\documentclass[10pt,journal,compsoc]{article}   

\usepackage{arxiv}
\usepackage{amsfonts}
\usepackage{amssymb}
\usepackage{amsbsy} 
\usepackage{amsmath}
\usepackage{cuted}
\usepackage{flushend}
\usepackage{caption}
\usepackage{subcaption}
\usepackage[dvips]{graphicx}
\usepackage[T1]{fontenc}
\usepackage[latin1]{inputenc}
\usepackage[table,xcdraw]{xcolor}

\usepackage{hyperref}
\hypersetup{
    unicode=false,          
    pdftoolbar=true,        
    pdfmenubar=true,        
    pdffitwindow=false,     
    pdfstartview={FitH},    
    pdftitle={IEEE submission},    
    pdfauthor={Prof.Fassman},     
    pdfsubject={RBIG-Information},   
    pdfcreator={Prof.Fassman},   
    pdfproducer={Prof.Fassman}, 
    pdfkeywords={keyword1} {key2} {key3}, 
    pdfnewwindow=true,      
    colorlinks=true,       
    linkcolor=blue,          
    citecolor=blue,        
    filecolor=blue,      
    urlcolor=blue           
}

\usepackage{multirow}

\usepackage{amsmath,epsfig}
\usepackage{array}
\usepackage{soul}
\usepackage[normalem]{ulem} 
\usepackage{times}
\usepackage[all]{xy}
\usepackage{flushend}
\usepackage{url}
\usepackage{rotating}
\usepackage{epstopdf}

\usepackage{float}
\usepackage{cite}
\usepackage{balance}
\usepackage{dsfont}

\def\x{{\mathbf x}}
\def\y{{\mathbf y}}
\def\X{{\mathbf X}}

\def\R{{\mathbf R}}
\def\A{{\mathbf A}}
\newcommand{\vect}[1]{\boldsymbol{#1}}

\title{Information Theory Measures via \\ Multidimensional Gaussianization}
\author{
  Valero Laparra\thanks{\url{https://www.uv.es/lapeva/}} \\
  Image Processing Laboratory \\
  Universitat de Val{\`e}ncia\\
  Val{\`e}ncia, Spain\\
  \texttt{valero.laparra@uv.es} \\
  \And
  J.~Emmanuel~Johnson\thanks{\url{https://jejjohnson.netlify.app}} \\
  Image Processing Laboratory \\
  Universitat de Val{\`e}ncia\\
  Val{\`e}ncia, Spain\\
  \texttt{juan.johnson@uv.es} \\
  \And
  Gustau Camps-Valls\thanks{\url{https://www.uv.es/gcamps/}} \\
  Image Processing Laboratory \\
  Universitat de Val{\`e}ncia\\
  Val{\`e}ncia, Spain\\
  \texttt{gcamps@uv.es} \\
  \And
  Raul Santos-Rodr\'iguez\thanks{\url{https://www.raulsantosrodriguez.com/}} \\
  Engineering Mathematics Department \\
  University of Bristol\\
  Bristol, UK\\
  \texttt{enrsr@bristol.ac.uk} \\
  \And
  Jesus Malo\thanks{\url{https://isp.uv.es/excathedra.html}} \\
  Image Processing Laboratory \\
  Universitat de Val{\`e}ncia\\
  Val{\`e}ncia, Spain\\
  \texttt{jesus.malo@uv.es} \\
}

\begin{document}
\maketitle


\begin{abstract}
Information theory is an outstanding framework to measure uncertainty, dependence and relevance in data and systems. It has several desirable properties for real world applications: it naturally deals with multivariate data, it can handle heterogeneous data types, and the measures can be interpreted in physical units. However, it has not been adopted by a wider audience because obtaining information from multidimensional data is a challenging problem due to the curse of dimensionality. Here we propose an indirect way of computing information based on a multivariate Gaussianization transform. Our proposal mitigates the difficulty of multivariate density estimation by reducing it to a composition of tractable (marginal) operations and simple linear transformations, which can be interpreted as a particular deep neural network. We introduce specific Gaussianization-based methodologies to estimate total correlation, entropy, mutual information and Kullback-Leibler divergence. We compare them to recent estimators showing the accuracy on synthetic data generated from different multivariate distributions. We made the tools and datasets publicly available to provide a test-bed to analyze future methodologies. Results show that our proposal is superior to previous estimators particularly in high-dimensional scenarios; and that it leads to interesting insights in neuroscience, geoscience, computer vision, and machine learning.
\end{abstract}


\section{Introduction}
     \label{sec:intro}
     First and second order estimators as mean, variance, covariance and correlation are ubiquitous in any data or system analysis. They provide summaries of the data, the uncertainty, the redundancy, and the similarity between data distributions. In a similar vein, the higher order summaries given by \emph{Information Theory Measures}~(ITMs)~\cite{Shannon1948} capture more complex features of the data or systems. 
Unlike simpler measures, ITMs allow one to compare different data sources because they are expressed in meaningful units related to the way signals may be encoded, transmitted and stored. This makes ITMs very valuable in situations where the Gaussian assumption is not fulfilled. 


ITMs have been applied in a number of fields.
For example, in \emph{pattern recognition} and \emph{machine learning}, ITMs have been used in general purpose techniques for feature extraction \cite{Nenadic07,HildETP06}, feature selection \cite{PengLD05,BalaganiP10}, or dictionary learning \cite{QiuPC14}, but also in more specific problems like regularization learning \cite{AchilleS18} or the description of the information bottleneck in deep learning \cite{WangCRWCC17}. ITMs are also useful for speech recognition~\cite{Qiao10} and data hashing \cite{CakirHBS19}.
In \emph{computer vision} and \emph{image processing}, ITMs have a central role in color constancy \cite{Marin-FranchF13},
sensor design \cite{Laparra2015}, shape registration \cite{HuangPM06}, or stereo vision \cite{Hirschmuller08}. In \emph{remote sensing} and \emph{geoscience}, entropy and mutual information have been used to analyze image classifiers\cite{paul2018spectral}, feature extraction and weighting ~\cite{zhang2018mutual}, and image registration methods \cite{xu2016multimodal} among others.
In \emph{neuroscience}, ITMs quantify the communication between different parts of the brain~\cite{Lizier11,Saproo14}, and efficiency of communication
may explain the emergence of neural structures and behaviors \cite{Barlow61,Barlow01,Olshausen96,Schwartz01,Malo06,Laparra15,Malo10,GomezVilla19}.
Comprehensive reviews of ITMs in biological systems can be found in \cite{Bialek16,Timme2018}.


Despite the widespread interest of ITMs, most people tend to use 2nd order measures in multivariate scenarios because they are easy to compute and very reliable. ITMs suffer from \emph{the curse of dimensionality}~\cite{Bellman61,LeeVerleysen07} since they require an accurate PDF estimation from the data which becomes infeasible in high dimensions. This prohibits their use in applications with medium-to-high data dimensionality and thus have been mostly restricted to low dimensional problems.


In this paper we propose the use of Gaussianization as a way to circumvent the curse of dimensionality in the estimation of several information-theoretic measures from data.
The core of our proposal is the Rotation-Based Iterative Gaussianization (RBIG) proposed in ~\cite{Laparra2011b}. The method learns an invertible transformation to a multivariate Gaussian domain. As consequence one could estimate the probability of points in the original domain efficiently by applying the change of variables formula. RBIG breaks the hard PDF estimation problem into a set of tractable and computationally efficient blocks.
This formulation can be interpreted as a generative model parametrized by deep learning architecture, and can be actually cast in the \emph{density destructors} framework~\cite{Inouye2018DeepDD,Johnson19}.


Nevertheless, our proposal for ITMs is not based on PDF estimation, but on a different property. 
We show how any differentiable Gaussianization can provide an estimate of \emph{total correlation}
as a by-product of the transformation. We demonstrate that this by-product is particularly easy to compute in RBIG as a sum of marginal entropy differences in each layer. This efficient estimation of \emph{total correlation} via RBIG allows us to introduce a set of practical estimators of other relevant ITMs for continuous variables. 


The specific contributions of this paper are:
\begin{enumerate}
    \item {\it Gaussianization strategies to estimate ITMs.}
    We prove the special role of total correlation in RBIG. This explains why RBIG is the best choice among Gaussianization transforms to estimate ITMs.

    \item \textit{Formal definition of Gaussianization-based ITMs.} We introduce specific RBIG-based methodologies to estimate \emph{total correlation} ($T$), \emph{entropy} ($H$), \emph{Kullback-Leibler divergence} ($D_{\textrm{KL}}$), and \emph{mutual information} ($I$), from multidimensional data.

    \item \textit{Systematic comparison with other estimates.} We provide exhaustive experimental evidence of performance of the RBIG-based ITMs in examples involving different families of distributions with known analytical result. In every case, bias and variance is given as a function of the number of samples and dimensions. More importantly, RBIG estimates are systematically compared to other methodologies reported    in~\cite{szabo14information}.

    \item \textit{Applications in real data for each ITM.}
    We illustrate how the RBIG-based estimators can be applied to real-world data science problems from diverse fields such as visual neuroscience, geoscience, computer vision, and learning analysis.

    \item \textit{A toolbox to compute ITMs, \url{https://isp.uv.es/RBIG4IT.htm}. } We made publicly available two implementations of a toolbox (Matlab and Python), including the estimators for the ITMs and auxiliary routines to reproduce the experiments. This can be used as a solid test-bed to systematically check future information estimates.
\end{enumerate}


The structure of the paper is as follows. 
First we give a short introduction to the information theory measures studied in this work. 
After that we explain the theoretical advantages of RBIG to estimate these measures. 
Finally for each measure we show how to compute it using RBIG, we show results on synthetic data comparing the performance with other methods, and applications in different disciplines, from 
computer vision and visual neuroscience to machine learning and geosciences.


     \section{Background}


\subsection{Information Theory Measures}

\begin{figure}[t!]
    \centering
    \includegraphics[width=7.5cm]{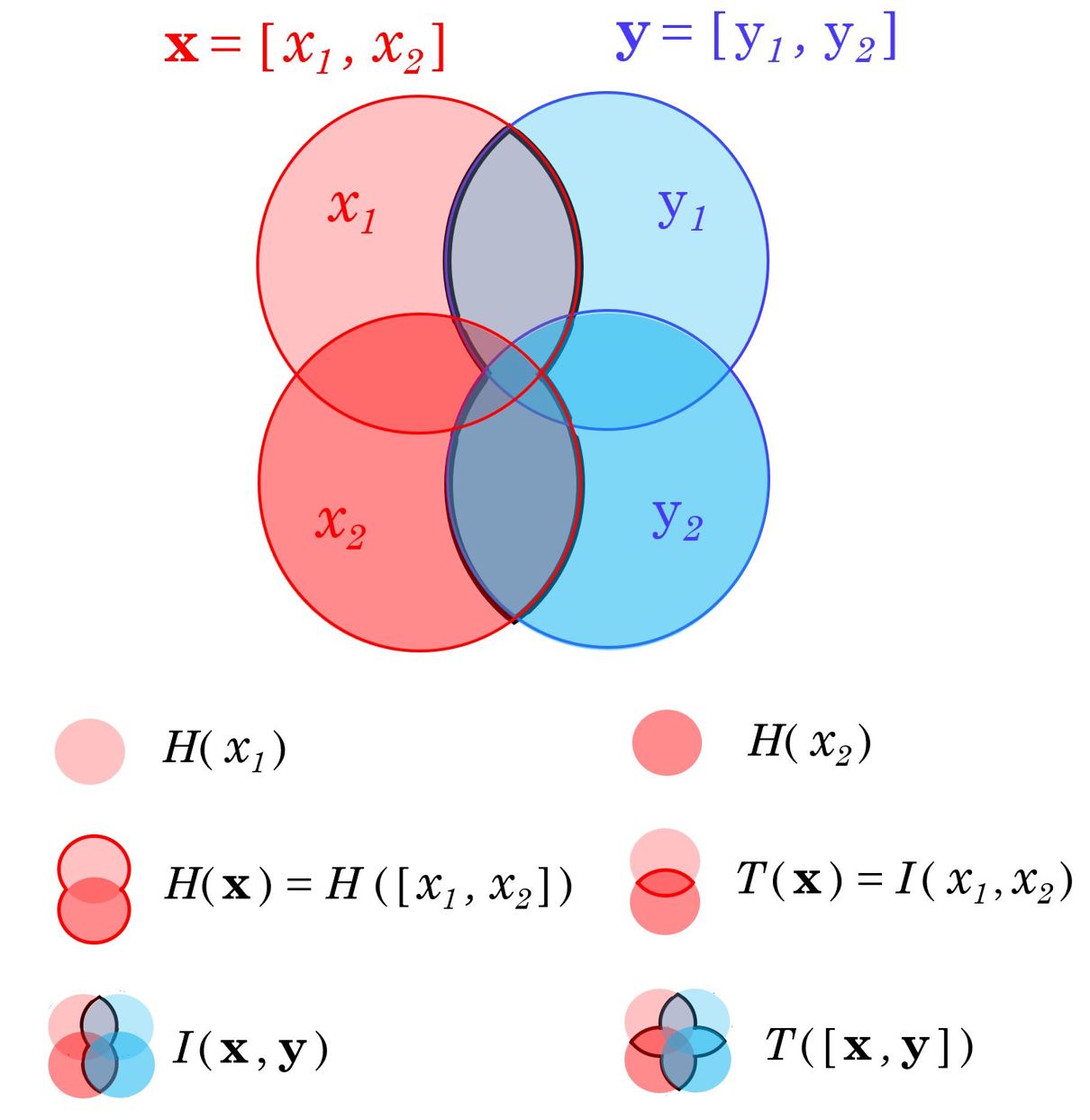}
    \caption{Conceptual scheme of information theoretic measures. $\x = [x_1,x_2]$ and $\y=[y_1,y_2]$ are two-dimensional random variables. Areas represent amounts of information, and intersections represent shared information among the corresponding variables and dimensions. Examples of entropy, total correlation and mutual information are given.}
    \label{fig:Fig_1}
\end{figure}

A comprehensive review of the information theoretic measures discussed in this work (namely total correlation, entropy, Kullback-Leibler divergence, and mutual information) is available elsewhere.
For instance, \cite{Cover05} is an excellent reference for differential entropy, $H$, mutual information, $I$, and Kullback-Leibler Divergence, $D_{\textrm{KL}}$; and~\cite{Watanabe1960} is an appropriate source for total correlation, $T$, also known as multi-information~\cite{Studeny98}. However, let us recall here the relation between $H$, $I$, and $T$, through the Venn Diagram in Fig.~\ref{fig:Fig_1}, which allows us to introduce the basic identities required throughout the work. 
This diagram represents the uncertainty and shared information in two multivariate random variables, $\x = [x_1,\ldots,x_{D_x}] \in \mathbb{R}^{D_x}$, and $\y =  [y_1,\ldots,y_{D_y}] \in \mathbb{R}^{D_y}$.
In particular, we consider two-dimensional variables, $D_x = D_y = 2$, just for visualization purposes, but the interpretation of the diagram is general and dimension could be bigger and not necessarily the same for $\x$ and $\y$. In this diagram each set represents the information provided by each univariate component.

The \emph{univariate differential entropy} ($H(x_i)$ or $H(y_i)$ in the diagram) is the expected value of the provided information by the $i$-th dimension of the variable:
\begin{equation}
    H(x_i) = \mathbb{E}_{x_i} [-\log(p_{x_i}(x_i))],
    \label{eq:entropy}
\end{equation}
where we have used the definition of information given by Shannon~\cite{Shannon1948}. In our diagram, the entropy corresponds to the area of the set. This definition can be extended to multidimensional variables. In this case, the \emph{joint differential entropy}, $H(\x) = H([x_1,\ldots,x_{D_x}])$, is given by the union of the univariate sets.

The \emph{Total correlation} describes the information shared by \emph{several} univariate components, and hence it can be defined either within a vector, i.e. $T(\x)$ is the intersection of the sets corresponding to $x_i$; or for concatenation of an arbitrary number of multivariate variables, i.e. $T([\x,\y])$ is the intersection of all the components in $[\x,\y] \in \mathbb{R}^{D_x+D_y}$. This intersection accounts for the redundancy between the dimensions of the considered set of variables. Therefore, $T$ can be computed as the difference between the sum of the entropies of the univariate sets minus the union:
\begin{equation}
    T(\x) = \sum_{i=1}^{D_x} H(x_i) - H(\x).
    \label{eq:TC}
\end{equation}
The \emph{mutual information} also describes the shared information between random variables but only between \emph{two of them}, which can be univariate or multivariate, and they are not restricted to have the same dimension.
For example, as in $I(x_1,x_2)$ or in $I(\x,\y)$.
The Venn diagram illustration is useful to show that, given the shared information nature of $I$, it is related to the differential entropies:
\begin{equation}
    I(\x,\y) = H(\x) + H(\y) -  H([\x,\y]).
    \label{H_I}
\end{equation}
Note that, depending on the group of variables or dimensions considered as input some measures may be equivalent. For instance, in our two-dimensional example $T(\x) = T([x_1,x_2]) = I(x_1,x_2)$.

The \emph{Kullback-Leibler Divergence} $D_{\textrm{KL}}$ cannot be easily included in the Venn Diagram of Fig.~\ref{fig:Fig_1} because its definition depends not only on the entropy but also on the cross-entropy, $D_{\textrm{KL}(\x|\y)} =  \mathbb{E}_{x_i} [-\log(p_{y_i}(y_i))] - H(x_i)$. The $D_{\textrm{KL}}$ measures a divergence (not a distance) between two PDFs~\cite{Cover05}.

It is worth noting that all four measures ($H$, $T$, $I$, and $D_{\textrm{KL}}$) are defined in the same units.


\subsection{Conventional Estimators}
\label{sec:intro_methods}

While plenty of methods focus on the estimation of $H$, $T$, $I$ and $D_{\textrm{KL}}$ for one dimensional or two-dimensional variables, there are few methods that deal with variables of more dimension. Our comparison is based on the broad family of ITM estimators included in~\cite{szabo14information}, which can be used in the general multivariate case, and we used their default parameterizations in all the experiments. 

We selected a number of conventional estimators that compute Shannon entropy of a continuous random variable and, by extension, can estimate the mutual information as seen in equation \ref{eq:MI_as_H}.

The simplest ITM estimators assume that the data distributions are from the exponential family, in our case we used the Gaussian distribution. With this assumption, the aforementioned ITMs are straightforward to calculate. We refer to these estimates as the \emph{exponential family} (\textbf{expF}).
The Gaussian family has attractive mathematical properties.
The Sharma-Mittal entropy yields an analytical expression for the entropy of a Gaussian distribution~\cite{Nielsen_2011} which also includes the derivative of the log-Normalizer of the distribution which acts as a corrective term yielding better estimates of the distribution.
The experiments below show that the Gaussian assumption performs well on simple distributions, but has lower performance for distributions which are characterized by their tails like the T-Student or the Pareto distribution. This is a key factor in many scientific fields where actual data are not Gaussian.

The next family of popular methods are those using binning strategies \cite{KumarPothapakula2019}.
These binning methods include algorithms like the ridge histogram estimation, the smooth kernel density estimator (KDE),
and the adaptive $k$-Nearest Neighbors (\textbf{kNN}) estimator \cite{Kozachenko1987ASE,Goria05}. These methods are non-parametric which allows them to be more flexible and should be able to fit more complex distributions. However, these methods are sensitive to the parameters chosen to define the neighborhoods and there is no intuitive method to select the appropriate parameters. Besides, their parameterization is data-dependent. 
We select the most robust method, the kNN estimator, as representative of this family in our experiments. kNN is adaptive in nature due to the neighbour structure from a distance matrix.
In all the experiments we choose the default value for $k$. Standard kNN algorithms are known to have scaling problems due to the high dimensionality. Therefore, we also include the scaled version which uses partition trees \cite{Stowell09},
and here is referred to as \textbf{KDP}.
We also utilize the batch method which attempts to estimate the entropy in batches and then it takes the mean of the estimates over the ensemble (\textbf{Ens}). This has been shown to be effective in applied settings such as image registration \cite{Kybic2004}.

We also include a more recent method which estimates these measures via von Mises Expansions (\textbf{vME}) \cite{NIPS2015_5911} which uses a functional Taylor expansion that computes the second order entropy. This non-parametric approach has been shown to have a fast rate of convergence compared to the traditional kNN family.

     \label{sec:it}

     \section{Multivariate Gaussianization}

     \label{sec:gauss_trans}

\begin{figure*}
\centering
\includegraphics[width = 18cm]{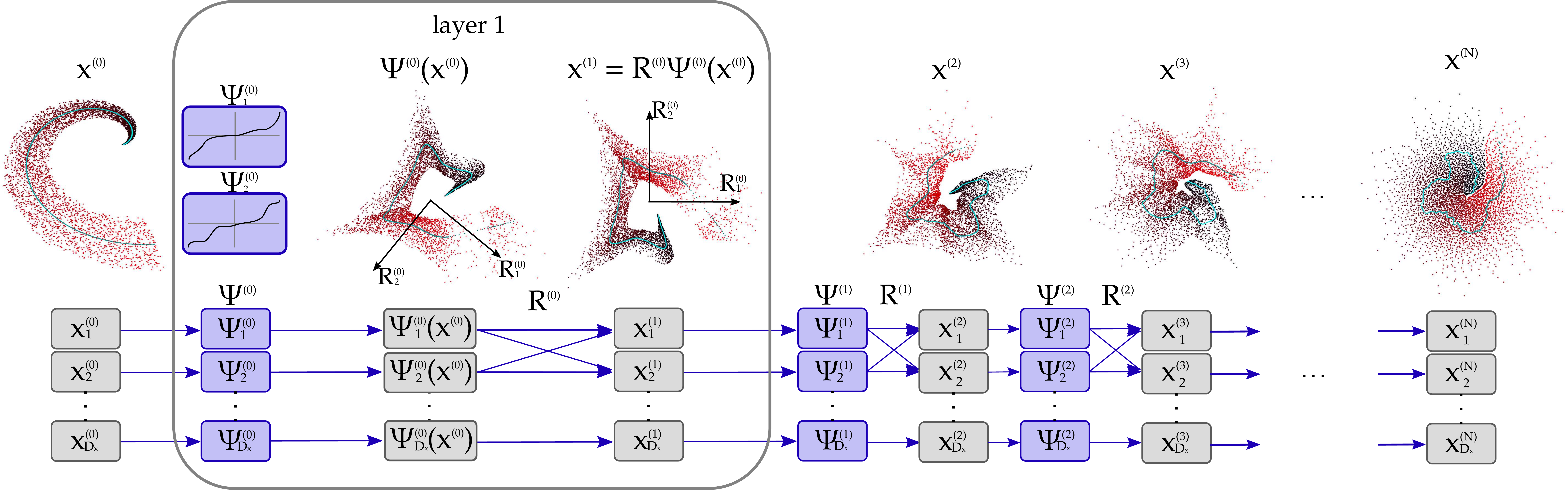} \caption{\label{rbig}
\small Gaussianization of an arbitrary non-Gaussian dataset using the Rotation-Based Iterative Gaussianization (RBIG) \cite{Laparra2011b}.
This Gaussianization algorithm is a cascade of nonlinear+linear transforms: the marginal Gaussianizations $\psi^{(n)}$ and the rotations $\R^{(n)}$.
The dots are colored and a blue line passing through the dataset has been added in order to follow how the dataset is modified in each layer of the network. In the first layer box we explicitly show the specific marginal Gaussianization transforms $\psi^{(0)}_i$ for the each dimension $i$ and the specific rotation $\R^{(0)}_i$ directions for the first iteration. 
Although the toy example is only bidimensional, in the bottom a representation of the RBIG network for generic dimensionality datasets is presented. See Section \ref{sec:RBIG} for details.
}
\end{figure*}

Given samples drawn from an arbitrary PDF, $\x \sim p(\x)$, a Gaussianization transform is a feature map,
\begin{equation}
    \x  \xrightarrow{\,\,\, G_x \,\,\,} \x' \label{Gauss},
\end{equation}
so that the output follows a zero-mean unit-covariance Gaussian, $\x' \sim \mathcal{N}(\x',\mathbf{0},\mathbf{I})$, where $\mathbf{0}$ is a vector of zeros (for the means) and $\mathbf{I}$ is the identity matrix (for the covariance).
See an example in Fig.~\ref{rbig}.

\subsection{Gaussianization is useful to estimate information}
If Gaussianization mappings are differentiable, as is usually the case~\cite{Friedman74,Friedman84,Chen2001,Lyu2009,Laparra2011b,Johnson19,Inouye2018DeepDD
,Kingma2014VAES,Rezende2015NF,BalleLS15}, they may be useful to estimate ITMs because of two reasons:
(1)~these mappings can be used first to estimate the PDFs via the standard transform of probability under smooth mappings~\cite{Stark95}, $p(\x) = |\nabla G_x(\x)| \, \mathcal{N}(\x',\mathbf{0},\mathbf{I})$, and afterwards, estimate the corresponding measure from the PDFs, and
(2)~they can be used to estimate the total correlation from data samples, and then, the remaining information theory measures can be related to $T$.
Gaussianization is appropriate to compute $T(\x)$ because the components of its output, $\x' \sim \mathcal{N}(\x',\mathbf{0},\mathbf{I})$, are statistically independent, and hence $T(\x')=0$. Therefore, the variation of total correlation under smooth mappings~\cite{Lyu2009}, which for transforms that do not preserve dimension (rectangular Jacobian $\nabla G$) generalizes to,
\begin{eqnarray}
    \Delta T(\x,\x') \!\! &=& \!\! T(\x) - T(\x')  \nonumber \\
                    \!\!  &=& \!\! \sum_{i=1}^{D_x} H(x_i) - \sum_{i=1}^{D_{x'}} H(x'_i)  \label{deltaT} \\ \!\!  & & \!\! + \frac{1}{2}\mathbb{E}_x \Big( \log |\nabla G_x(\x)^\top  \nabla G_x(\x)| \Big) \nonumber
\end{eqnarray}
in the case of Gaussianization, where $D_{x'}=D_{x}$ and $T(\x')=0$, leads to:
\begin{equation}
    T(\x) = \sum_{i=1}^{D_x} H(x_i) - \frac{D_x}{2} \log(2\pi e) + \mathbb{E}_x \Big( \log |\nabla G_x(\x)| \Big) \label{eq:T_definition}
\end{equation}
These two strategies (either estimating the PDF, or estimating $T$) illustrate the theoretical and practical interest of differentiable Gaussianization transforms. However, direct application of these ideas for an arbitrary Gaussianization may not be straightforward. In the first strategy, errors in the estimated PDF may accumulate when integrating along the domain, and multidimensional integration is not trivial. In the second strategy, Eq.~\eqref{deltaT} relies on an average over all the samples, the matrix $\nabla G_x(\x)$, which is multivariate (i.e. time consuming and also prone to estimation errors).

Therefore, all differentiable Gaussianizations are eventually applicable for information estimation but only certain transforms could overcome the above issues.

\subsection{Previous differentiable Gaussianization transforms}
The idea of designing a method to estimate multivariate densities through iterative marginal operations in order to avoid the curse of dimensionality was originally proposed in \cite{Friedman84}. This work was based on the idea of projection pursuit~\cite{Friedman74} which is also one of the seminal works for what later will be known as Independent Components Analysis (ICA).
A step forward was given in \cite{Chen2001} where the projection-pursuit-like Gaussianization was based on an iterative cascade of two operations: a multivariate \emph{linear ICA} and a set of univariate (marginal) gaussianizations based on mixtures of Gaussians. The advantage was the use of \emph{fast linear ICA} techniques developed in the late 90's~\cite{Hyvarinen01}. However the problem of the above methods inspired on projection pursuit is their focus on the multivariate (time-consuming) problem: the ICA transform looking for the most non-Gaussian direction of the data in each iteration.

More recently, transforms to latent spaces with Gaussian PDF have been proposed in the context of deep-learning. Examples include Variational AutoEncoders \cite{Kingma2014VAES}, Generative Adversarial Networks \cite{Goodfellow2014GANS} and Invertible Flows \cite{Rezende2015NF}. Each family tackles the underlying PDF estimation problem from a slightly different {\em algorithmic} perspective, but they share many {\em conceptual} properties.
They look for two main components: the first looks for a function that maps samples from a latent space $\x'$ to the observed space $\x$, and the second aims to find a function that maps data from the observed space $\x$ to some latent space $\x'$. The latter component are called {\em density  destructors}~\cite{Inouye2018DeepDD} because their aim is removing the structure of the input data distribution.
These density destructors had not been used to estimate information-theoretic variables nor had been connected to previous projection-pursuit transforms.
However, this gap has been filled~\cite{Johnson19} by linking the deep-learning family and the projection-pursuit family with special emphasis on information theory through a specific algorithm that may be interpreted in both ways: the Rotation-Based Iterative Gaussianization (RBIG)~\cite{Laparra2011b}.

In the context of estimating information theory quantities the particular Gaussianization we propose to use, RBIG, has two major advantages with regard to other techniques:
\begin{enumerate}
    \item The cascade of nonlinear+linear transforms was shown to converge to a Gaussian even if the linear transforms are random rotations. This alleviates the requirement for linear ICA in classical projection-pursuit techniques, and the requirement to learn the typically large amount of parameters of the linear transforms in the deep-learning family.

\item RBIG has special properties regarding total correlation that allow to follow the strategy suggested by Eq.~\eqref{deltaT}, but using only marginal operations, thus overcoming the requirement to compute and average the Jacobian.
\end{enumerate}

The first of these properties was proved in the original paper~\cite{Laparra2011b}, and the second, more relevant for information theory, is analyzed in detail below in Section~\ref{sec:key}.

\subsection{The Rotation-Based Iterative Gaussianization}
\label{sec:RBIG}
The Gaussianization strategy that we follow, RBIG, was introduced in \cite{Laparra2011b} and is illustrated in Fig.\ref{rbig}. It is based on the sequential (layer wise) application of two operations: nonlinear marginal Gaussianizations, $\Psi$, and linear rotations, $\R$:
\begin{equation}
      \x^{(n+1)} = \R^{(n)} \Psi^{(n)}(\x^{(n)})
      \label{rbig_iter}
\end{equation}
where $n$ refers to the step in the sequence and convergence takes $N$ steps, $n=1,\ldots,N$.
The original data (the data in the step 0) is $\x^{(0)} = \x$, and the output of the transform is $\x^{(N)} = \x'$.
The marginal Gaussianization consists of a set of marginal operations (dimension-wise): marginal equalization from the cumulative density function, and marginal transform from uniform to Gaussian using the inverse cumulative density function of a standard univariate Gaussian. 
Of course all the components of the marginally Gaussianized variable follow a standard univariate Gaussian, ${\Psi}_i^{(n)}(x_i^{(n)}) \sim \mathcal{N}(0,1) \,\,\, \forall i = 1,\ldots,D_x$, that can be collectively grouped as $\boldsymbol{\Psi}^{(n)}(\x^{(n)}) = [\Psi_1^{(n)}(x_1^{(n)}),\ldots,\Psi_{D_x}^{(n)}(x_{D_x}^{(n)})]$. However, note that at some intermediate point in the series the joint PDF is not a multivariate Gaussian (yet). See for instance the evolution in Fig.~\ref{rbig} which was computed with RBIG.

The following operation is a rotation, $\R$, which may be sophisticated for quicker convergence (e.g. PCA or ICA matrices), but convergence is also guaranteed even for random orthogonal matrices~\cite{Laparra2011b}. The number of layers can be easily decided using a convergence criteria~\cite{Laparra2011b}.

\section{The Special Role of Total Correlation in RBIG}
\label{sec:key}

RBIG is particularly convenient to propose ITM estimators because
it breaks the multivariate estimation of total correlation in multiple univariate entropy estimations.
This ability to break the multivariate problem emerges naturally from the way the RBIG network is constructed 
and can be deduced from the following property.

\noindent{\bf Lemma.}
The change of total correlation in each RBIG layer reduces to a sum of univariate entropies:
\begin{eqnarray}
    \Delta T^{(n)} &=& T\Big(\x^{(n)}\Big) - T\Big(\x^{(n+1)}\Big) \nonumber \\[0.2cm]
    &=& \frac{D_x}{2} \log(2\pi e) - \sum_{i=1}^{D_x} H\Big(x^{(n+1)}_i\Big)
    \label{lemma}
\end{eqnarray}

\noindent{\bf Proof.} The difference in total correlation before and after the application of a layer is:
\begin{equation}
    \Delta T^{(n)} = T\Big(\x^{(n)}\Big) - T\Big( \R^{(n)} \Psi^{(n)}(\x^{(n)}) \Big), \nonumber
\end{equation}
since total correlation is invariant under marginal invertible transforms~\cite{Studeny98} and $\Psi^{(n)}$ is a marginal invertible transform we have:
\begin{equation}
    \Delta T^{(n)} = T\Big(\Psi^{(n)}(\x^{(n)})\Big) - T\Big( \R^{(n)}  \Psi^{(n)}(\x^{(n)}) \Big) \nonumber
\end{equation}
And then, applying Eq.~\eqref{deltaT}, we have:
\begin{eqnarray}
   \hspace{-0.5cm} \Delta T^{(n)} = \sum_{i=1}^{D_x} H\Big(  \Psi_i^{(n)}(x_i^{(n)}) \Big) \!\!\!\!\!&-&\!\!\!\!\! \sum_{i=1}^{D_x} H\Big( \big( \R^{(n)} \Psi_i^{(n)}(x_i^{(n)}) \big)  \Big) \nonumber \\
    \!\!\!\!\!&+&\!\!\!\!\! \mathbb{E}\Big( \log(|\R^{(n)}|) \Big)
    \nonumber
\end{eqnarray}
where the first term is just $D_x$ times the known entropy of the marginal $\mathcal{N}(0,1)$, the second term is the sum of marginal entropies of the transformed variable, $\x^{(n+1)}$, and the last term vanishes because $\R^{(n)}$ is restricted to be a rotation and has unit-determinant. This leads to Eq.~\eqref{lemma}. 

The univariate nature of the computation of variations of $T$ in RBIG layers
makes this specific Gaussianization better suited to obtain ITMs from $T$
than other Gaussianization transforms. In RBIG one is not forced to
compute the matrices $\nabla_x G(\x)$ and average them over the multidimensional
space as in Eq.~\eqref{eq:T_definition}.

     \section{RBIG-based Estimators}

    In the following subsections we propose RBIG-based estimators for different information-theoretic quantities: $T$ , $H$, $D_{KL}$, and $I$. In every case, we follow this structure:
\begin{enumerate}
\item We derive the RBIG-based estimator.
\item We validate its performance by comparing the result with alternative estimators on synthetic data from known PDFs where the analytical result is known.
\item We give examples of applicability in practical problems with real-world data where the PDF is unknown.
\end{enumerate}

      \subsection{Total Correlation, \texorpdfstring{$T(\x)$}{T(x)}}
      \label{sec:TC}

Total correlation (also known as multiinformation) is the amount of information shared among the different dimensions of a multidimensional distribution. 
The definition was given in Eq.~\eqref{eq:TC} and illustrated in Fig.~\ref{fig:Fig_1}.

\subsubsection{Estimation using RBIG}
When applying RBIG to particular data, $\x$, the output 
has zero total correlation, i.e. $T(\x^{(N)}) = 0$. Therefore, by applying  Eq.~\eqref{lemma}, the total correlation of the original data is the sum of the $T$ variation at each layer:
\begin{eqnarray}
      \Tilde{T}(\x) &=& \sum_{n=0}^{N-1} \Delta \Tilde{T}^{(n)} \label{Estim_T} \\
                    &=& \frac{(N-1) D_x}{2} \log(2\pi e) - \sum_{n=1}^{N} \sum_{i=1}^{D_x} \Tilde{H}(x_i^{(n)}). \nonumber
                    \label{eq:TC_from_RBIG}
\end{eqnarray}
Note that the estimation of this (challenging) multivariate magnitude reduces to a set of (easy-to-compute) univariate marginal entropy estimations (sec. \ref{sec:key}). Intuitively, each RBIG layer removes part of the total correlation present in the original data, introducing as many layers as necessary until it is removed.

\subsubsection{Validation on known PDFs}
\label{sec:val_TC}

\begin{figure*}
    \centering
    \begin{tabular}{m{0.8mm}m{41mm}m{41mm}m{41mm}m{41mm}}
      & $D = 3$ & $D = 10$ & $D = 50$ & $D = 100$ \\
\parbox[t]{2mm}{\multirow{1}{*}{\rotatebox[origin=c]{90}{Gaussian}}} &
	\includegraphics[width=0.95\linewidth]{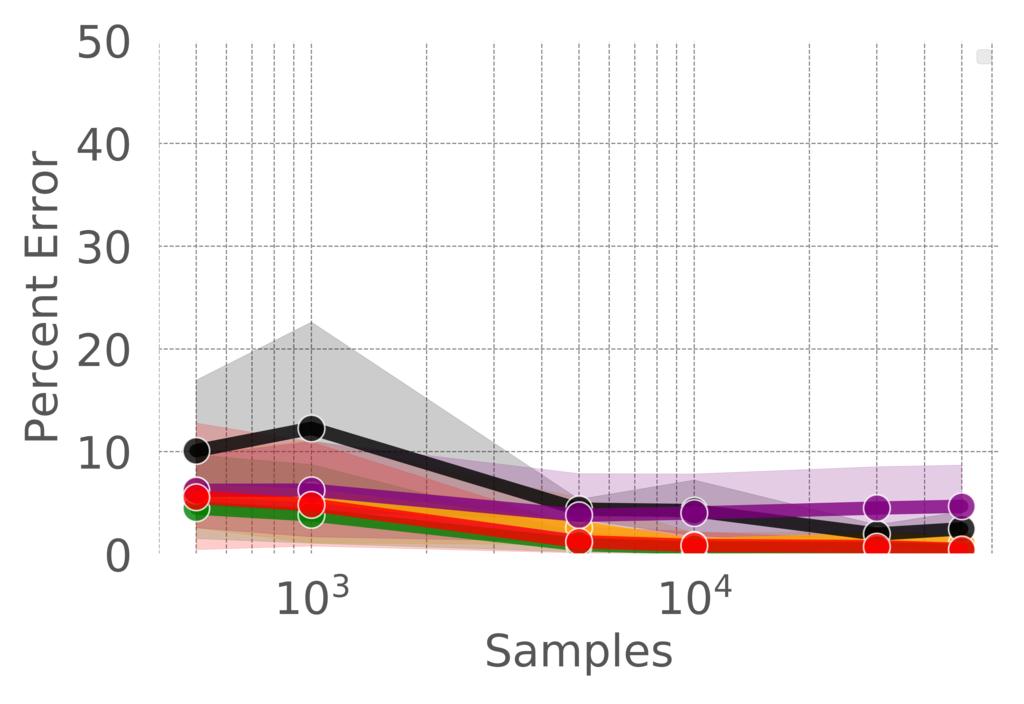} &
    \includegraphics[width=0.95\linewidth]{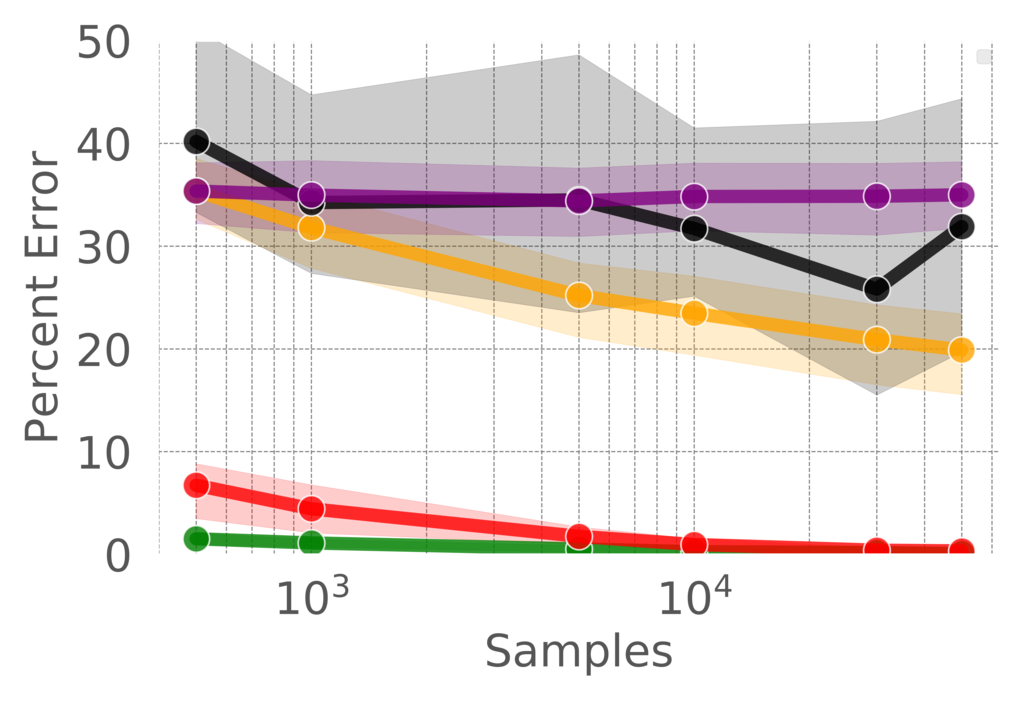} &
    \includegraphics[width=0.95\linewidth]{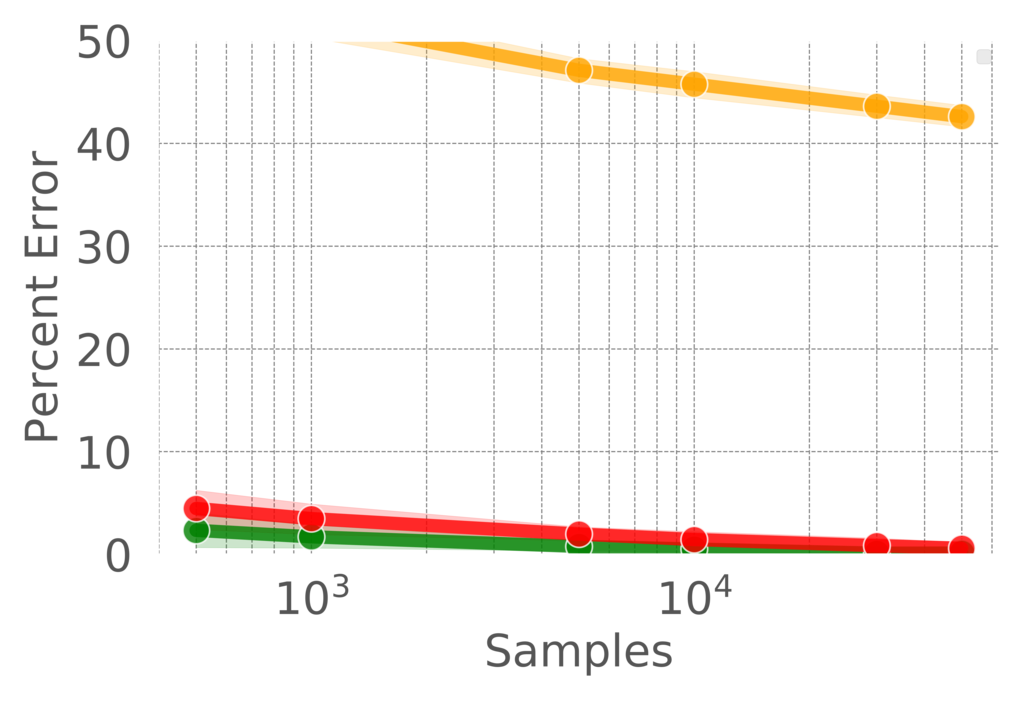} &
    \includegraphics[width=0.95\linewidth]{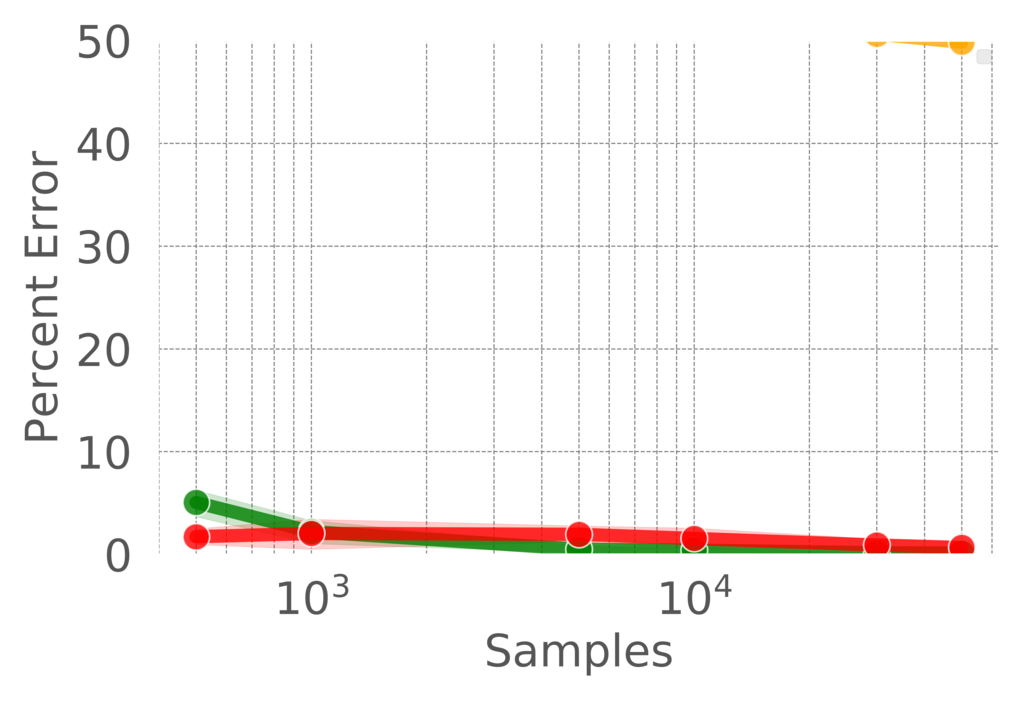} \\
\hline
\parbox[t]{2mm}{\multirow{1}{*}{\rotatebox[origin=c]{90}{Uniform}}} &
    \includegraphics[width=0.95\linewidth]{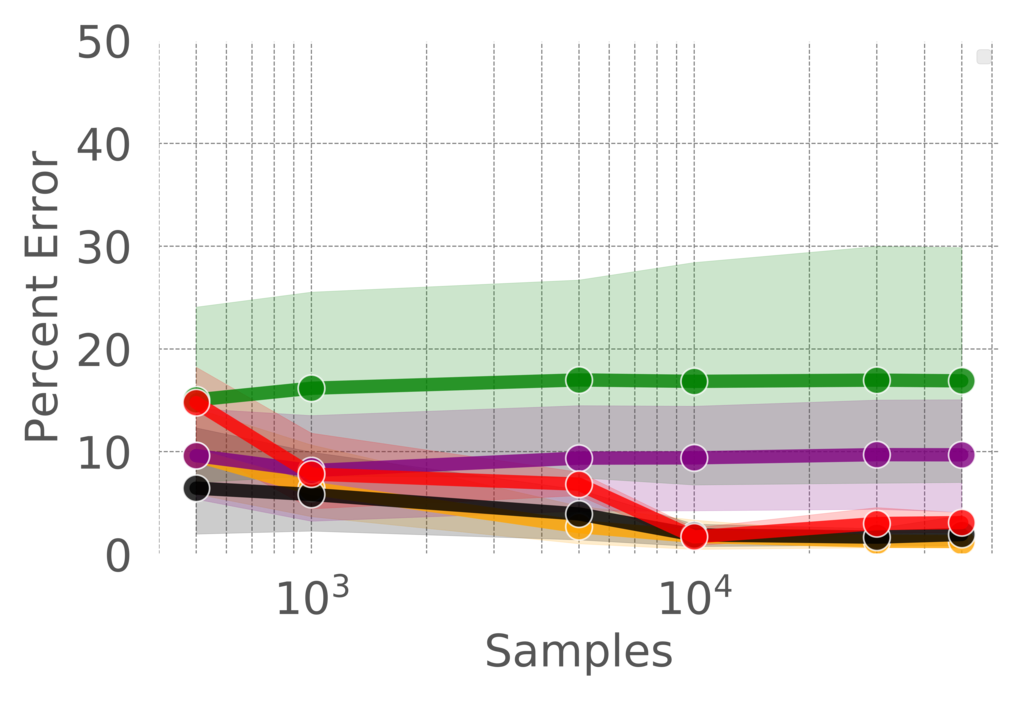} &
    \includegraphics[width=0.95\linewidth]{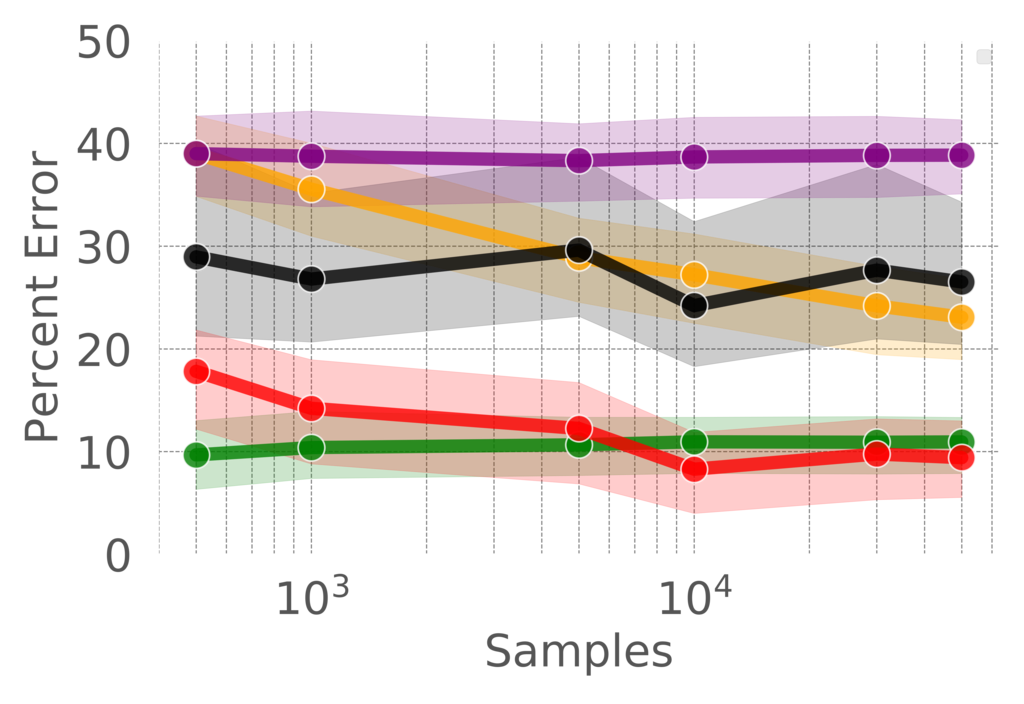} &
    \includegraphics[width=0.95\linewidth]{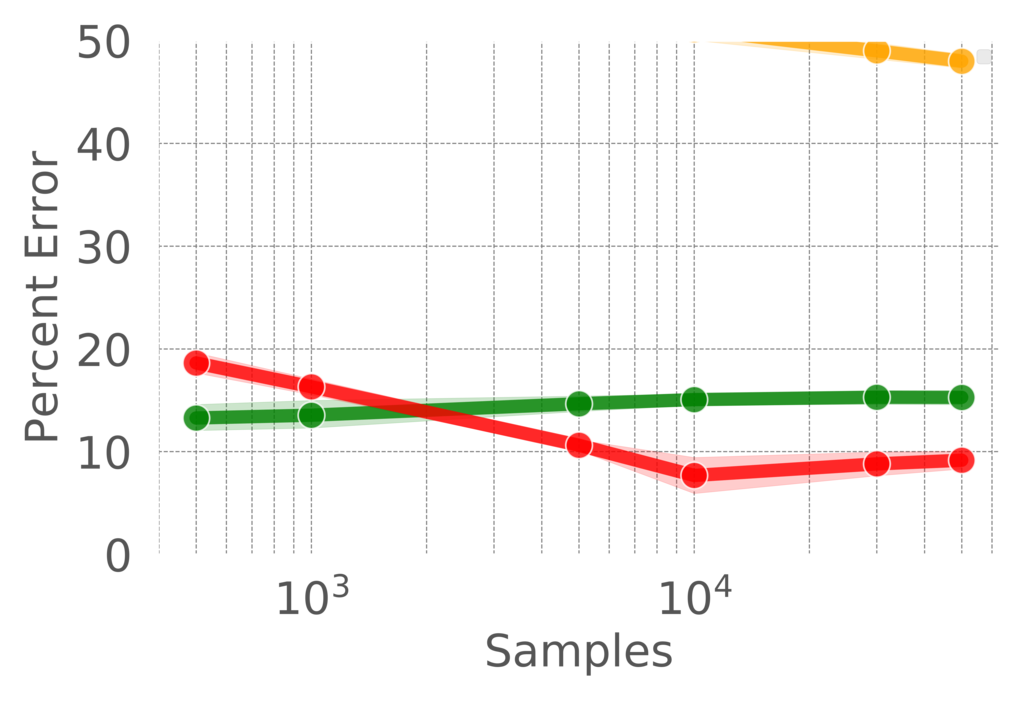} &
    \includegraphics[width=0.95\linewidth]{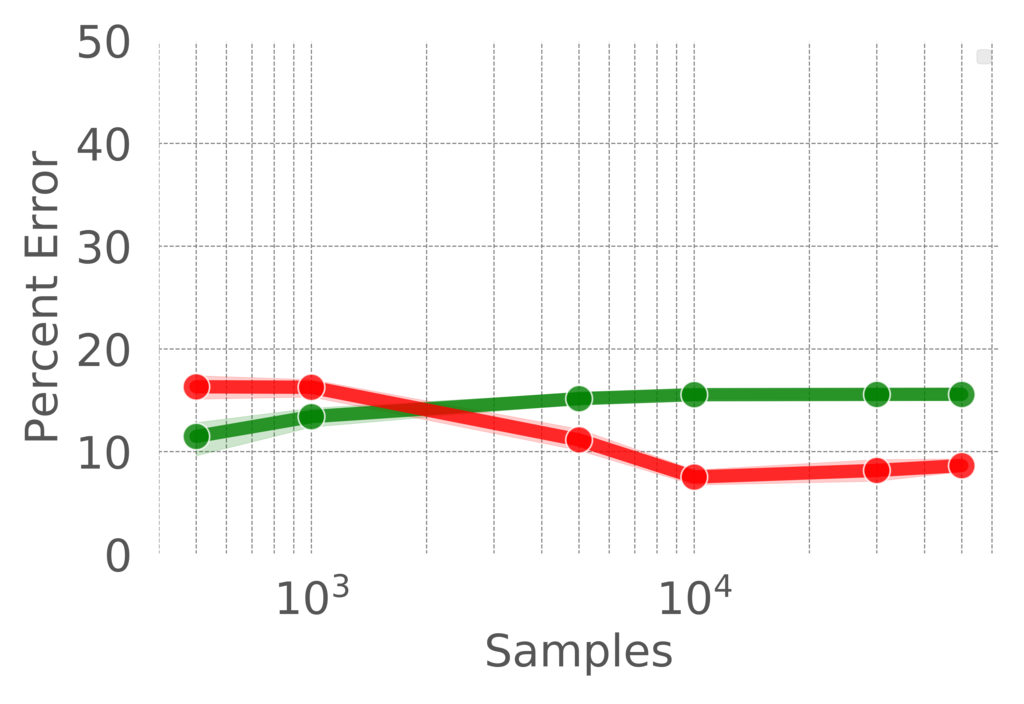}\\
\hline
\parbox[t]{2mm}{\multirow{18}{*}{\rotatebox[origin=c]{90}{Student}}} &
     \includegraphics[width=0.95\linewidth]{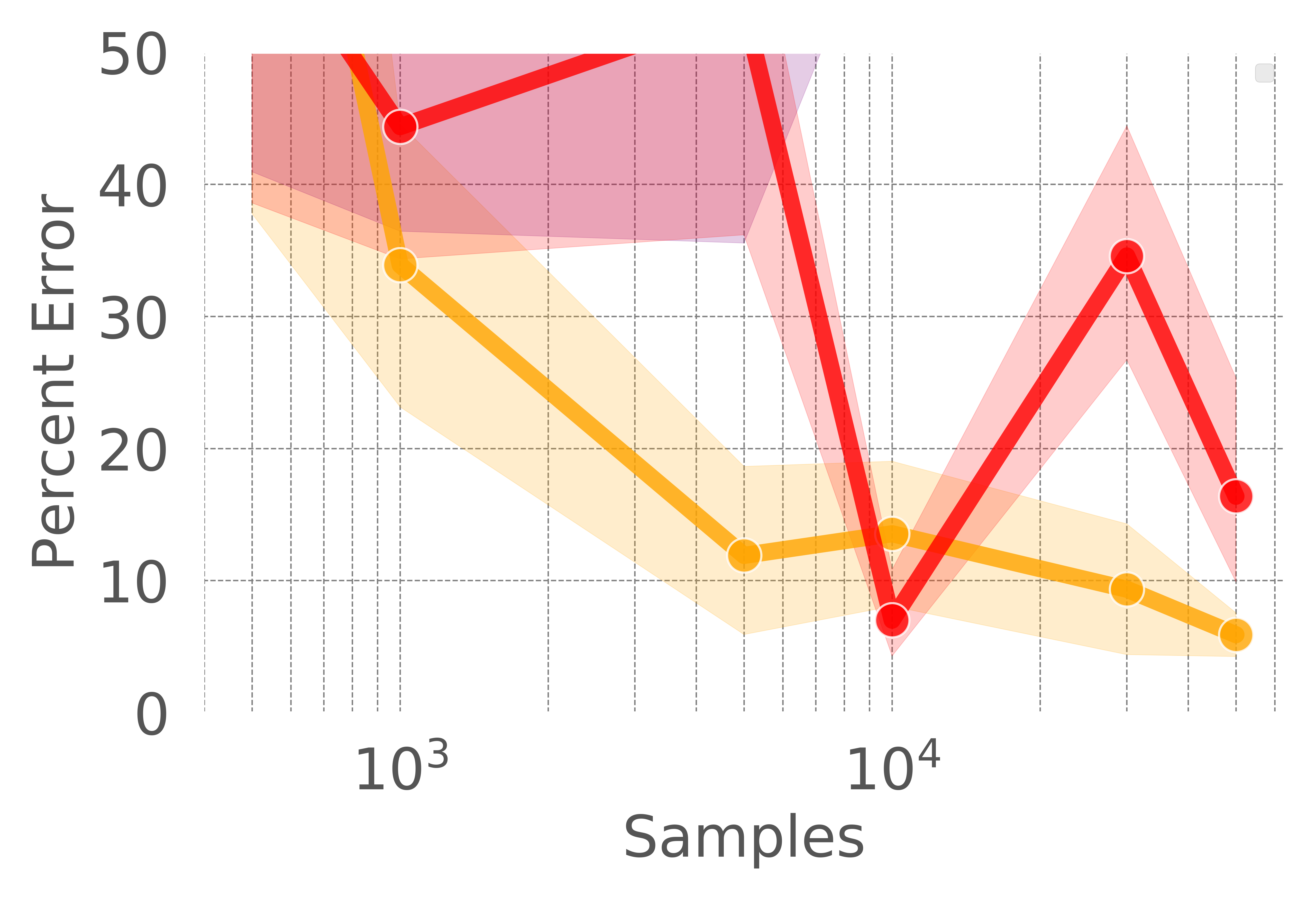} &
     \includegraphics[width=0.95\linewidth]{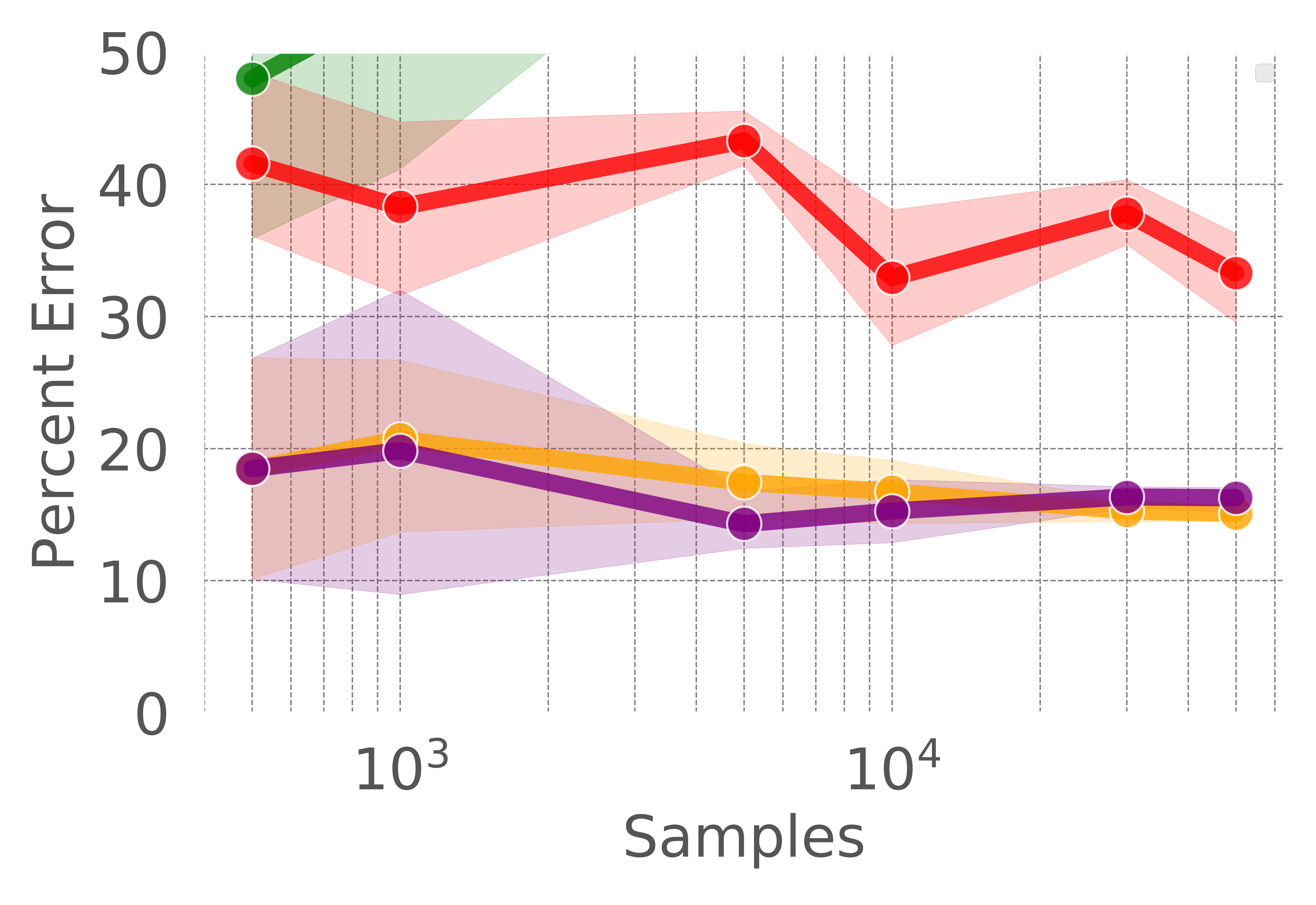} &
     \includegraphics[width=0.95\linewidth]{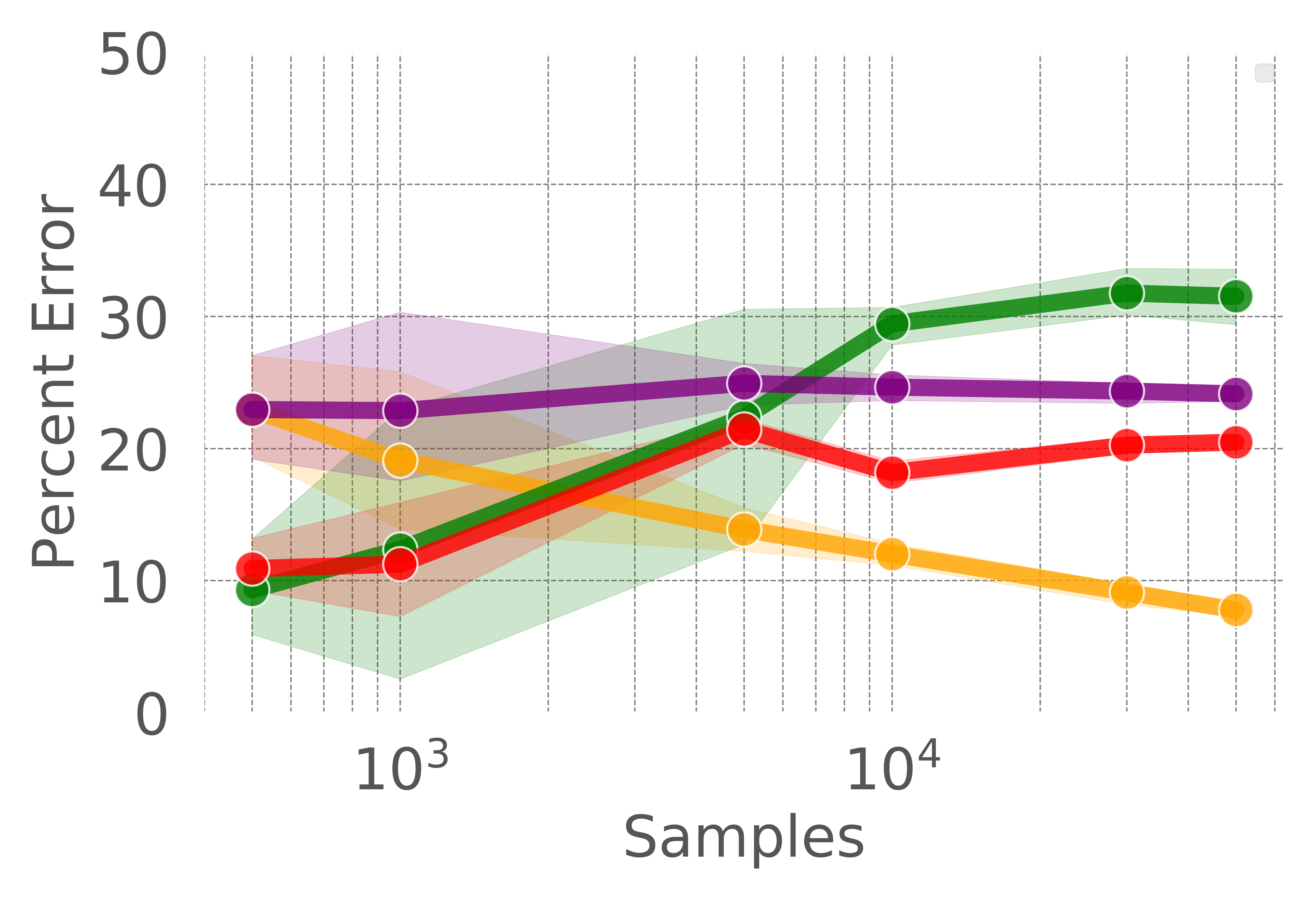} &
     \includegraphics[width=0.95\linewidth]{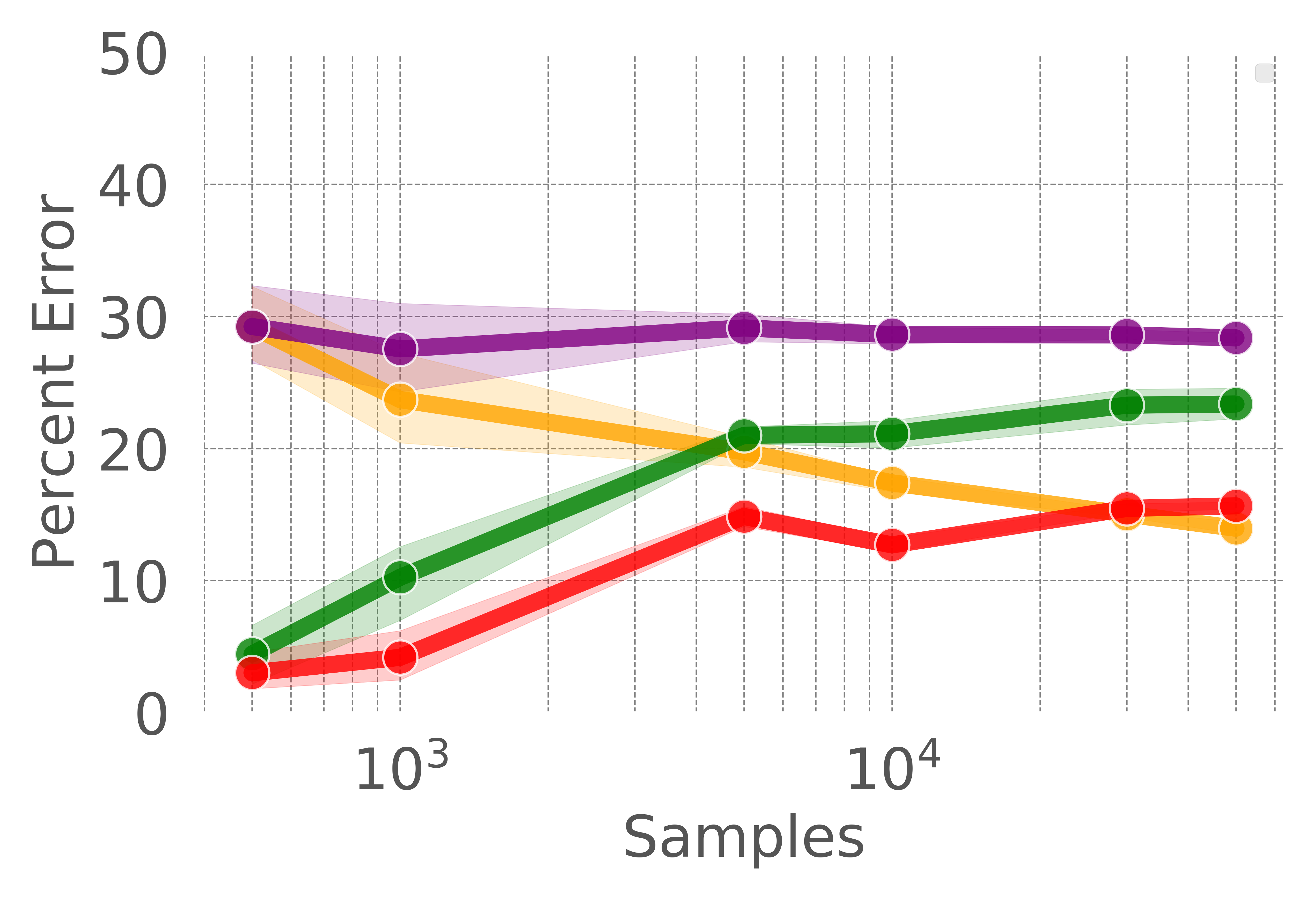} \\
&    \includegraphics[width=0.95\linewidth]{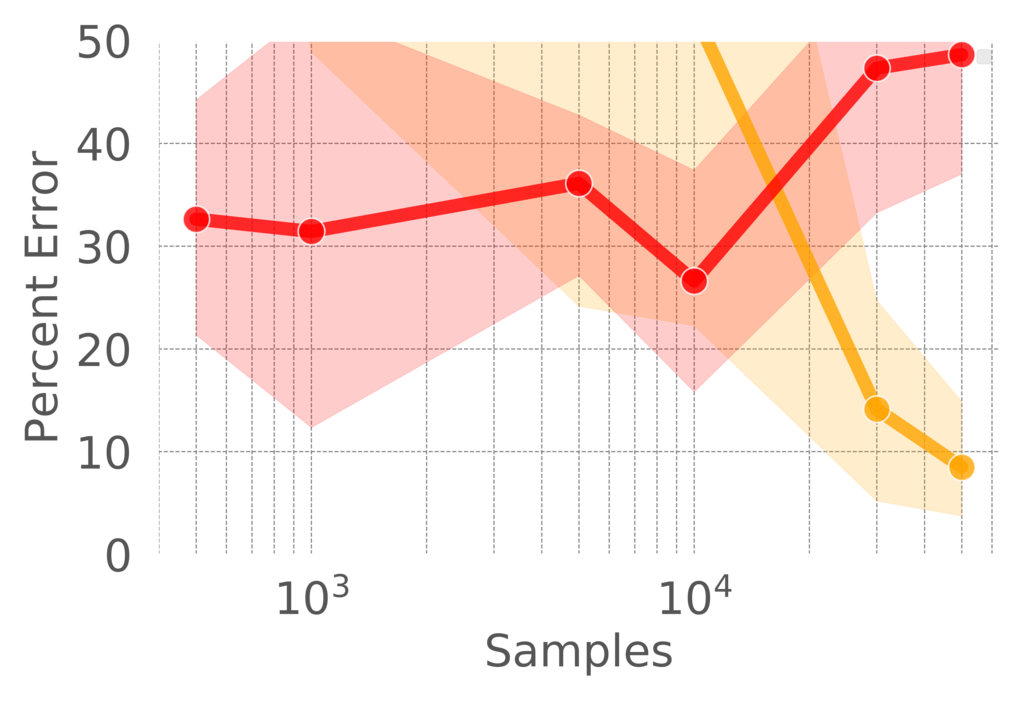} &
     \includegraphics[width=0.95\linewidth]{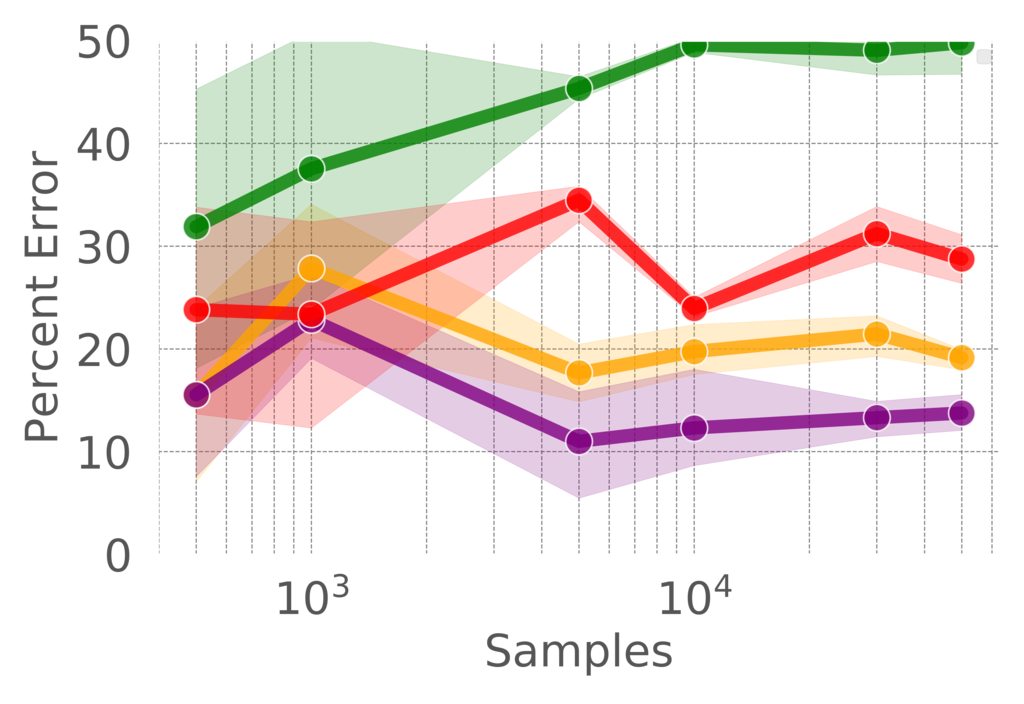} &
     \includegraphics[width=0.95\linewidth]{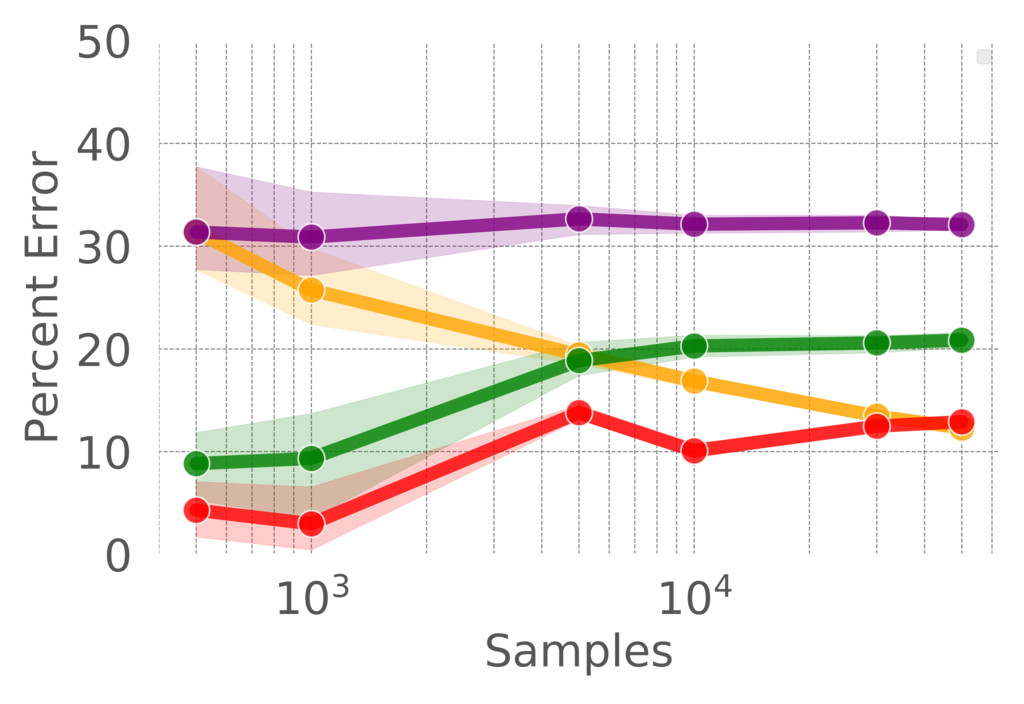} &
     \includegraphics[width=0.95\linewidth]{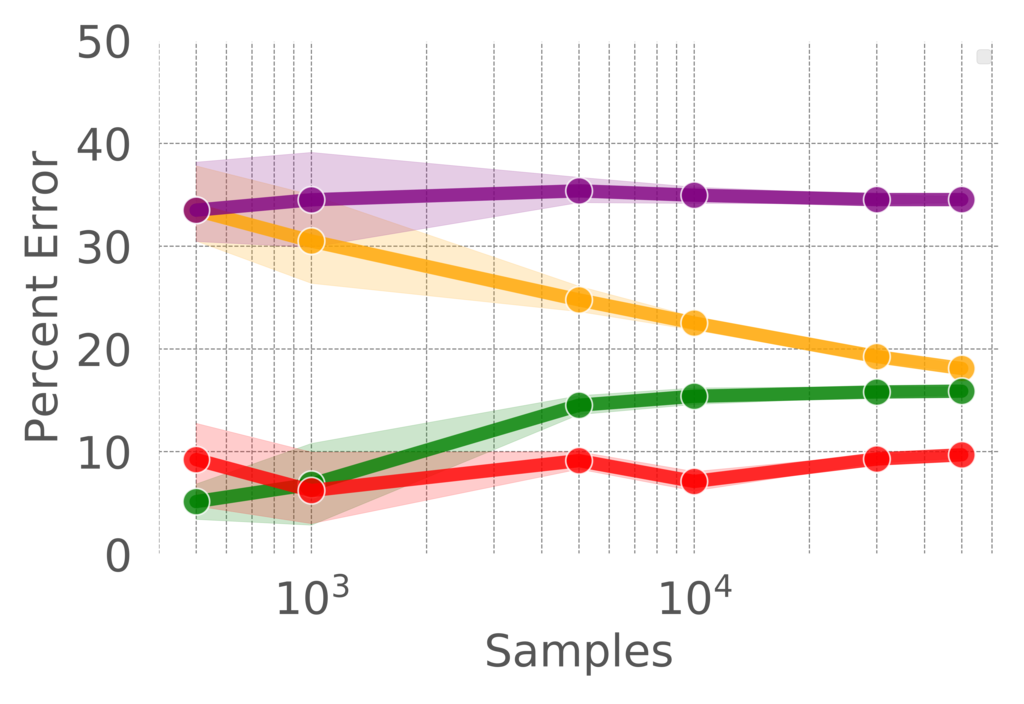} \\
&    \includegraphics[width=0.95\linewidth]{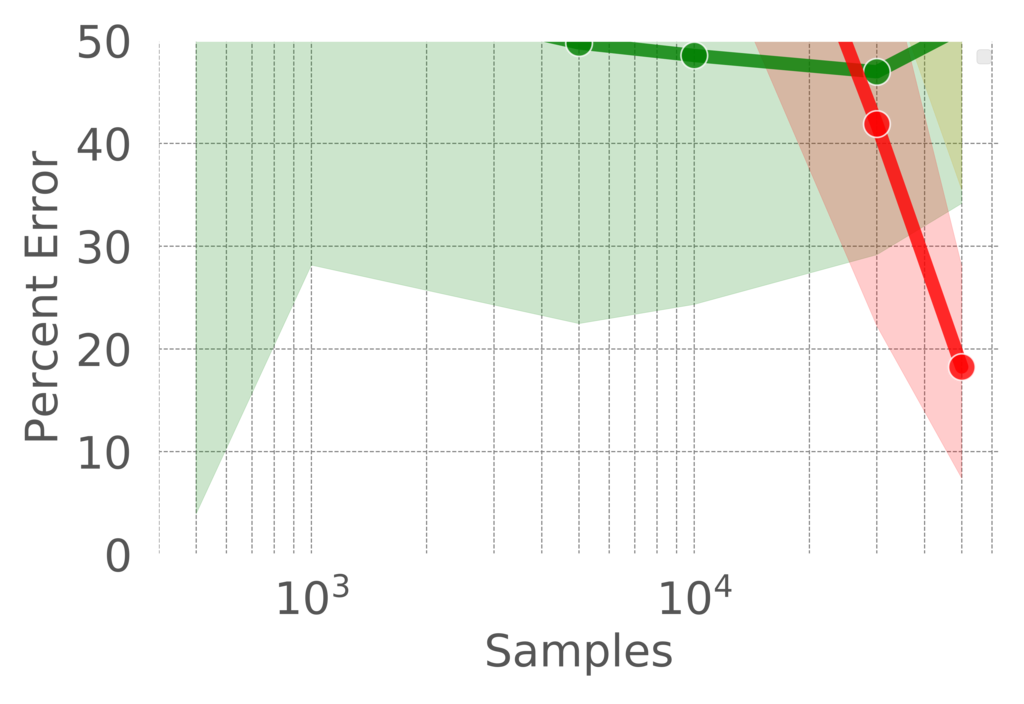} &
     \includegraphics[width=0.95\linewidth]{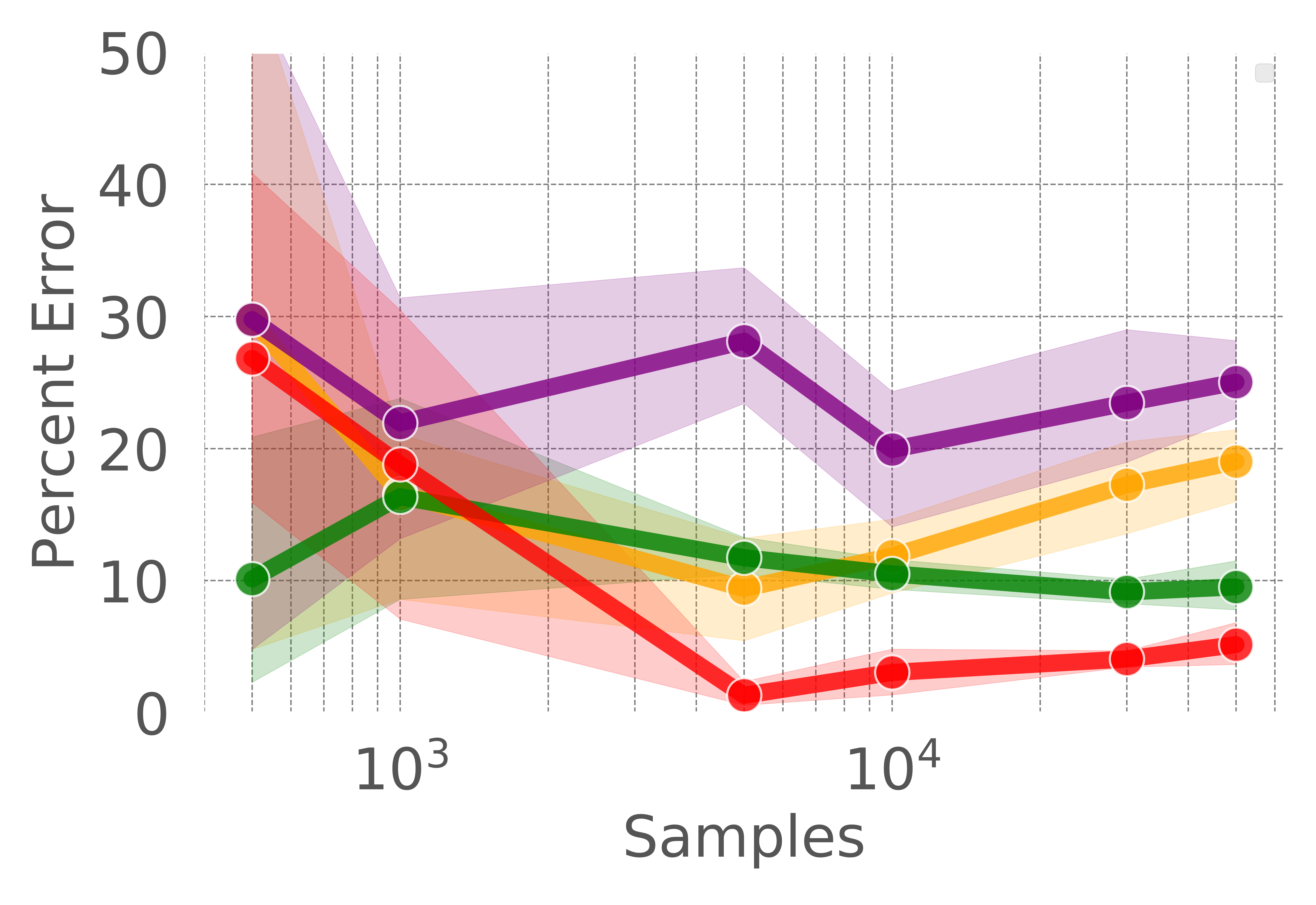} &
     \includegraphics[width=0.95\linewidth]{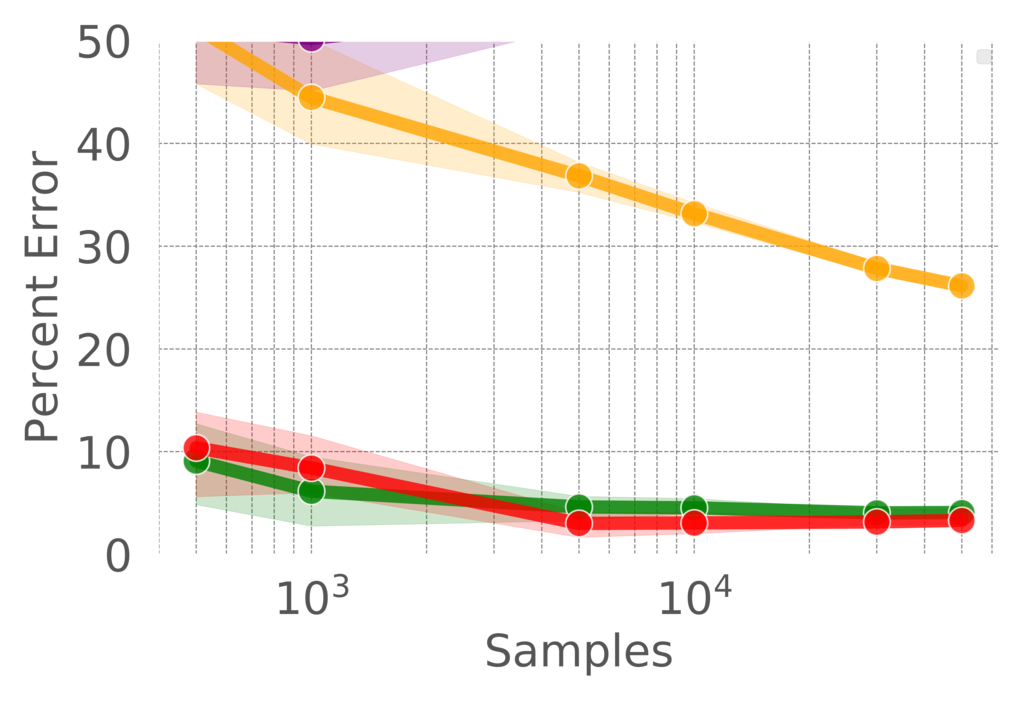} &
     \includegraphics[width=0.95\linewidth]{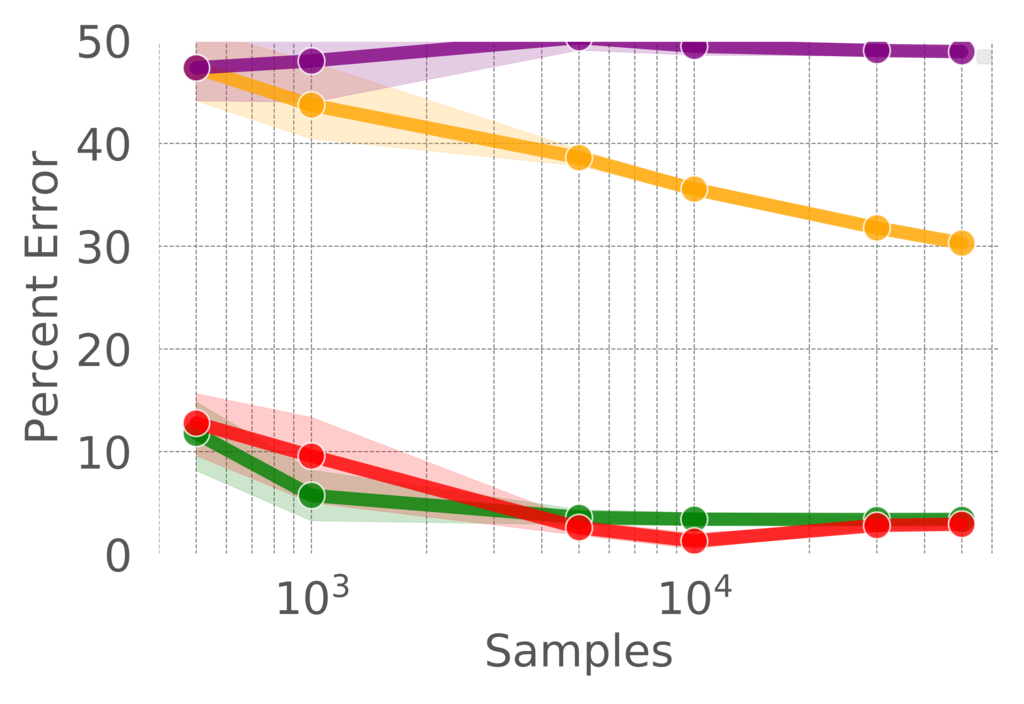}
    \end{tabular}
    \begin{tabular}{m{150mm}}
    \includegraphics[width=0.95\linewidth, trim={0 100mm 0 0}, clip]{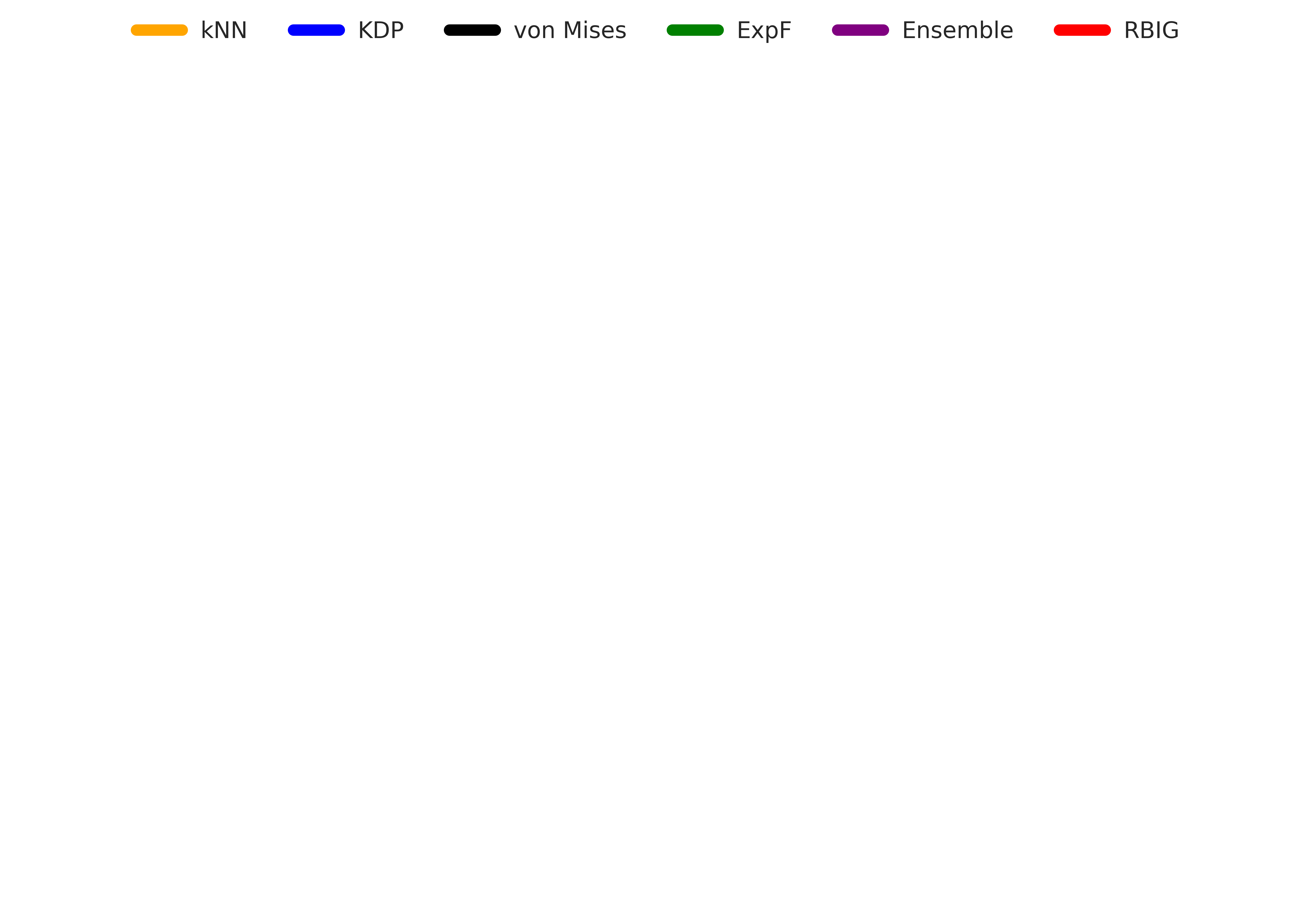}\\
\end{tabular}
    \caption{Total correlation estimation results in relative mean absolute error. Results for different distributions are given: Gaussian, uniform and the Student PDFs ($\nu = 3,5,20$ for each row respectively). Each column correspond to an experiment of a particular number of dimensions $D$. Mean and standard deviation are given for five trials.}
    \label{fig:TC_gauss}
\end{figure*}

We validate the performance of the proposed method on three different types of multivariate distributions:
\begin{itemize}
    \item {\em Gaussian distribution.} We consider data drawn from Gaussian distributions with zero mean and random covariance matrices. Each coefficient in the covariance matrix is generated randomly from a uniform ${\mathcal U}(0,1)$ distribution in each trial, the covariance matrix is enforced to be symmetric.
    \item {\em Linearly transformed uniform distribution.} Data is generated from a multidimensional uniform distribution and multiplied by a random squared matrix generated from a uniform ${\mathcal U}(0,1)$ distribution. The transformation matrix is different in each trial.
    \item {\em Multivariate Student distribution.} Different values of the degrees of freedom, $\nu$, that controls the weight of the tails are tested. The coefficients of the symmetric scale matrix (or shape matrix) are generated sampling from a uniform distribution. The diagonal is enforced to have a fixed value (in our case $10$) in order to keep the total correlation in a controlled regime.
\end{itemize}
Results for different number of samples and dimensions and comparison with the performance of different estimators are given in Fig. \ref{fig:TC_gauss}. Results are given in percentage of absolute error with regard to the analytic value of $T$. In Appendix \ref{app:formulas} we give details on how to compute these values analytically for each distribution. Table \ref{tab:TC} shows a summary of the relative mean absolute errors when using $10^4$ samples.

Results show that, while some methods behave well only in specific distributions, RBIG is consistent and always obtains a good performance for all distributions. 
Only when the distribution is Gaussian, RBIG is outperformed by \emph{expF}. This was expected since \emph{expF} assumes a Gaussian distribution. Being a non-parametric model, RBIG will generally provide more sensible estimates in real scenarios where the data generating mechanism is typically unknown. Importantly, note that the benefits of RBIG are more visible when the data dimensionality increases. This effect is visible also in the measures of the following sections.

\begin{table}[h!]
\begin{center}
\hspace{-0cm}
\caption{Relative mean absolute errors in percentage for total correlation estimation on known PDFs. Best value in dark gray, second best value in bright gray.}\label{tab:TC}
\begin{tabular}{|l|l|l|l|l|l|l|l|l|}
\hline
            &      & $D_x$ & \textbf{RBIG}                 & \textbf{kNN}                  & \textbf{KDP} & \textbf{expF}                 & \textbf{vME} & \textbf{Ens}  \\
\hline
\parbox[t]{3mm}{\multirow{4}{*}{\rotatebox[origin=c]{90}{Gaussian}}}   &   -   & 3            & \cellcolor[HTML]{C0C0C0}0.87  & 0.94                          & 76.65        & \cellcolor[HTML]{656565}0.63  & 4.27         & 4.03                          \\
  &              & 10           & \cellcolor[HTML]{C0C0C0}0.97  & 23.48                         & >100 & \cellcolor[HTML]{656565}0.27  & 31.72        & 34.83                         \\
  &              & 50           & \cellcolor[HTML]{C0C0C0}1.45  & 45.77                         & >100 & \cellcolor[HTML]{656565}0.52  & >100 & 54.74                         \\
  &              & 100          & \cellcolor[HTML]{C0C0C0}1.55  & 52.78                         & >100 & \cellcolor[HTML]{656565}0.41  & >100 & 59.94                         \\
\hline
\parbox[t]{2mm}{\multirow{4}{*}{\rotatebox[origin=c]{90}{Rotated}}}     &   -            & 3            & \cellcolor[HTML]{656565}1.70  & \cellcolor[HTML]{C0C0C0}1.80  & 82.90        & 16.80                         & 1.90         & 9.40                          \\
 &               & 10           & \cellcolor[HTML]{656565}8.30  & 27.20                         &  >100 & \cellcolor[HTML]{C0C0C0}11.00 & 24.20        & 38.70                         \\
 &               & 50           & \cellcolor[HTML]{656565}7.70  & 51.10                         &  >100 & \cellcolor[HTML]{C0C0C0}15.10 & >100 & 59.40                         \\
 &               & 100          & \cellcolor[HTML]{656565}7.50  & 57.80                         & >100 & \cellcolor[HTML]{C0C0C0}15.50 & >100 & 64.50                         \\
\hline
\parbox[t]{2mm}{\multirow{14}{*}{\rotatebox[origin=c]{90}{Student}}}   & \parbox[t]{2mm}{\multirow{4}{*}{\rotatebox[origin=c]{90}{$\nu=3$}}} & 3            & \cellcolor[HTML]{656565}7.01  & \cellcolor[HTML]{C0C0C0}13.55 &  >100 & 94.03                         & >100 & 66.59                         \\
&                  & 10           & 32.93                         & \cellcolor[HTML]{C0C0C0}16.73 &  >100 & 67.32                         & >100 & \cellcolor[HTML]{656565}15.27 \\
&                  & 50           & \cellcolor[HTML]{C0C0C0}18.18 & \cellcolor[HTML]{656565}12.02 &  >100 & 29.44                         & >100 & 24.65                         \\
&                  & 100          & \cellcolor[HTML]{656565}12.71 & \cellcolor[HTML]{C0C0C0}17.41 &  >100 & 21.12                         & >100 & 28.63                         \\
\cline{2-9}
&  \parbox[t]{2mm}{\multirow{4}{*}{\rotatebox[origin=c]{90}{$\nu=5$}}}    & 3            & \cellcolor[HTML]{656565}26.61 & \cellcolor[HTML]{C0C0C0}52.76 &  >100 & 89.74                         & 81.85        & 133.12                        \\
&                  & 10           & 23.94                         & \cellcolor[HTML]{C0C0C0}19.74 &  >100 & 49.60                         & >100 & \cellcolor[HTML]{656565}12.31 \\
&                  & 50           & \cellcolor[HTML]{656565}10.10 & \cellcolor[HTML]{C0C0C0}16.87 &  >100 & 20.29                         & >100 & 32.14                         \\
&                  & 100          & \cellcolor[HTML]{656565}7.10  & 22.53                         &  >100 & \cellcolor[HTML]{C0C0C0}15.39 & >100 & 34.96                         \\
\cline{2-9}
& \parbox[t]{2mm}{\multirow{4}{*}{\rotatebox[origin=c]{90}{$\nu=20$}}}   & 3            & \cellcolor[HTML]{C0C0C0}88.27 &  >100 & >100 & \cellcolor[HTML]{656565}48.56 & >100 & >100 \\
&                  & 10           & \cellcolor[HTML]{656565}3.05  & 11.86                         &  >100 & \cellcolor[HTML]{C0C0C0}10.51 & >100 & 19.93                         \\
&                  & 50           & \cellcolor[HTML]{656565}3.07  & 33.17                         &  >100 & \cellcolor[HTML]{C0C0C0}4.54  & >100 & 52.62                         \\
&                  & 100          & \cellcolor[HTML]{656565}1.31  & 35.56                         &  >100 & \cellcolor[HTML]{C0C0C0}3.43  & >100 & 49.46 \\
\hline
\end{tabular}
\end{center}
\end{table}

\subsubsection{Experiments on real-world data: visual neuroscience}
Here we use RBIG estimations of $T$ to measure the reduction of redundancy along the neural pathway in the visual brain.
The Efficient Coding Hypothesis argues that the retina-cortex pathway evolved to get an optimal representation of natural images~\cite{Barlow61,Barlow01}. Emergence of biological-like mechanisms from redundancy reduction optimization is the classical way to confirm this hypothesis~\cite{Olshausen96,Schwartz01,Malo06,Laparra15}.
However, an alternative confirmation relies on evaluating the statistical performance of biologically plausible networks that have not been optimized for information-theoretic goals~\cite{Malo10,GomezVilla19}.
Biological networks substantially reducing signal redundancy suggests that the hypothesis is correct.

In this example we compute the redundancy of real image data~\cite{VanHateren98} injected through physiologically sensible networks~\cite{Carandini12} tuned to fit visual psychophysics as in~\cite{Watson02,Malo10,LaparraJOSA17,GomezVilla19}.
RBIG is used to compute the redundancy (total correlation shared by the elements of the image signal) at the input of the biological network, and at the output. Substantial $\Delta T$ means that the biological network is efficient in encoding the considered signal.

This example is technically interesting because 
the RBIG estimate of $\Delta T$ can be compared with a reliable reference of $\Delta T$ computed via Eq.~\eqref{deltaT}.
In this visual neuroscience problem, the reference (or ground truth for comparison) can be obtained from the
analytic Jacobian of the transform applied by the biological network, i.e. $\nabla G(\vect{x})$ in Eq.~\eqref{deltaT}, which is known~\cite{Martinez18}.

\begin{figure*}[t!]
    \centering
    \hspace{0cm}\includegraphics[width=17cm]{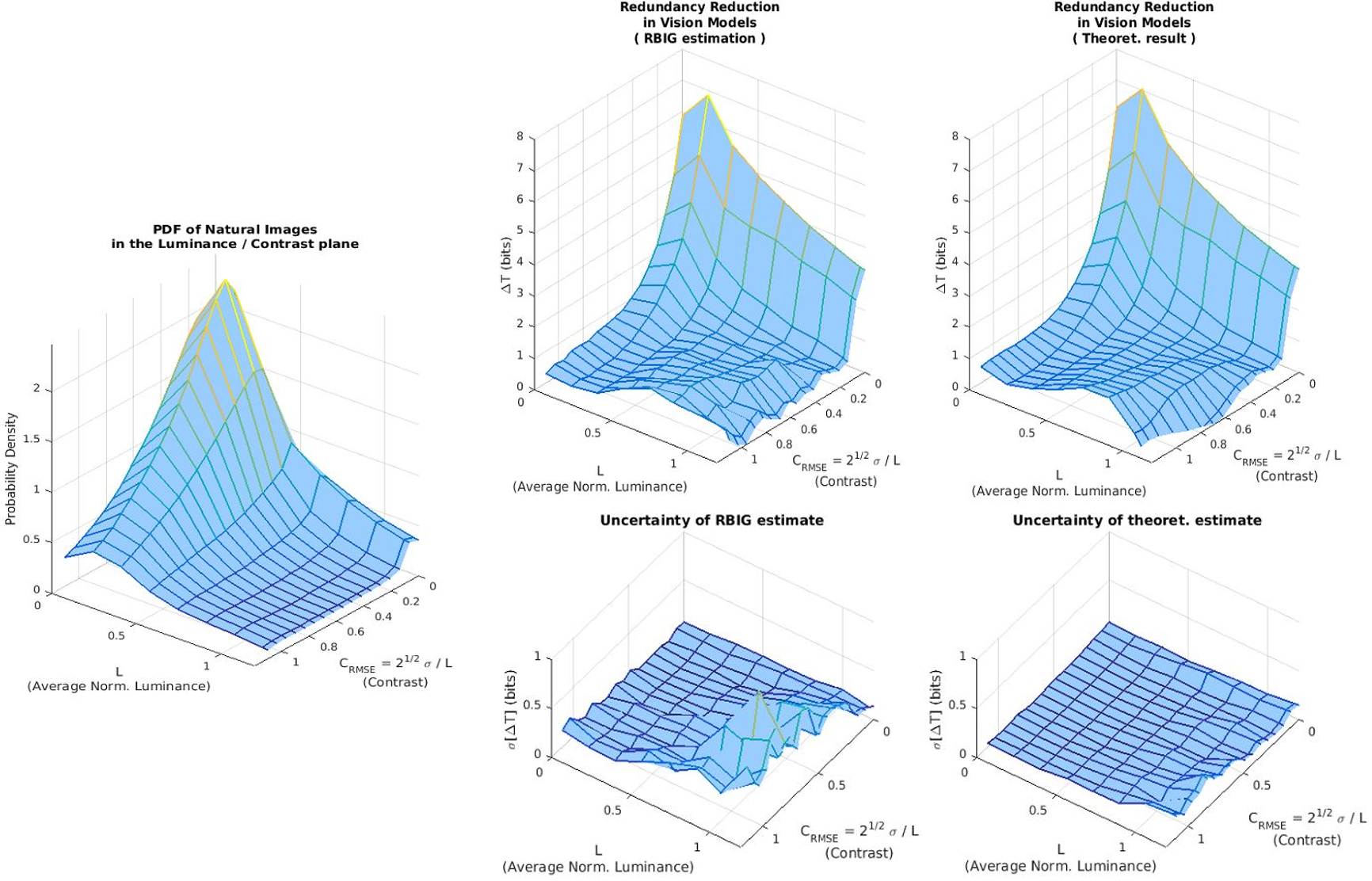}
    \caption{Efficient Coding Hypothesis in Visual Neuroscience from RBIG estimations of total correlation: redundancy reduction in the human visual system. Left panel shows the PDF of natural images (VanHateren database~\cite{VanHateren98}) at the luminance/contrast plane. Surfaces at the top-right show the redundancy reduction $\Delta T$ along the considered biological network (see~\cite{Carandini12,Martinez18,GomezVilla19} for background on these networks) at different points of the image space. Estimations are done with RBIG (center) and with the theoretical reference computed with the analytical Jacobian of the network~\cite{Martinez18} via Eq.~\eqref{deltaT} (right).
    This represents the efficiency of the visual brain in transmitting information about natural images.
    The surfaces at the bottom display the uncertainty of the RBIG and reference estimates computed from 10 realizations.}
    \label{fig:Fig_vision}
\end{figure*}

In our experiment, $1.2 \cdot 10^6$ image samples of 3-pixels were considered across the space of luminance and contrast. 
The responses of the biological model to these images were computed and the reduction of redundancy for stimuli from each region (or bin) across the image
space was estimated using (1) RBIG, and (2) the reference computed using the analytic Jacobian.
Results of $\Delta T$ 
is compared to the PDF of natural images in the luminance/contrast space in Fig.~\ref{fig:Fig_vision}. 

These results show that the efficiency of biological vision, $\Delta T$, is bigger in more populated regions of the image space, in line with the Efficient Coding Hypothesis. It is important to stress that this match between biological system and natural images holds regardless of the image database and regardless of minor differences in the biological model. Note that these results (obtained from the radiometrically calibrated VanHateren database~\cite{VanHateren98} and a narrow interaction kernel in the Divisive Normalization\footnote{Image data and code for vision model and analysis are available at \texttt{http://isp.uv.es/code/redundancy\_natural\_images.zip}}) are consistent with similar results in~\cite{GomezVilla19}, that used hyperspectral scenes~\cite{Foster16}; or results in~\cite{Malo19}, that used the colorimetrically calibrated IPL database~\cite{Laparra12}. In these cases, different kernels in Wilson-Cowan and 
Divisive Normalization vision models were used. 
The similarity of all these results indicates that the connection between neural efficiency and image statistics is pretty robust. From the technical point of view, the RBIG estimate using real image data closely follows the theoretical reference. Note that the deviations from the reference are small in general, and slightly bigger only in the region where the number of samples is small (the high-luminance / high-contrast region).

      \subsection{Entropy, \texorpdfstring{$H(\x)$}{H(x)}}
      \label{sec:entropy}
      
Entropy is a measure of how unpredictable a process is. The definition was given in Eq.~\eqref{eq:entropy} and illustrated in Fig.~\ref{fig:Fig_1}.

\subsubsection{Estimation using RBIG}
We build upon the straightforward estimation of the total correlation $T$ with RBIG in Eq.~\ref{Estim_T}, and the relation given in Eq.~\ref{eq:TC}, to define the RBIG-based estimator for $H$:
\begin{equation}
   \Tilde{H}(\x) = \sum_{i=1}^{D_x} \Tilde{H}(x_i) - \Tilde{T}(\x).
\end{equation}
Note that $\Tilde{T}$ only consists of marginal entropy estimations and the only extra computations in $\Tilde{H}(\x)$ are also marginal, that is $\Tilde{H}(x_i)$. 

\subsubsection{Validation on known PDFs}

Here we validate the performance of the proposed estimator in datasets with the same configuration described in \ref{sec:val_TC}.
See the details of the synthetic datasets in Appendix \ref{app:formulas}.
Table \ref{tab:Entropy} summarizes the bias (percentage of error) for a fixed number of samples.
Exhaustive results (bias and variance as a function of the kind of distribution, dimension and number of samples) are given in the supplementary material.

Results show again that RBIG estimation obtains good performance for all considered distributions.
The relative error 
is under $5\%$ in most of the cases and below $20\%$ in all cases.
In the Gaussian case, RBIG outperforms the other methods, and obtains similar accuracies to the estimator that assumes Gaussian distribution.
For the other distributions RBIG is the best one in most cases, and close to the first in the case it is not.
Again the performance of RBIG is clearly better in situations with multiple dimensions.

\begin{table}[b!]
\begin{centering}
\caption {Relative mean absolute errors in percentage for entropy estimation on known PDFs. Best value in dark gray, second best value in bright gray.}\label{tab:Entropy}
\begin{tabular}{|l|l|l|l|l|l|l|l|l|}
\hline
        &          & $D_x$  & \textbf{RBIG}                 & \textbf{kNN}                  & \textbf{KDP} & \textbf{expF}                 & \textbf{vME}                  & \textbf{Ens}            \\
\hline
\parbox[t]{3mm}{\multirow{4}{*}{\rotatebox[origin=c]{90}{Gaussian}}}   & & 3            & \cellcolor[HTML]{656565}0.90  & 1.70                          & 112.30       & \cellcolor[HTML]{C0C0C0}1.10  & 8.80                          & 12.00                        \\
  &                & 10           & \cellcolor[HTML]{C0C0C0}1.20  & 27.90                         & 179.80       & \cellcolor[HTML]{656565}0.10  & 34.70                         & 40.30                        \\
  &                & 50           & \cellcolor[HTML]{C0C0C0}0.90  & 32.20                         & 107.40       & \cellcolor[HTML]{656565}0.10  & 108.40                        & 38.10                        \\
  &                & 100          & \cellcolor[HTML]{C0C0C0}0.70  & 30.70                         & 89.60        & \cellcolor[HTML]{656565}0.10  & 94.20                         & 34.60                        \\
\hline
\parbox[t]{2mm}{\multirow{4}{*}{\rotatebox[origin=c]{90}{Rotated}}}     & & 3            & \cellcolor[HTML]{C0C0C0}4.00  & 6.20                          & 171.40       & 36.80                         & \cellcolor[HTML]{656565}3.70  & 30.10                        \\
  &                & 10           & \cellcolor[HTML]{656565}12.20 & 38.50                         & 241.80       & 17.90                         & \cellcolor[HTML]{C0C0C0}31.80 & 53.90                        \\
  &                & 50           & \cellcolor[HTML]{656565}6.80  & 44.60                         & 136.60       & \cellcolor[HTML]{C0C0C0}13.50 & 87.90                         & 51.60                        \\
  &                & 100          & \cellcolor[HTML]{656565}5.50  & 42.50                         & 110.70       & \cellcolor[HTML]{C0C0C0}11.50 & 94.30                         & 47.20                        \\
\hline
\parbox[t]{2mm}{\multirow{14}{*}{\rotatebox[origin=c]{90}{Student}}}   & \parbox[t]{2mm}{\multirow{4}{*}{\rotatebox[origin=c]{90}{$\nu=3$}}} & 3            & \cellcolor[HTML]{656565}0.75  & \cellcolor[HTML]{C0C0C0}0.76  & 34.93        & 11.90                         & 3.13                          & 1.99                         \\
  &                & 10           & 2.82                          & \cellcolor[HTML]{656565}1.44  & 137.19       & 15.55                         & 53.19                         & \cellcolor[HTML]{C0C0C0}1.77 \\
  &                & 50           & \cellcolor[HTML]{C0C0C0}6.03  & \cellcolor[HTML]{656565}3.47  & 195.30       & 22.11                         & 175.94                        & 7.09                         \\
  &                & 100          & \cellcolor[HTML]{656565}6.61  & \cellcolor[HTML]{C0C0C0}8.57  & 228.62       & 24.44                         & 166.09                        & 13.72                        \\
\cline{2-9}
&  \parbox[t]{2mm}{\multirow{4}{*}{\rotatebox[origin=c]{90}{$\nu=5$}}}    & 3          & \cellcolor[HTML]{656565}0.51  & \cellcolor[HTML]{C0C0C0}0.70  & 24.75        & 3.42                          & 1.38                          & 1.94                         \\
  &                & 10           & \cellcolor[HTML]{656565}1.12  & 1.25                          & 96.73        & 5.52                          & 59.29                         & \cellcolor[HTML]{C0C0C0}1.21 \\
  &                & 50           & \cellcolor[HTML]{656565}2.82  & \cellcolor[HTML]{C0C0C0}4.84  & 146.63       & 9.59                          & 202.48                        & 8.89                         \\
  &                & 100          & \cellcolor[HTML]{656565}3.11  & \cellcolor[HTML]{C0C0C0}10.67 & 184.02       & 11.36                         & 195.17                        & 16.24                        \\
\cline{2-9}
& \parbox[t]{2mm}{\multirow{4}{*}{\rotatebox[origin=c]{90}{$\nu=20$}}}   & 3            & \cellcolor[HTML]{656565}0.54  & \cellcolor[HTML]{C0C0C0}0.71  & 19.19        & 0.76                          & 1.32                          & 1.56                         \\
  &                & 10           & \cellcolor[HTML]{C0C0C0}0.48  & 0.84                          & 69.83        & 1.56                          & 46.62                         & \cellcolor[HTML]{656565}0.37 \\
  &                & 50           & \cellcolor[HTML]{656565}0.95  & 6.66                          & 107.74       & \cellcolor[HTML]{C0C0C0}3.31  & 219.86                        & 11.13                        \\
  &                & 100          & \cellcolor[HTML]{656565}0.68  & 13.45                         & 138.98       & \cellcolor[HTML]{C0C0C0}4.20  & 214.41                        & 19.35                       \\
\hline
\end{tabular}
\end{centering}
\end{table}

\subsubsection{Experiments on real-world data: geosciences}
Entropy computed with RBIG has been already applied in geoscience and remote sensing for Earth observation. For instance, in \cite{Laparra2015} entropy was used to analyze which is the optimal configuration of spectral and spatial resolution for an hyperspectral image sensor. In \cite{Johnson18EGU} 
entropy was used to analyze effect of using different spatio-temporal configurations for 
studying global essential climate variables related to the water cycle, like evaporation, precipitation and soil moisture. 

Here we show a novel example were we analyze the spatial patterns of air temperature globally. 
This example illustrates the kind of interesting regularities that may be discovered by the RBIG-based entropy measures. 
We use the data from the Earth Science Data Lab (ESDL) \footnote{\url{https://www.earthsystemdatalab.net/}}\cite{Mahecha19esdc}. These data consist of 3-dimensional cubes (2 spatial and one temporal dimensions) for different variables. In our case we select the air temperature at 2 m (ERAInterim) as a variable for the years 2001-2008 and the whole globe. For each temporal acquisition we have a temperature product with $1440 \times 720$ spatial samples (in a 0.25 degrees grid resolution) weekly sampled (46 images per year).

\begin{figure*}
    \centering
        \begin{tabular}{m{55mm}m{55mm}m{55mm}}
        \includegraphics[width=\linewidth]{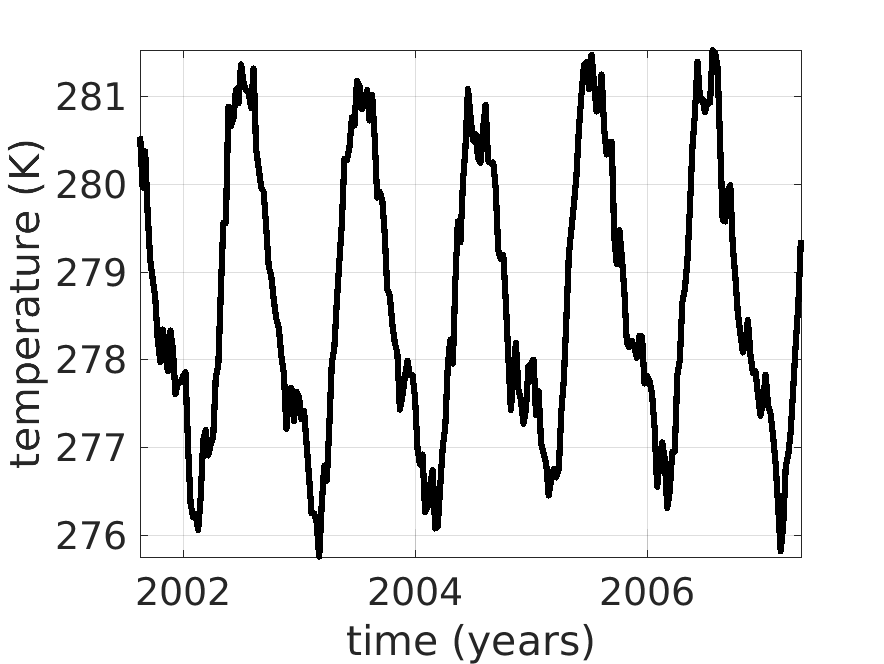} &
        \includegraphics[width=\linewidth]{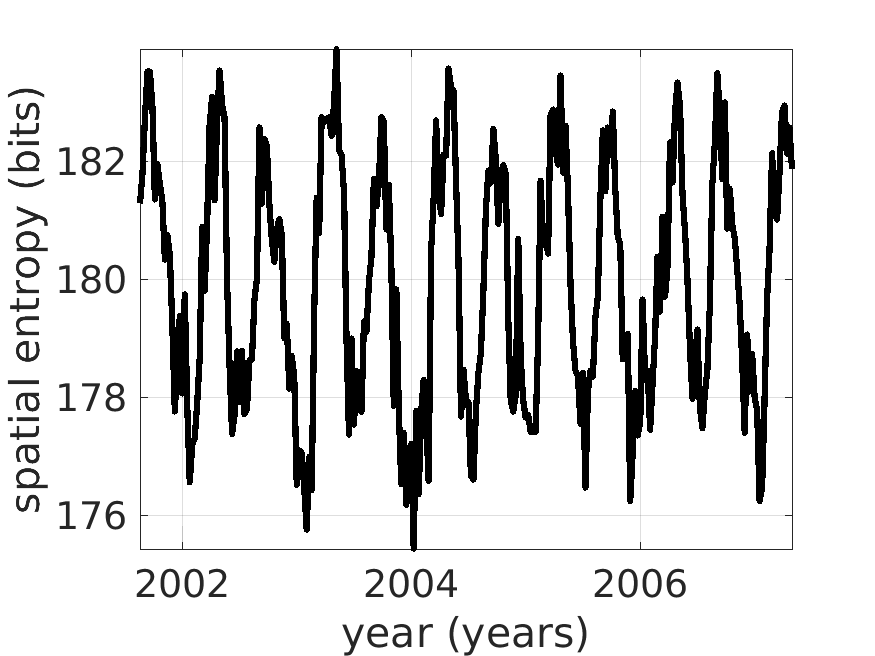} &
        \includegraphics[width=\linewidth]{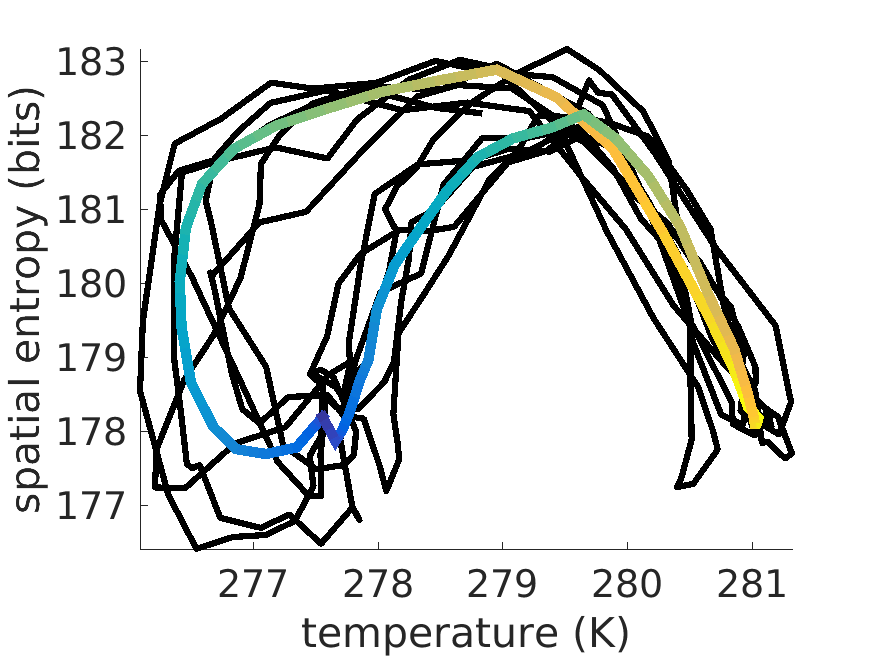}
        \end{tabular}
    \caption{Regularities in geoscience data from RBIG estimations of entropy. Left: mean global temperature at each time step. Center: Spatial entropy at each time step (see text for details). Right: Evolution of the relation between temperature and spatial entropy for 2001-2008. The mean for all the studied years is given in the thicker line which colors correspond to the distance between the Earth and the sun for this period of time (blue closer, yellow farther away).}\label{fig:Real_entropy}
\end{figure*}

Figure \ref{fig:Real_entropy} shows the results of computing the entropy for patches of $5 \times 5$ spatial regions for each temporal acquisition. Results for the mean temperature of the whole globe are given for reference. It can be seen that the seasonal cycle is clear for temperature (left) and entropy (middle). However, while temperature displays one cycle per year as expected, the spatial entropy displays two cycles per year. 
The figure at the right shows the relations between the mean temperature and the spatial entropy showing a cyclic pattern.
This means that variability in the spatial patterns of the temperature have a different behavior than the mean temperature.
The spatial patterns are less variable during winter and summer, and are more variable in spring and autumn.
This kind of analysis can help to understand the spatial behavior of terrestrial ecosystems for example, and could help in characterizing their variability and dynamics.

      \subsection{Kullback-Leibler Divergence, \texorpdfstring{${\textrm{KL}}(\x|\y)$ }{KL(x|y)} }
      \label{sec:kld}

While $D_{\textrm{KL}}$ is a measure of divergence between PDFs, here we use the simpler notation $D_{\textrm{KL}}(\x|\y) = D_{\textrm{KL}}(p(\x)|p(\y))$, meaning that the considered variables follow the corresponding PDFs. 

\subsubsection{Estimation using RBIG}
\begin{figure}
    \centering
    \begin{tabular}{c}
    \includegraphics[width=8cm]{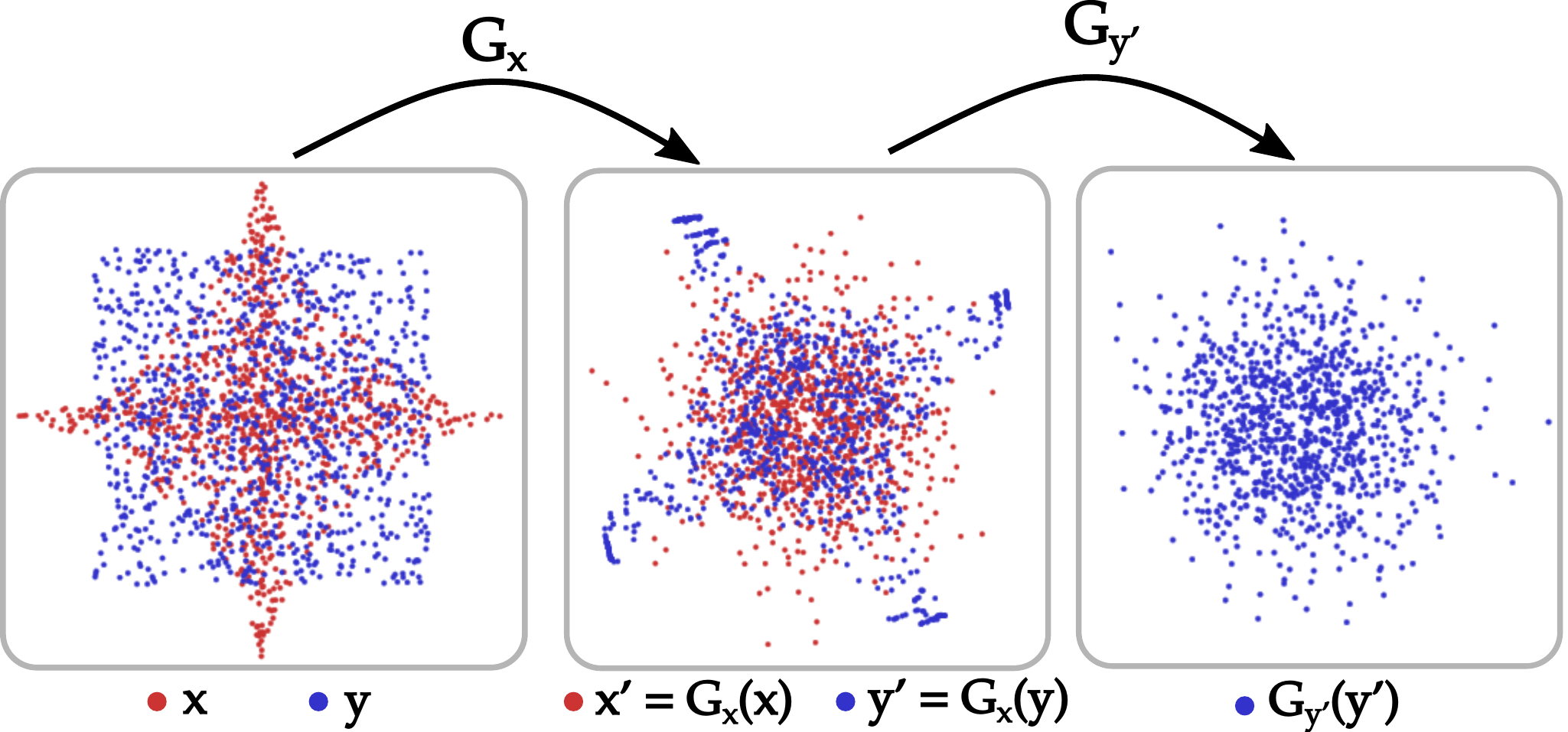} \end{tabular}
    \caption{Illustrative example of computing the $D_{\textrm{KL}}(\x|\y)$ using RBIG. First, both datasets are transformed using the transformation that Gaussianizes the dataset $\x$, i.e. $\x' = G_x(\x) \sim \mathcal{N}(\mathbf{0},\mathbf{I})$ and $\y' = G_x(\y)$. And secondly we compute how far is the new dataset $\y^*$ of being Gaussian. In order to do the second step we need to compute the total correlation of $\y'$. We do so using the equation \eqref{eq:TC_from_RBIG} which requires to compute the transformation that Gaussianizes $\y'$, i.e. $G_{y^*}$.}
    \label{fig:KLD_ilustrative}
\end{figure}

The proposed estimator is based on the property of KLD that states that the $D_{\textrm{KL}}$ is invariant under invertible and differentiable mappings~\cite{Qiao10}, i.e. $D_{\textrm{KL}}(\x|\y) = D_{\textrm{KL}}(f(\x)|f(\y))$ 
if the function $f(\cdot)$ is invertible and differentiable. 
As a result, we propose transforming both datasets, $\x$ and $\y$, to a domain where we know that one of them is Gaussian,
and then compute how non-Gaussian the other dataset is. Figure \ref{fig:KLD_ilustrative} illustrates this two-steps idea.

In the above context, we use the concept of \emph{Non-Gaussianity}~\cite{Cardoso03}.
Non-Gaussianity is defined as the divergence of a particular distribution with regard to a Gaussian distribution
with the same mean and covariance $J(\x) = D_{\textrm{KL}}(p(\x)|\mathcal{N}(\x,\bf{\mu},\Sigma))$.
Here we define the \emph{standard non-Gaussianity}, $J_s(\x)$ as the divergence between $p(\x)$ and a standard Gaussian, $J_s(\x) = D_{\textrm{KL}}(p(\x)|\mathcal{N}(\x,\mathbf{0},\mathbf{I}))$. Now, by using the Pythagorean property of the
divergence~\cite{Cardoso03}, we can write the standard non-Gaussianity estimate as:
\begin{equation}
     J_s(\x) = D_{\textrm{KL}}\Bigg( p(\x)\Bigg| \prod_{i=1}^{D_x} p(x_i) \Bigg) + D_{\textrm{KL}}\Bigg(\prod_{i=1}^{D_x} p(x_i)\Bigg| \prod_{i=1}^{D_x} \mathcal{N}(0,1) \Bigg), \nonumber
\end{equation}
and, since the first term is the definition of total correlation~\cite{Studeny98}, and the second term is the divergence of factorized PDFs, the standard non-Gaussianity can be readily written as: 
\begin{equation}
J_s(\x) = T(\x) + \sum_{i=1}^{D_x} D_{\textrm{KL}}(p_{x_i}(x_i)|\mathcal{N}(0,1)).
\label{eq:nongaussianity}
\end{equation}

This leads to propose the following KLD estimator. 

\noindent{\bf Proposition:} Taking the RBIG transform $G_x(\cdot)$ learned from samples, $\x$, the RBIG-based estimator of divergence $D_{\textrm{KL}}(\x|\y)$ can be computed as:
\begin{eqnarray}
     \hspace{-1cm} \Tilde{D}_{\textrm{KL}}(\y|\x) &=& J_s(G_x(\y))
     \label{estim_kld_yx}
\end{eqnarray}
where, following eq.~\eqref{eq:nongaussianity}, $J_s(\cdot)$ depends on an RBIG-based estimation of $T$, and a set of marginal divergences.

\noindent{\bf Proof:} Consider the transformation $G_x$ obtained training RBIG using the data $\x$. Invoking the invariance of $D_{\textrm{KL}}(\y|\x)$ under invertible and differentiable transforms~\cite{Qiao10}, and considering that the RBIG transforms are the invertible and differentiable~\cite{Laparra2011},
we can write:
\begin{eqnarray}
      D_{\textrm{KL}}(\y|\x) &=& D_{\textrm{KL}}(G_x(\y)|G_x(\x)). \label{KLD_J}
\end{eqnarray}
Taking into account that when applying the transformation $G_x$ over $\x$ the obtained dataset follows a standard Gaussian, i.e. $G_x(\x) \sim \mathcal{N}(\mathbf{0},\mathbf{I})$, then:
\begin{eqnarray}
      D_{\textrm{KL}}(\y|\x) &=& D_{\textrm{KL}}(G_x(\y)|\mathcal{N}(\mathbf{0},\mathbf{I})) = J_s(G_x(\y)).\nonumber
\end{eqnarray}

\subsubsection{Validation on known PDFs}
\label{sec:kld_validation}
Here we evaluate the $D_{\textrm{KL}}$ among two datasets $\x$ and $\y$. Four different combinations of different datasets are analyzed:
\begin{itemize}
    \item {\em Gaussians with different means.} Both datasets are drawn from Gaussian distributions with the same diagonal covariance ($\Sigma_1 = \Sigma_2 = \mathbf{I}$) but different means ($\mu_1=\mathbf{0}$, $\mu_2$ may take different values $\mu_2=[0.2,0.4,0.6]$.

\item {\em Gaussians with different covariance matrices.} The data are also generated from Gaussians but in this case with the same mean (i.e. $\mu_1 = \mu_2 = \mathbf{0}$) but different covariances (i.e. $\Sigma_1 \neq \Sigma_2$). We parametrized $\Sigma_2 = \sigma_2 \mathbf{Q} + \mathbf{I}$ and tested three different values $\sigma_2=[0.5,0.75,0.9]$, where $\mathbf{Q}$ is a covariance matrix generated randomly in each trial and with zeros in the diagonal.

\item {\em Gaussian vs Student.} One dataset is generated from a Gaussian distribution and the other dataset from a multivariate Student distribution. We keep fixed the parameters of the Gaussian distribution while modifying the $\nu$ parameter of the multivariate Student. We parametrized the Gaussian distribution using a multivariate Student with $\nu_1=100$. In particular $\Sigma_1 = \Sigma_2 = \mathbf{I}$, $\mu_1=\mu_2=\mathbf{0}$, and $\nu_2 = [2,4,7]$.

\item {\em Student vs Student.} Data is generated from two multivariate Student distributions.  We keep fixed the parameters of one distribution while modifying the $\nu$ parameter of the other one. In particular $\Sigma_1 = \Sigma_2 = \mathbf{I}$, $\mu_1=\mu_2=\mathbf{0}$, and $\nu_1=8$, and $\nu_2 = [2,4,7]$.
\end{itemize}
Details of how to compute analytically the $D_{\textrm{KL}}$ value for each configuration are given in Appendix~\ref{app:formulas}.
Table~\ref{tab:KLD} summarizes the results for a representative sample size. 
In this case, divergence is not directly available from the {\bf KDP} and {\bf Ens} estimators, so they are not included in the comparison.
Bias/variance results for different number of samples are given in the supplementary material. 
Results show again that RBIG estimation obtains good performance.
BRIG is second best only in the all-Gaussian cases where, not surprisingly, it is outperformed by \emph{expF}, the method that assumes Gaussian models.
When different distributions come in, RBIG estimations are clearly better.
Actually, in this non-Gaussian case, while all other methods yield huge errors, i.e. the relative error bigger than $1000\%$ in particular sample/dimension configurations, RBIG bias never diverges.

\begin{table}[t!]
\caption {Relative mean absolute errors in percentage for $D_{\textrm{KL}}$ estimation on known PDFs. Best value in dark gray, second best value in bright gray.}\label{tab:KLD}
\begin{center}
\begin{tabular}{|l|l|l|l|l|l|l|}
\hline
   &  & \textbf{dim} & \textbf{RBIG}    & \textbf{kNN}         & \textbf{expF}       & \textbf{vME}      \\
\hline

\parbox[t]{3mm}{\multirow{12}{*}{\rotatebox[origin=c]{90}{Gaussian, different means}}} & \parbox[t]{2mm}{\multirow{4}{*}{\rotatebox[origin=c]{90}{$\mu_2=0.2$}}}  & 3    & \cellcolor[HTML]{C0C0C0}13.49  & 16.90  & \cellcolor[HTML]{656565}10.28  & 92.28      \\
&                                                                        & 10           & \cellcolor[HTML]{C0C0C0}19.50  & 22.47                           & \cellcolor[HTML]{656565}3.27   &  >1000                        \\
&                                                                       & 50           & \cellcolor[HTML]{C0C0C0}32.04  & 40.85                           & \cellcolor[HTML]{656565}13.34  &  >1000                        \\
&                                                                       & 100          & 47.40                          & \cellcolor[HTML]{C0C0C0}41.24   & \cellcolor[HTML]{656565}28.14  &  >1000                        \\
\cline{2-7}
&\parbox[t]{2mm}{\multirow{4}{*}{\rotatebox[origin=c]{90}{$\mu_2=0.4$}}}      & 3            & \cellcolor[HTML]{656565}3.55   & 7.98         & \cellcolor[HTML]{C0C0C0}5.57   & 24.22                         \\
   &                                                                    & 10           & \cellcolor[HTML]{C0C0C0}5.25   & 22.93                           & \cellcolor[HTML]{656565}2.02   & 604.17                        \\
   &                                                                    & 50           & \cellcolor[HTML]{C0C0C0}8.67   & 40.91                           & \cellcolor[HTML]{656565}3.40   &  >1000                        \\
   &                                                                    & 100          & \cellcolor[HTML]{656565}4.83   & 43.49                           & \cellcolor[HTML]{C0C0C0}8.70   &  >1000                        \\
\cline{2-7}
&\parbox[t]{2mm}{\multirow{4}{*}{\rotatebox[origin=c]{90}{$\mu_2=0.6$}}}                                                  & 3            & \cellcolor[HTML]{656565}3.75   & 6.39                            & \cellcolor[HTML]{C0C0C0}3.89   & 12.23                         \\
   &                                                                    & 10           & \cellcolor[HTML]{C0C0C0}2.81   & 24.72                           & \cellcolor[HTML]{656565}1.85   & 213.72                        \\
   &                                                                    & 50           & \cellcolor[HTML]{C0C0C0}13.83  & 43.11                           & \cellcolor[HTML]{656565}1.83   & 897.65                        \\
   &                                                                    & 100          & \cellcolor[HTML]{C0C0C0}42.42  & 46.00                           & \cellcolor[HTML]{656565}5.11   & 686.96                        \\
\hline

\parbox[t]{3mm}{\multirow{12}{*}{\rotatebox[origin=c]{90}{Gaussians, different covs.}}} & \parbox[t]{2mm}{\multirow{4}{*}{\rotatebox[origin=c]{90}{$\mu_2=0.5$}}}  & 3    & \cellcolor[HTML]{C0C0C0}24.93  & 27.30         & \cellcolor[HTML]{656565}4.89   & 63.90                         \\
&                                                                       & 10           & \cellcolor[HTML]{C0C0C0}18.80  & 103.65                          & \cellcolor[HTML]{656565}2.64   &  >1000                        \\
&                                                                       & 50           & \cellcolor[HTML]{C0C0C0}23.62  & 173.62                          & \cellcolor[HTML]{656565}8.42   &  >1000                       \\
&                                                                       & 100          & \cellcolor[HTML]{C0C0C0}32.56  & 200.33                          & \cellcolor[HTML]{656565}17.59  &  >1000                        \\
\cline{2-7}
&\parbox[t]{2mm}{\multirow{4}{*}{\rotatebox[origin=c]{90}{$\mu_2=0.75$}}}                                             & 3            & \cellcolor[HTML]{C0C0C0}21.04  & 24.77                           & \cellcolor[HTML]{656565}3.72   & 36.64       \\
 &                                                                      & 10           & \cellcolor[HTML]{C0C0C0}10.44  & 96.85                           & \cellcolor[HTML]{656565}1.86   & 605.00                        \\
 &                                                                      & 50           & \cellcolor[HTML]{C0C0C0}10.07  & 159.16                          & \cellcolor[HTML]{656565}5.70   &  >1000                        \\
 &                                                                      & 100          & \cellcolor[HTML]{C0C0C0}13.66  & 179.67                          & \cellcolor[HTML]{656565}11.40  &  >1000                        \\
\cline{2-7}
&\parbox[t]{2mm}{\multirow{4}{*}{\rotatebox[origin=c]{90}{$\mu_2=0.9$}}}                                                & 3            & \cellcolor[HTML]{C0C0C0}17.12  & 25.95                           & \cellcolor[HTML]{656565}3.40   & 26.15         \\
  &                                                                     & 10           & \cellcolor[HTML]{C0C0C0}6.77   & 94.42                           & \cellcolor[HTML]{656565}1.60   & 448.87                        \\
  &                                                                     & 50           & \cellcolor[HTML]{656565}3.40   & 152.46                          & \cellcolor[HTML]{C0C0C0}4.81   &  >1000                        \\
  &                                                                     & 100          & \cellcolor[HTML]{656565}5.96   & 170.28                          & \cellcolor[HTML]{C0C0C0}9.43   &  >1000 \\
\hline

\parbox[t]{3mm}{\multirow{12}{*}{\rotatebox[origin=c]{90}{Gaussian vs. Student}}} & \parbox[t]{2mm}{\multirow{4}{*}{\rotatebox[origin=c]{90}{$\nu=2$}}}  & 3            & \cellcolor[HTML]{656565}5.08   & 29.58                           & 793.53                         & \cellcolor[HTML]{C0C0C0}5.78  \\
&                                                              & 10           & \cellcolor[HTML]{656565}32.72  & \cellcolor[HTML]{C0C0C0}83.51   & 1278.91                        & 596.63                        \\
&                                                              & 50           & \cellcolor[HTML]{656565}59.37  & \cellcolor[HTML]{C0C0C0}468.33  & 2783.43                        & >1000                            \\
&                                                              & 100          & \cellcolor[HTML]{656565}42.11  & \cellcolor[HTML]{C0C0C0}1024.30 & 4330.18                        & >1000                            \\
\cline{2-7}
&\parbox[t]{2mm}{\multirow{4}{*}{\rotatebox[origin=c]{90}{$\nu=4$}}}                                                & 3             & \cellcolor[HTML]{656565}17.09  & 95.02                           & 148.08                         & \cellcolor[HTML]{C0C0C0}22.52 \\
&                                                              & 10           & \cellcolor[HTML]{656565}42.84  & \cellcolor[HTML]{C0C0C0}157.63  & 219.26                         & 963.37                        \\
&                                                              & 50           & \cellcolor[HTML]{656565}60.53  & 584.46                          & \cellcolor[HTML]{C0C0C0}547.48 & >1000                            \\
&                                                              & 100          & \cellcolor[HTML]{656565}41.71  & 1214.61                         & \cellcolor[HTML]{C0C0C0}962.45 & >1000                            \\
\cline{2-7}
&\parbox[t]{2mm}{\multirow{4}{*}{\rotatebox[origin=c]{90}{$\nu=7$}}}                                                & 3             & \cellcolor[HTML]{656565}8.34   & 271.61                          & \cellcolor[HTML]{C0C0C0}35.78  & 59.69                         \\
&                                                              & 10           & \cellcolor[HTML]{656565}38.78  & 307.82                          & \cellcolor[HTML]{C0C0C0}49.77  &  >1000                        \\
&                                                              & 50           & \cellcolor[HTML]{656565}48.80  & 713.36                          & \cellcolor[HTML]{C0C0C0}145.15 &  >1000                        \\
&                                                              & 100          & \cellcolor[HTML]{656565}26.01  & 1399.34                         & \cellcolor[HTML]{C0C0C0}278.93 & >1000                            \\

\hline
\parbox[t]{3mm}{\multirow{12}{*}{\rotatebox[origin=c]{90}{Student vs. Student}}} & \parbox[t]{2mm}{\multirow{4}{*}{\rotatebox[origin=c]{90}{$\nu=2$}}}  & 3            & \cellcolor[HTML]{656565}9.08   & \cellcolor[HTML]{C0C0C0}13.87   & 3442.45                        &  >1000                        \\
 &                                                             & 10           & \cellcolor[HTML]{656565}20.57  & \cellcolor[HTML]{C0C0C0}57.60   & 7462.58                        & 346.61                        \\
 &                                                             & 50           & \cellcolor[HTML]{656565}85.14  & \cellcolor[HTML]{C0C0C0}405.47  & 19991.36                       & >1000                            \\
 &                                                             & 100          & \cellcolor[HTML]{656565}242.80 & \cellcolor[HTML]{C0C0C0}939.24  & 35064.60                       & >1000                            \\
\cline{2-7}
 &\parbox[t]{2mm}{\multirow{4}{*}{\rotatebox[origin=c]{90}{$\nu=4$}}}                                                & 3            & \cellcolor[HTML]{656565}9.51   & \cellcolor[HTML]{C0C0C0}47.03   & 1502.19                        & 48.89                         \\
 &                                                             & 10           & \cellcolor[HTML]{656565}36.33  & \cellcolor[HTML]{C0C0C0}139.12  & 2561.86                        &  >1000                        \\
 &                                                             & 50           & \cellcolor[HTML]{656565}37.29  & \cellcolor[HTML]{C0C0C0}656.95  & 7997.12                        & >1000                            \\
 &                                                             & 100          & \cellcolor[HTML]{656565}60.52  & \cellcolor[HTML]{C0C0C0}1441.18 & 13033.03                       & >1000                            \\
\cline{2-7}
&\parbox[t]{2mm}{\multirow{4}{*}{\rotatebox[origin=c]{90}{$\nu=7$}}}                                                & 3             & \cellcolor[HTML]{656565}13.13  & \cellcolor[HTML]{C0C0C0}126.41  & 589.47                         & 128.84                        \\
 &                                                              & 10           & \cellcolor[HTML]{656565}23.13  & \cellcolor[HTML]{C0C0C0}301.97  & 1070.70                        & >1000                        \\
 &                                                             & 50           & \cellcolor[HTML]{656565}28.34  & \cellcolor[HTML]{C0C0C0}976.95  & 3689.57                        & >1000                            \\
 &                                                                      & 100          & \cellcolor[HTML]{656565}145.88 & \cellcolor[HTML]{C0C0C0}2046.95 & 6370.43                        &  >10000                            \\
\hline
\end{tabular}
\end{center}
\end{table}

\subsubsection{Experiments on real-world data: computer vision}
Here we show how RBIG-based estimations of $D_{\textrm{KL}}$ may be useful to analyze image databases widely used in Computer Vision: MNIST~\cite{Lecun98} and CIFAR-10~\cite{Krizhevsky09}.
Divergence between classes may describe the difficulty/simplicity of class separation in image recognition problems, and can help designing automatic classification schemes. 

\begin{figure}
    \centering
    \begin{tabular}{c}
    \includegraphics[width=8cm, trim=42.5mm 0 0 0]{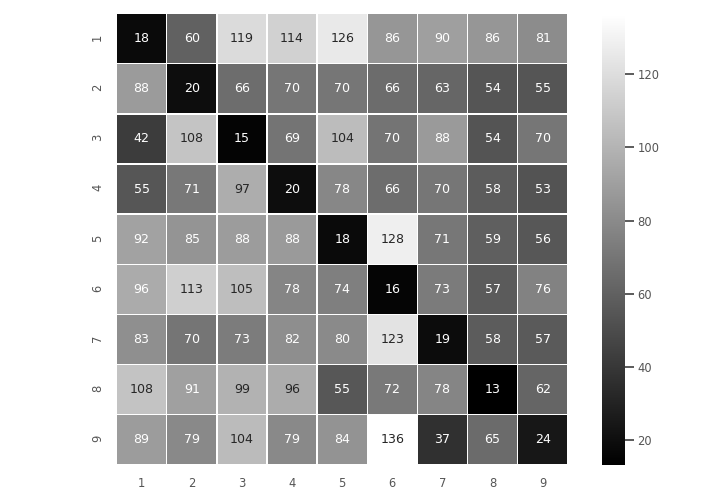} \\
    \includegraphics[width=9cm, trim=50mm 0 0 0]{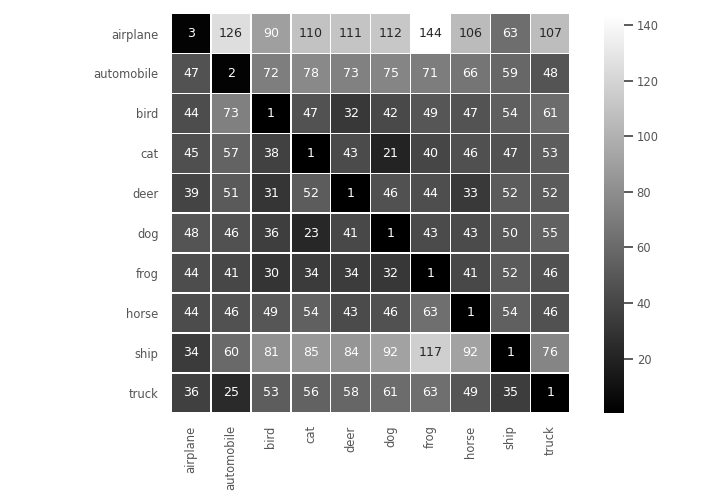} \\
    \end{tabular}
    \caption{Analysis of computer vision datasets using RBIG estimations of divergence.
    Heatmap showing the $D_{\textrm{KL}}$ results between classes for (top) MNIST and (bottom) CIFAR-10 datasets.}
    \label{fig:KLD_mnist_cifar}
\end{figure}

The MNIST digit dataset contains 60000 training and 10000 test images of ten handwritten digits (0 to 9) that are $28\times28$ pixels in size. The gray scale levels were converted to the range $[\epsilon, 1-\epsilon], \epsilon\sim \mathcal N(0, 0.1)$.
The CIFAR-10 natural images dataset which contains 50000 training and 10000 test RGB images that are of size $3\times32\times32$ pixels. The color levels were converted to the range $[\epsilon, 1-\epsilon], \epsilon\sim \mathcal N(0, 0.1)$.
We train our RBIG algorithm to calculate the $D_{\textrm{KL}}$ between all of the pairs of classes in both directions ($D_{\textrm{KL}}(\x|\y)$ and $D_{\textrm{KL}}(\y|\x)$) for the MNIST dataset and the CIFAR dataset.
Figure \ref{fig:KLD_mnist_cifar} shows heatmaps for the MNIST and CIFAR-10 datasets. It can be seen that in both cases, digits and objects, the PDF of similar classes have lower $D_{\textrm{KL}}$. For instance in the MNIST case the digits 7 and 9 are close in both directions (from 7 to 9, and from 9 to 7, note that $D_{\textrm{KL}}$ is not symmetric). In the case of the objects the closest ones are \emph{cat} and \emph{dog}, or \emph{horse} and \emph{deer}.

      \subsection{Mutual Information, \texorpdfstring{$I(\x,\y)$}{I(x,y)}}
      \label{sec:MI}

In order to compute $I$ using RBIG we could apply:
\begin{eqnarray}
    I(\x,\y) &=& H(\x) + H(\y) - H([\x,\y])
            \label{eq:MI_as_H}
\end{eqnarray}
computing each entropy factor using RBIG as explained in sec.~\ref{sec:entropy}.
However note that while $I$ can be small (variables could be even independent), the three quantities in eq.~\ref{eq:MI_as_H}
may be large for multivariate data. Therefore relatively small errors in the computation of the entropy can be
large in comparison with the value of $I$ for variables with low statistical relation.
Here, we introduce a way of computing $I(\x,\y)$ which depends only on the computation of one measure.

\subsubsection{Computation using RBIG}
The proposed estimate is based on a well-known property of mutual information: it is invariant over a reparametrization of the space of each variable using smooth and uniquely invertible transformations \cite{Kraskov2004}.
Basically our proposal is to transform the two datasets, $\x$ and $\y$, to a domain where both follow a standard Gaussian. This removes all the total correlation contained on each dataset. 

\noindent{\bf Proposition:} The total correlation remaining among both gaussianized datasets is equivalent to the mutual information between the original datasets:
\begin{equation}
\Tilde{I}(\x, \y) = \Tilde{T}([\x',\y']),
\label{I_rbig}
\end{equation}
where $\x'=G_x(\x)$ and $\y'=G_y(\y)$. Therefore, the computation of $\Tilde{I}(\x, \y)$ is reduced to one measurement. Moreover, according to Eq.~\ref{Estim_T}, it reduces to a set of univariate operations.

\noindent{\bf Proof:} First we apply RBIG to both variables, $\x$ and $\y$, the new variables, $\x' = G_x(\x)$ and $\y' = G_y(\y)$, are Gaussian distributions with zero mean and identity covariance. Their joint entropies can be expressed as the sum of the marginal entropies, i.e. their total correlation is zero:  $ H(\x') = \sum_{i=1}^{D_x} H(x'_i)$ and $ H(\y') = \sum_{i=1}^{D_y} H(y'_i)$. Therefore if we use the definition of $I$ in entropy terms we have:
\begin{eqnarray*}
    I(\x',\y') &=& H(\x') + H(\y') - H([\x',\y']) \\
            &=& \sum_{i=1}^{D_x} H(x'_i) + \sum_{i=1}^{D_y} H(y'_i) - H([\x',\y'])
\end{eqnarray*}
Let us consider the variable result of stacking $\x'$ and $\y'$, $\mathbf{z} = [\x', \y']$. Previous equation can be expressed as:
\begin{eqnarray*}
    \Tilde{I}(\x',\y') &=& \sum_{i=1}^{D_x + D_y} \Tilde{H}(z_i) - \Tilde{H}(\mathbf{z}) = \Tilde{T}(\mathbf{z})
\end{eqnarray*}
On the other hand, note that $I$ is invariant under the reparametrization of the space of each variable using smooth and uniquely invertible transformations \cite{Kraskov2004}. If we compute $G_x(\x)$ and $G_y(\y)$ using RBIG the requirement for the transformations is satisfied. Taking this into account:
\begin{eqnarray*}
\Tilde{I}(\x, \y) &=& \Tilde{I}(\x', \y') \\
&=& \Tilde{T}(\mathbf{z}) \\ &=& \Tilde{T}([\x',\y']).
\end{eqnarray*}

\subsubsection{Validation on known PDFs}
Here we consider two different situations:
\begin{itemize}
    \item{\em Gaussian vs Gaussian with different covariance.} We generate data from Gaussians distributions with the same mean but difference covariance following the same procedure described in sec.~\ref{sec:kld_validation}.
    \item{\em Student vs Student.} We compare data generated from two multivariate Student distribution. The process is similar to the one described in sec.~\ref{sec:kld_validation}, yet here we use the same value for both distributions, i.e. $\nu = \nu_1 = \nu_2$, but we consider different values for $\nu = [3,5,20]$. The difference between both distributions came from the difference in the shape matrices.
        The shape matrices were different in each trial but keeping the value of mutual information in a controlled regime: the diagonal is forced to have a value of $10$ and the off-diagonal coefficients are generated sampling from a uniform distribution ${\mathcal U}(0,1)$.
\end{itemize}
Details of how to compute analytically the $I$ value for each configuration are given in the appendix \ref{app:formulas}. Table \ref{tab:MI} summarizes the results for a given number of samples. Exhaustive comparison can be found in the supplementary material. While the computation of mutual information in multiple dimensions is a particularly challenging problem, RBIG achieves reasonable low error rates. As in the previous experiments, the estimation performed using RBIG clearly outperforms the other methods.

\begin{table}[t!]
\begin{center}
\caption{Relative mean absolute errors in percentage for mutual information estimation on known PDFs. Best value in dark gray, second best value in light gray.}\label{tab:MI}
 \setlength{\tabcolsep}{5pt} 
\begin{tabular}{|l|l|l|l|l|l|l|l|l|}
\hline
        &          & $D_x$  & \textbf{RBIG}                 & \textbf{kNN}                  & \textbf{KDP} & \textbf{expF}                 & \textbf{vME}                  & \textbf{Ens}            \\
\hline
\parbox[t]{2mm}{\multirow{4}{*}{\rotatebox[origin=c]{90}{Gaussians}}}  & & 3            & \cellcolor[HTML]{C0C0C0}10.60 & 26.00                        & 149.10       & \cellcolor[HTML]{656565}9.20  & 13.20        & 48.50                         \\
    &              & 10           & \cellcolor[HTML]{656565}9.60  & 76.30                        & 102.60       & \cellcolor[HTML]{C0C0C0}23.70 & 311.00       & 91.00                         \\
    &              & 50           & \cellcolor[HTML]{656565}6.80  & 104.70                       & 100.70       & \cellcolor[HTML]{C0C0C0}39.50 & 68.00        & 105.50                        \\
    &              & 100          & \cellcolor[HTML]{656565}11.70 & 107.20                       & >1000 & \cellcolor[HTML]{C0C0C0}42.60 & 77.40        & 106.10                        \\
\hline
\parbox[t]{2mm}{\multirow{12}{*}{\rotatebox[origin=c]{90}{Student vs. Student}}}   & \parbox[t]{2mm}{\multirow{4}{*}{\rotatebox[origin=c]{90}{$\nu=3$}}} & 3    & \cellcolor[HTML]{656565}35.72 & 95.32                        & >1000 & \cellcolor[HTML]{C0C0C0}63.73 & >1000 & 86.58                         \\
    &              & 10           & 22.26                         & \cellcolor[HTML]{656565}2.66 & 118.51       & \cellcolor[HTML]{C0C0C0}18.14 & >1000 & 66.77                         \\
    &              & 50           & \cellcolor[HTML]{656565}1.51  & 88.38                        & 104.50       & \cellcolor[HTML]{C0C0C0}36.10 & 810.02       & 105.83                        \\
    &              & 100          & \cellcolor[HTML]{656565}15.34 & 98.66                        & >1000 & \cellcolor[HTML]{C0C0C0}65.71 & 789.55       & 105.34                        \\
\cline{2-9}
&  \parbox[t]{2mm}{\multirow{4}{*}{\rotatebox[origin=c]{90}{$\nu=5$}}}    & 3            & \cellcolor[HTML]{656565}18.51 & 118.04   & >1000 & \cellcolor[HTML]{C0C0C0}56.49 & >1000 & 96.41                         \\
  &                & 10           & \cellcolor[HTML]{656565}3.07  & 24.83                        & 113.89       & \cellcolor[HTML]{C0C0C0}9.39  & >1000 & 101.26                        \\
  &                & 50           & \cellcolor[HTML]{656565}10.91 & 102.89   & 105.08       & \cellcolor[HTML]{C0C0C0}25.17 & 849.12       & 117.30                        \\
  &                & 100          & \cellcolor[HTML]{656565}24.43 & 105.41   & 101.10       & \cellcolor[HTML]{C0C0C0}42.57 & 805.44       & 110.58                        \\
\cline{2-9}
& \parbox[t]{2mm}{\multirow{4}{*}{\rotatebox[origin=c]{90}{$\nu=20$}}}   & 3            & 73.63                         & 194.16   & >1000 & \cellcolor[HTML]{656565}14.63 & >1000 & \cellcolor[HTML]{C0C0C0}15.36 \\
 &                 & 10           & \cellcolor[HTML]{C0C0C0}40.02 & 108.82   & 110.68       & \cellcolor[HTML]{656565}29.69 & >1000 & 208.20                        \\
 &                 & 50           & \cellcolor[HTML]{656565}29.98 & 149.53   & 102.93       & \cellcolor[HTML]{C0C0C0}36.30 & 946.93       & 154.88                        \\
 &                 & 100          & \cellcolor[HTML]{656565}37.21 & 128.27   & 101.44       & \cellcolor[HTML]{C0C0C0}43.77 & 844.41       & 127.67                      \\
\hline
\end{tabular}
\end{center}
\end{table}

\subsubsection{Experiments on real-world data: learning in neural networks}
RBIG estimates of mutual information have been used to analyze the transmission of information in artificial and biological neural networks. 
For instance, in \cite{Malmgren17igarss} RBIG estimates of $I$ between different features and the outputs of the network were used to compare different feature extraction methods before using a convolutional neural network (CNN). In \cite{IMM2018} RBIG estimates of $I$ were used to analyze overfitting in CNN predictions. In \cite{Malo19} RBIG was used to measure the visual information available from different neural layers in psychophysically tuned networks.

Here we present a novel analysis of a learning system 
in the context of the information bottleneck \cite{Tishby2015INFOBOTTLE}
in artificial neural networks (ANNs).
We illustrate the use of $I$ via RBIG to monitor the information shared between each layer of an ANN and the actual outputs while the network is learning to solve a regression problem. 

The regression problem considered here consists of predicting atmosphere temperatures from radiance measurements.
The input samples were acquired by the infrared atmospheric sounding interferometer (IASI) sensor in the MetOp satellite,
The data has originally 8461 spectral bands which have been reduced to 50 dimensions using PCA, i.e. $\x \in \mathbb{R}^{50}$. 
The desired output is the temperature at three different pressure levels in the atmosphere, $\y \in \mathbb{R}^{3}$:  $10^3$ hPa (land surface), $10$ hPa ($\sim 30$ km) and $10^{-2}$ hPa ($\sim 80$ km).
Labeled data were obtained using the European Centre for Medium-Range Weather Forecasts (ECMWF) analysis model (see \cite{Sobrino19} for a more exhaustive data description). The number of samples used is 100000: 80000 for training and 20000 for testing the network.

\begin{figure*}
    \centering
        \begin{tabular}{cc}
        a) & b) \\
        \includegraphics[width=6.9cm]{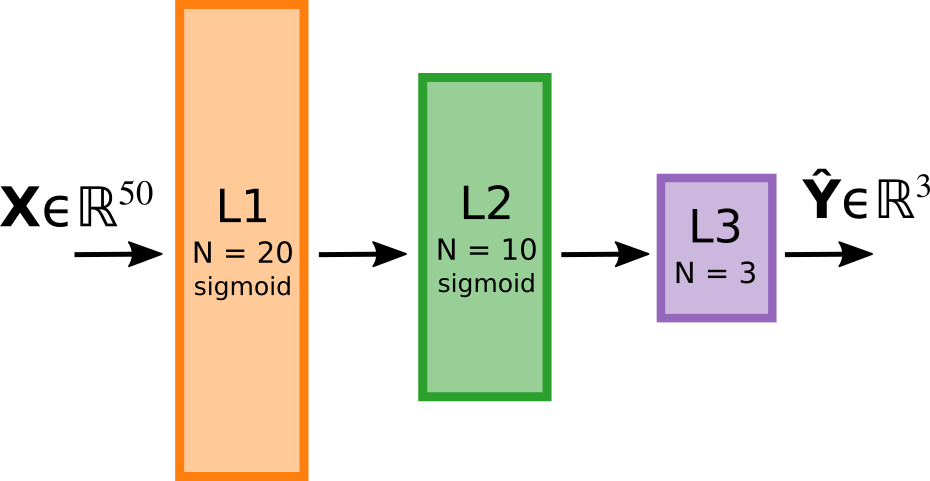} &
        \includegraphics[width=6.9cm]{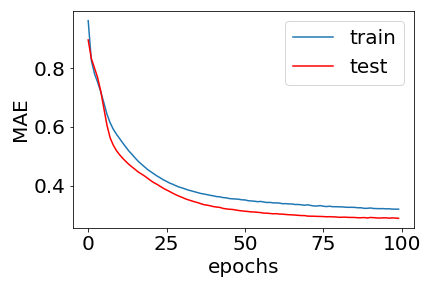} \\
        c) & d) \\
        \includegraphics[width=6.9cm]{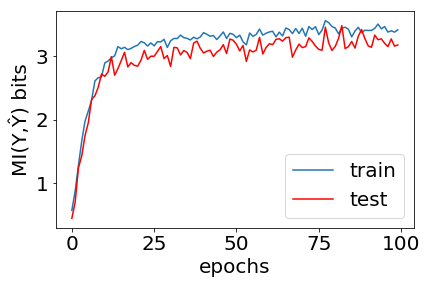} &
        \includegraphics[width=6.9cm]{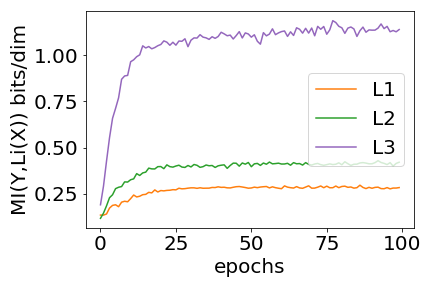}
        \end{tabular}
    \caption{Learning in artificial neural networks from RBIG estimations of mutual information: evolution of $I$ during the training of an ANN. a) Configuration of the considered neural network. b) Error evolution. c) Evolution of $I$ between the predicted output and the actual data.  d) Evolution of $I$ per dimension between the output of each layer and the actual data.}
    \label{fig:Real_MI_ANN}
\end{figure*}

In the experiment we illustrate how $I$ can be used as a metric to monitor the output of the network (as regular metrics do) but also it can be used to monitor the evolution of the intermediate layers during the training process. The experiment setup and results are illustrated in Fig.~\ref{fig:Real_MI_ANN}.
The considered ANN has three dense layers of 20, 10, and 3 neurons respectively, Fig. \ref{fig:Real_MI_ANN}a).
The first two layers use a sigmoid  activation function, the third layer does not have an activation function since we do not want to restrict the output domain. 
The ANN has been trained for 100 epochs to minimize the Mean Absolute Error (MAE) with the actual temperatures, Fig. \ref{fig:Real_MI_ANN}b). 
Interestingly, Fig. \ref{fig:Real_MI_ANN}c) shows how the $I$ between the predicted and the actual outputs evolves consistently with the MAE
for both training and test data.
Fig. \ref{fig:Real_MI_ANN}d) shows the evolution of the $I$ between the output of each layer and the output for the training data.

A number of consequences can be derived. First, RBIG-based estimations of $I$ can be used as a regular metric to monitor the predictions.
More interestingly, $I$ can be used to monitor the particular evolution of each layer (not only the last one), where conventional metrics 
(such as MAE or MSE) are not applicable. In the example we can see how the last layer (L3) learns very fast at the beginning and, at some point, the learning rate drops, and it keeps learning slowly. On the other hand, the first layer (L1) starts to learn slowly, it keeps learning at medium rate for longer time, and at some point it saturates and seems to stop learning. The behavior of the intermediate layer (L2) is in between L1 and L3.

      \section{Concluding Remarks} 
      
In this work we propose the use of the Rotation-Based Iterative Gaussianization (RBIG)
for computing Information Theory Measures on multidimensional data.
Given the special role of total correlation in RBIG, it is the best choice among Gaussianization
transforms to compute information measures. As opposed to other Gaussianization transforms, in RBIG the estimation of multivariate total correlation reduces to a set of univariate operations: it does not involve integrating multivariate quantities over a multivariate domain.
More specifically, we derive particular expressions to compute \emph{total correlation}, \emph{differential entropy}, \emph{Kullback-Leibler divergence} and \emph{mutual information} from observational data.

Experiments on data drawn from a variety of distributions where analytical results are available allow us a systematic assessment of the quality of the proposed method. The RBIG-based estimates are compared to representative methods of different families, namely,
nearest neighbors~\cite{Kozachenko1987ASE,Goria05}, partition trees~\cite{Stowell09}, ensemble methods~ \cite{Kybic2004}, the exponential family~\cite{Nielsen_2011}, and Von Mises expansion~\cite{NIPS2015_5911}, as included in the most comprehensive and up-to-date toolbox~\cite{szabo14information}. Experiments show that the proposed methodology is the best choice in general, and the advantages are particularly
relevant for high dimensional scenarios and when the Gaussian approximation is not reasonable.

We provide a toolbox with the RBIG-based estimators in Matlab and Python, and the routines and data to reproduce all the experiments on known distributions. This represents a standard procedure to systematically test future estimates of the considered information measures,  \url{https://isp.uv.es/RBIG4IT.htm}.

The practical use of information theory measures with real data is commonly restricted to very low dimensional scenarios (either one or two dimensions). In this work we show that the RBIG-based information estimates can address interesting scientific issues in a range of real-world problems. In particular, RBIG-based measures provide sensible qualitative insight
on the Efficient Coding Hypothesis in \emph{Visual Neuroscience},
on the discovery of spatio-temporal regularities in \emph{Geoscience} essential climate variables,
on the complexity of image databases in \emph{Computer Vision},
and on the learning rate and representations in \emph{Artificial Neural Networks} in a \emph{Remote Sensing} problem. 
These results suggest that the proposed estimators open a range of possibilities to use information-theory measures in questions that could not be addressed before.

      \label{sec:conclusions}



\bibliographystyle{IEEEtran}
\bibliography{biblio2}


\section{Appendix: expressions for experiments}
\label{app:formulas}
In this appendix we give details on the analytical formulas to compute the different information theory measures for the distributions employed in this paper.

\subsection{Total correlation}

\subsubsection{Gaussian}
For a Gaussian distribution, $\x \sim \mathcal{N}(\x,\mu,\Sigma)$, the analytic value of total correlation is:
\begin{eqnarray*}
T(\x) &=& \sum_{i=1}^{D_x} H(x_i) - H(\x) \\
&=& \sum_{i=1}^{D_x} \frac{1}{2} \log(2 e \pi \sigma_i^2) - \frac{1}{2} \log(|2 e \pi \Sigma|) \\
&=& \sum_{i=1}^{D_x} \log (\sigma_i) - \frac{1}{2}\log(|\Sigma|). \\
\end{eqnarray*}

\subsubsection{Linearly transformed uniform distribution}

The total correlation of a linearly transformed uniform distribution can be computed using the variation of total correlation under transformations defined in Eq.~\eqref{deltaT}.
By starting from a multidimensional distribution which marginals are independent, i.e. $T(\x) = 0$, we can easily compute the total correlation of a rotated and scaled version, $\y = \bf{M}  \x$:

\begin{equation*}
     T(\y) =  \sum_{i=1}^{D_y} H(y_i) - \sum_{i=1}^{D_x} H(x_i) - \frac{1}{2}\log(|\bf{M}^\top\bf{M}|).
     \label{TC_rot}
\end{equation*}
Since we know the applied linear transformation we can compute {$\log(|\bf{M}^\top\bf{M}|)$}. The marginal entropies for the transformed data are computed empirically from a dataset of $5 \cdot 10^5$ samples.

\subsubsection{Multivariate t-Student}

In \cite{Guerrero-Cusumano1998} the total correlation shared by the coefficients of samples that follow a multivariate t-Student distribution of dimension $D$, $\nu$ degrees of freedom, and scale matrix (or shape matrix) $\A$, was given:
\begin{eqnarray*}
 T(D,\nu,\A) &=& -\frac{1}{2} \log(\det(\A)) \\
 &+& \log\left[ \frac{\Gamma(\frac{D}{2})[\beta(\frac{\nu+1}{2},\frac{1}{2})]^D}{(\pi)^{\frac{D}{2}}\beta(\frac{\nu+D}{2},\frac{D}{2})} \right] \\
 &+& \frac{D(\nu+1)}{2} \left[ \psi \left(\frac{\nu+1}{2} \right) - \psi \left(\frac{\nu}{2} \right)\right]\\
 &-& \frac{\nu+D}{2} \left[ \psi \left(\frac{\nu+D}{2} \right) - \psi \left(\frac{\nu}{2} \right)\right].
 \end{eqnarray*}

\subsection{Entropy}

\subsubsection{Gaussian}

Entropy of a Gaussian distribution, $\x \sim \mathcal{N}(\mu,\Sigma)$, can computed in nats as:
\begin{eqnarray*}
     H(\x) &=& \frac{D_x}{2} + \frac{D_x}{2}\log(2 \pi) + \frac{1}{2} \log(|\Sigma|).
\end{eqnarray*}

\subsubsection{Linearly transformed uniform distribution}

As for the total correlation we use a linearly transformed version of a multidimensional uniform distribution which marginals are independent, i.e. $T(\x) = 0$. Therefore, using Eqs. \eqref{eq:TC} and \eqref{deltaT}, we can compute a reference based only on marginal entropy estimations as follows:
\begin{eqnarray*}
 H(\y) &=& \sum_{i=1}^{D_y} H(y_i) - T(\y) \\
 &=& \sum_{i=1}^{D_x} H(x_i) + {\frac{1}{2}\log(|\bf{M}^\top  \bf{M}|)}.
\end{eqnarray*}
This reference will be a true value if $H(x_i)$ is analytic. In our case $M$ is squared.

\subsubsection{Multivariate Student}

 In \cite{GUERREROCUSUMANO1996} an analytical definition of the entropy of a multivariate Student was given by
\begin{eqnarray*}
 H(D,\nu,\A) &=& \frac{1}{2} \log(|\A|) + log\left[ \frac{(\nu\pi)^{\frac{D}{2}}}{\Gamma(-\frac{D}{2})} \beta (\frac{D}{2},\frac{\nu}{2}) \right] \\
 &+& \frac{\nu+D}{2} \left[ \psi(\frac{\nu+D}{2}) - \psi(\frac{\nu}{2})\right].
 \end{eqnarray*}

\subsection{Kullback-Leibler Divergence}

\subsubsection{Gaussian}

$D_{\textrm{KL}}$ between two Gaussians $\x_1 \sim {\mathcal N}(\mu_1\mathbf{1},\Sigma_1)$ and $\x_2 \sim {\mathcal N}(\mu_2\mathbf{1},\Sigma_2)$ can be calculated as
\begin{equation*}
{{D}_{\textrm{KL}}(\x_1|\x_2) = \\ 
\frac{1}{2} \left( \text{tr}(\Sigma_2^{-1}\Sigma_1) + P - D + \log\frac{|\Sigma_2|}{|\Sigma_1|} \right),}
\label{eq:KLD_gauss}
\end{equation*}
where $\text{tr}(·)$ is the trace function, $P = (\mu_1 + \mu_2)^{\top} \Sigma_2^{-1} (\mu_1 + \mu_2)$, and $\mathbf{1}$ is a vector of ones.

\subsubsection{Multivariate Student}

The $D_{\textrm{KL}}$ between two multivariate Student-t distributions ($\x_1 \sim f_1(\mu_1=0,\Sigma_1=\mathbf{I},\nu_1)$ and $\x_2 \sim f_2(\mu_2=0,\Sigma_2=\mathbf{I},\nu_2)$) can be calculated as (\cite{Villa2018} eq. 8):
\begin{eqnarray*}
{D}_{\textrm{KL}}(\x_1|\x_2) =  \log\bigg(\frac{{K}_1}{{K}_2}\bigg) - E_1 + E_2,
\label{eq:KLD_student}
\end{eqnarray*}
with
\begin{eqnarray*}
K_1 = \dfrac{\Gamma(\dfrac{\nu_1+D}{2})}{\Gamma(\dfrac{\nu_1}{2}) \sqrt{(\pi\nu_1)^D}}
\end{eqnarray*}
and
\begin{eqnarray*}
E_1 &=& \frac{\nu_1 + D}{2} \mathbb{E}_{D,\nu_1} \left[ \log\bigg(1+ \frac{\X_1^\top \X_1}{\nu_1} \bigg)\right] \\
&=& \dfrac{\nu_1 + D}{2}  \left( \Psi\bigg(\frac{\nu_1+D}{2}\bigg) - \Psi\bigg(\dfrac{\nu_1}{2}\bigg) \right)
\end{eqnarray*}

\section{Appendix: Detailed results}
\label{app:figuras}

In this appendix we show the detailed results for the entropy estimation (sec. \ref{sec:entropy}), $D_{KL}$ estimation (sec. \ref{sec:kld}) and MI estimation (sec. \ref{sec:MI}). In each section these results are summarized in the corresponding table. Details on the experimental setup for each experiment are given in the corresponding section.

In all the plots we give the mean absolute error in percentage regard the value computed using the analytical formula for different number of data samples. Mean values (solid line) and standard deviation (shadow) are given for five trials. In each figure each column correspond to a particular number of dimensions in the data. Each color correspond to a particular method explained in sec.\ref{sec:intro_methods}: RBIG (red), kNN (yellow), KDP (blue), expF (green), vME (black) and ensemble (purple).

\begin{figure*}[H]
    \centering
    \begin{tabular}{m{0.8mm}m{41mm}m{41mm}m{41mm}m{41mm}}
      & $D = 3$ & $D = 10$ & $D = 50$ & $D = 100$ \\
\parbox[t]{2mm}{\multirow{1}{*}{\rotatebox[origin=c]{90}{Gaussian}}} &
	\includegraphics[width=3.9cm]{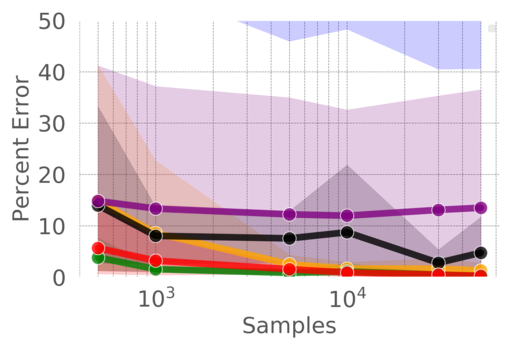} &
    \includegraphics[width=3.9cm]{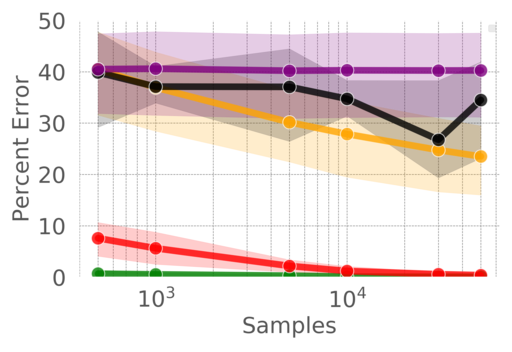} &
    \includegraphics[width=3.9cm]{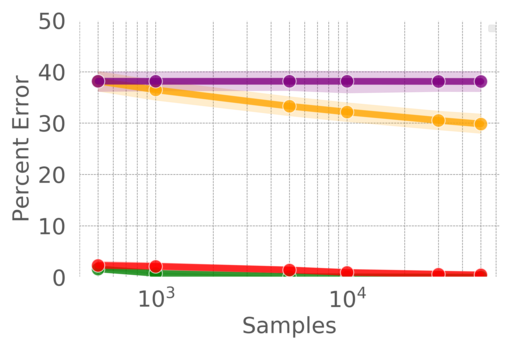} &
    \includegraphics[width=3.9cm]{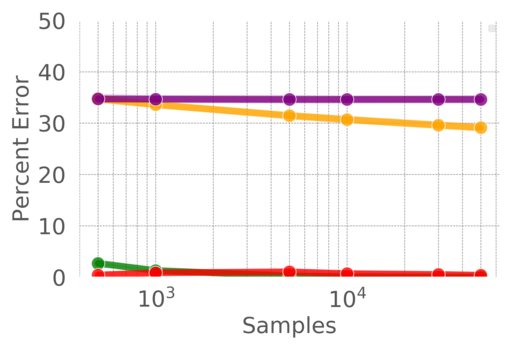} \\
\hline
\parbox[t]{2mm}{\multirow{1}{*}{\rotatebox[origin=c]{90}{Uniform}}} &
    \includegraphics[width=3.9cm]{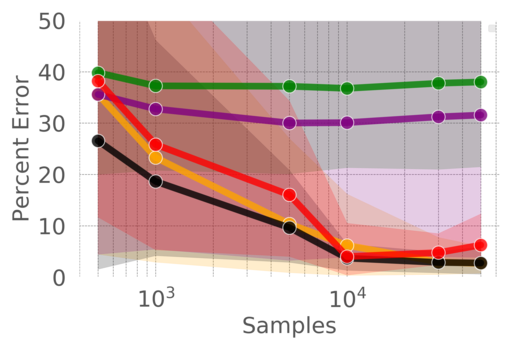} &
    \includegraphics[width=3.9cm]{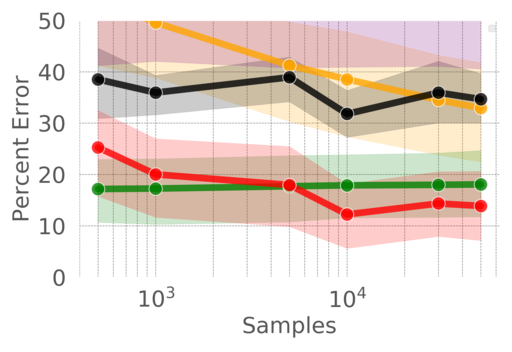} &
    \includegraphics[width=3.9cm]{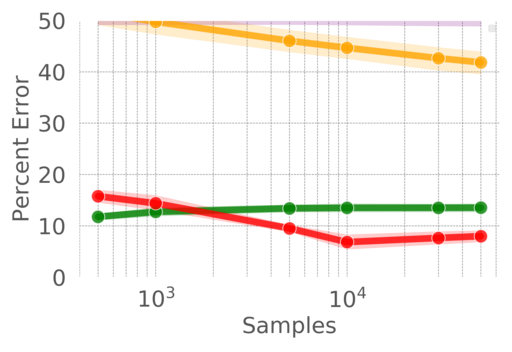} &
    \includegraphics[width=3.9cm]{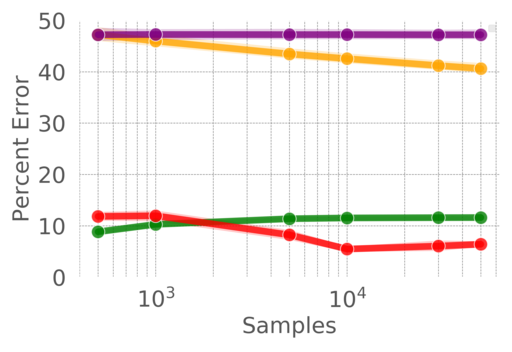}\\
\hline
\parbox[t]{2mm}{\multirow{18}{*}{\rotatebox[origin=c]{90}{Student-t}}} &
     \includegraphics[width=3.9cm]{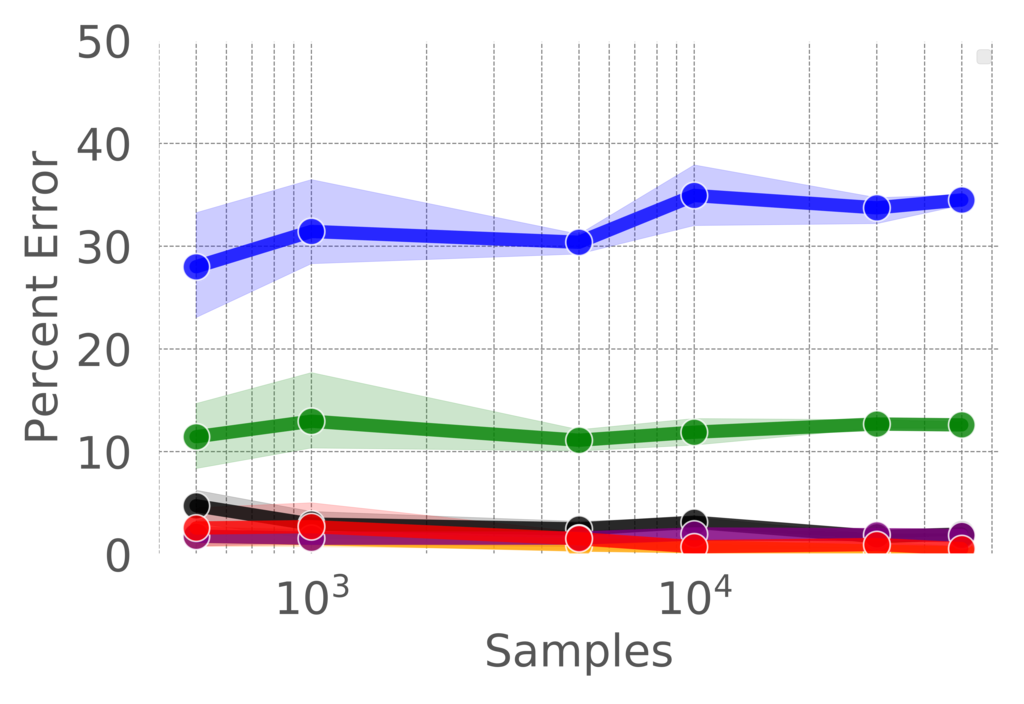} &
     \includegraphics[width=3.9cm]{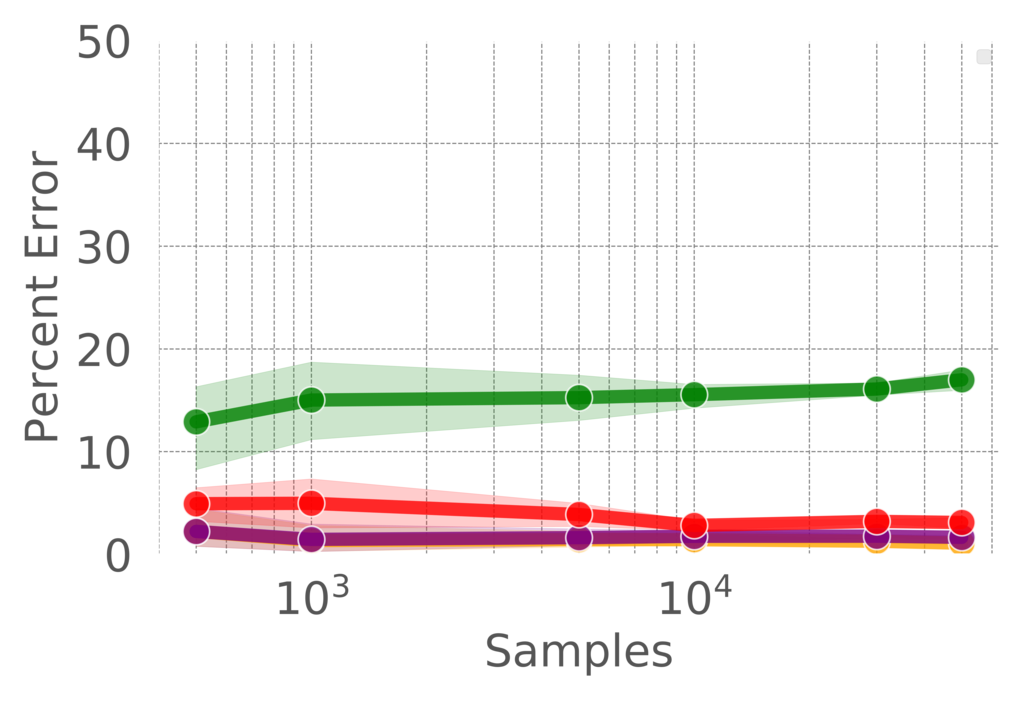} &
     \includegraphics[width=3.9cm]{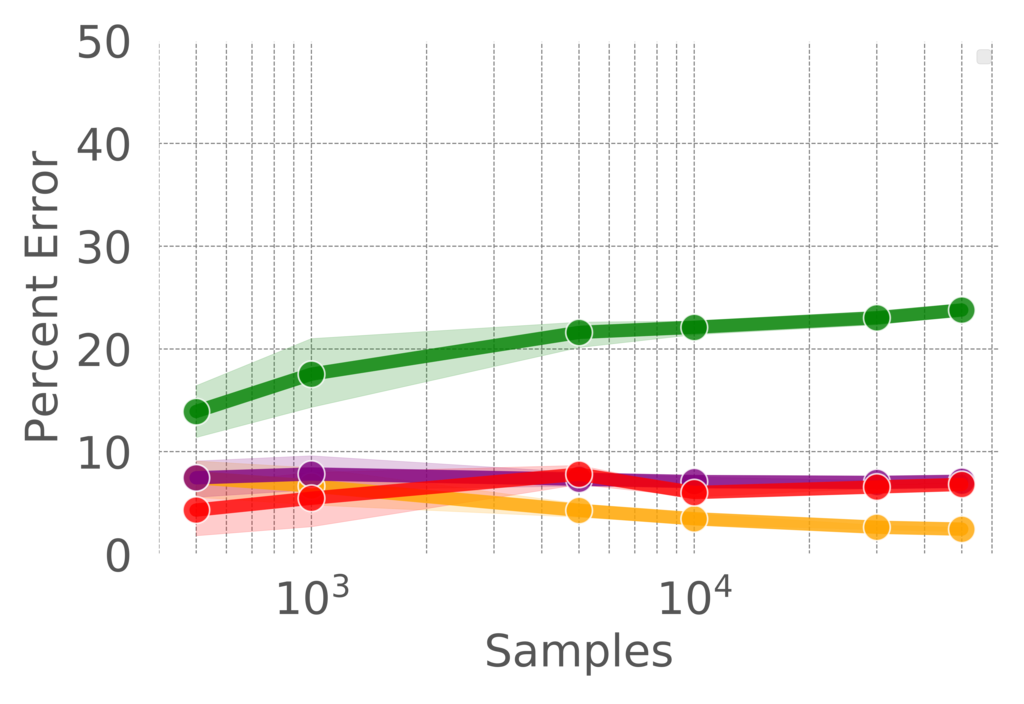} &
     \includegraphics[width=3.9cm]{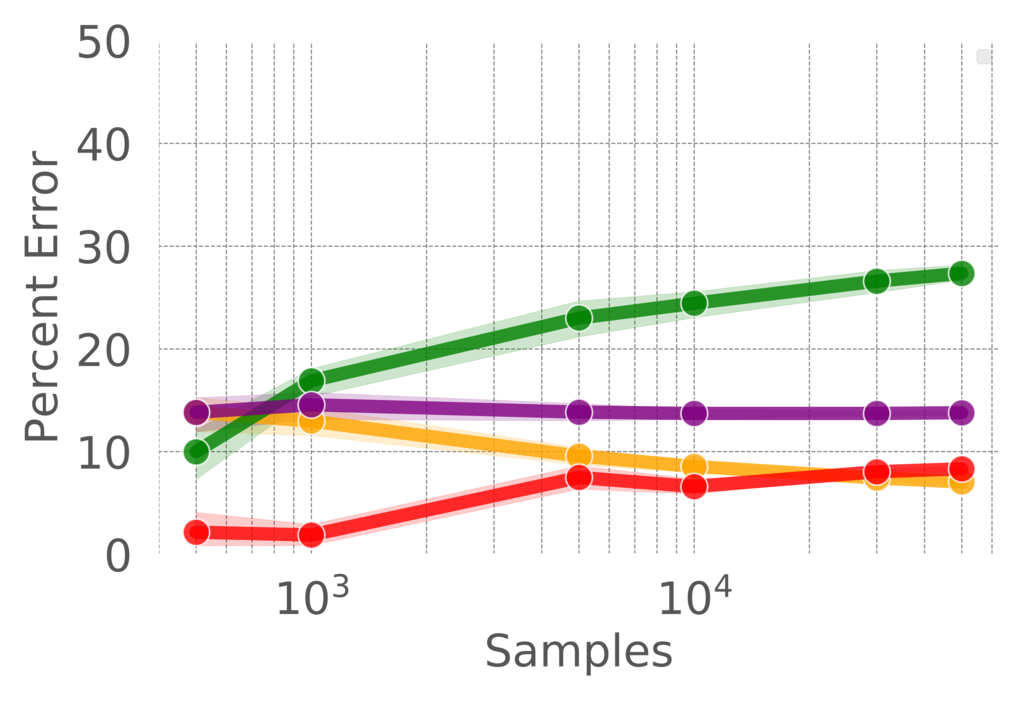} \\
&    \includegraphics[width=3.9cm]{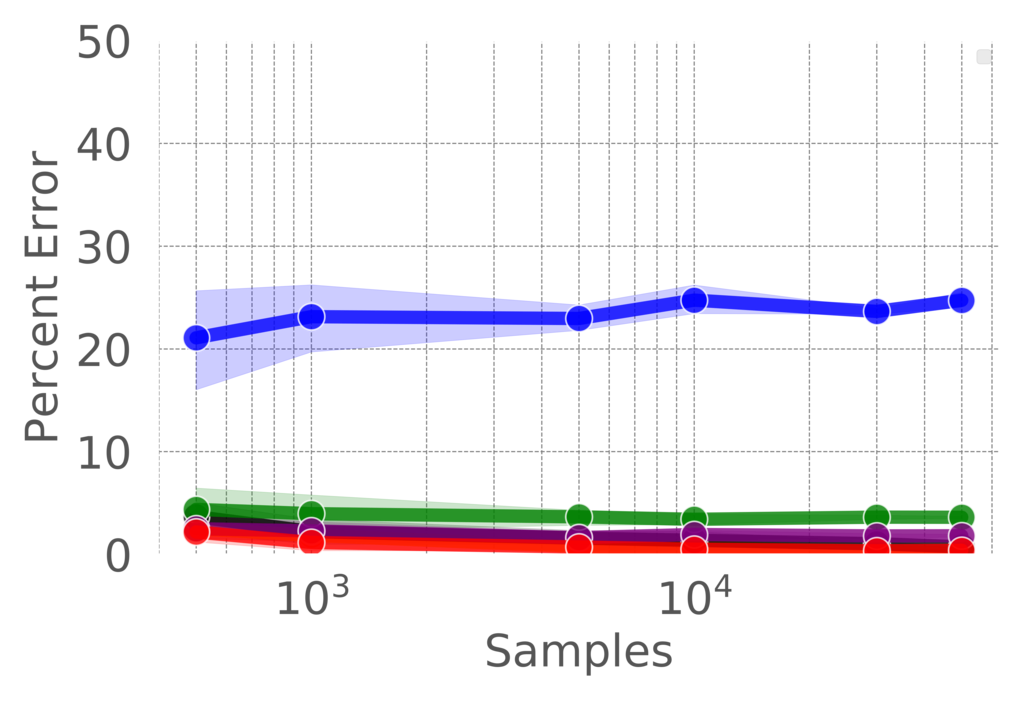} &
     \includegraphics[width=3.9cm]{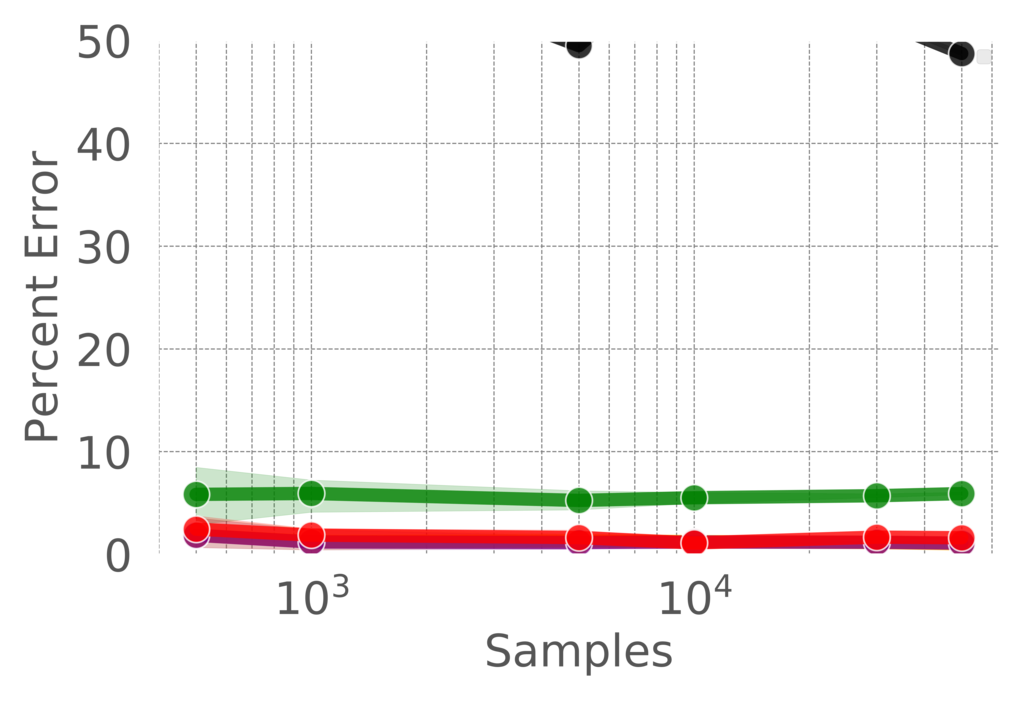} &
     \includegraphics[width=3.9cm]{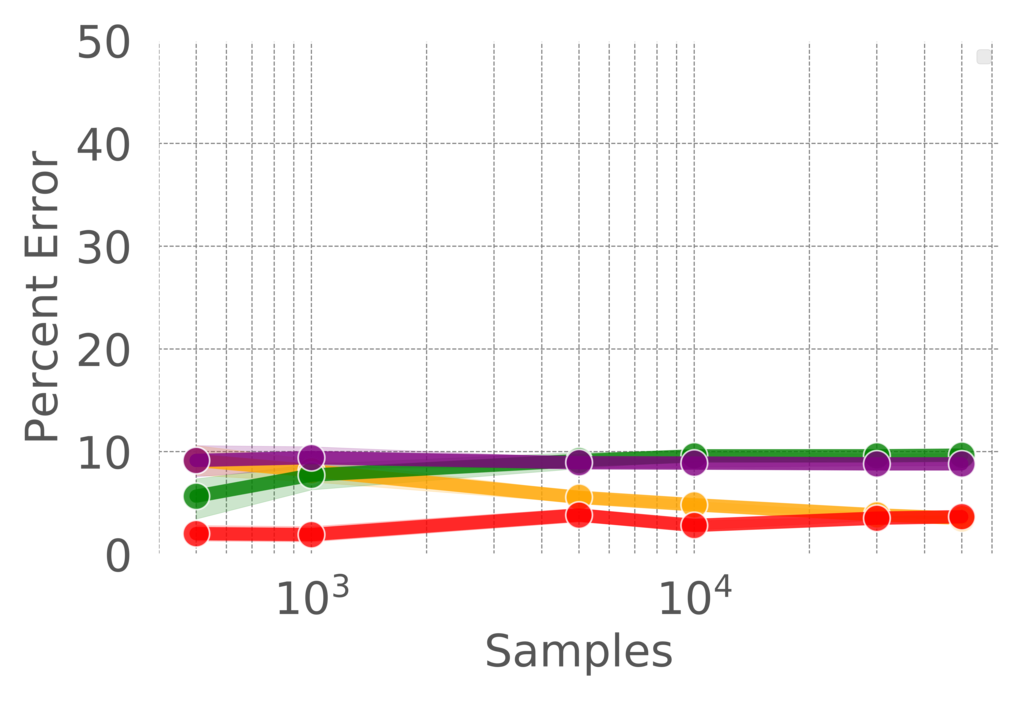} &
     \includegraphics[width=3.9cm]{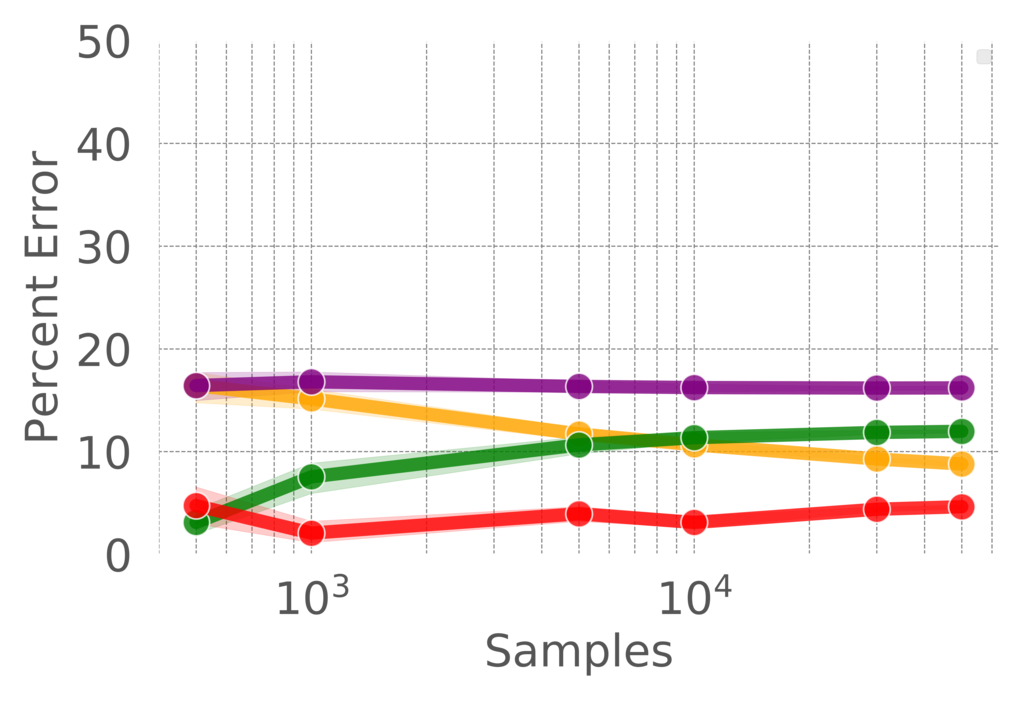} \\
&    \includegraphics[width=3.9cm]{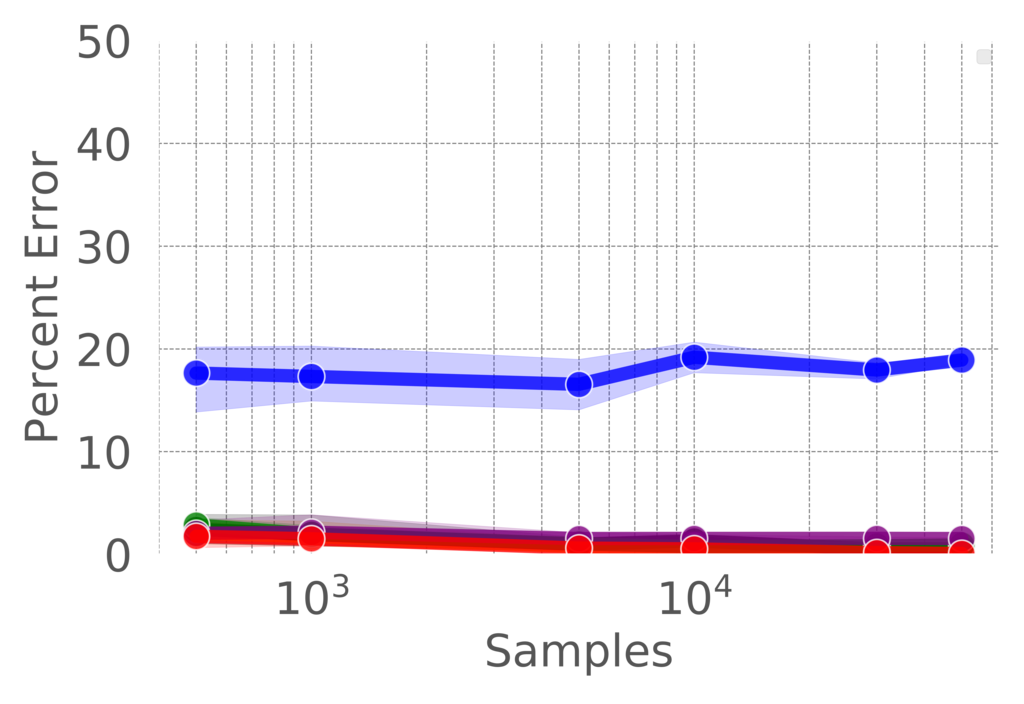} &
     \includegraphics[width=3.9cm]{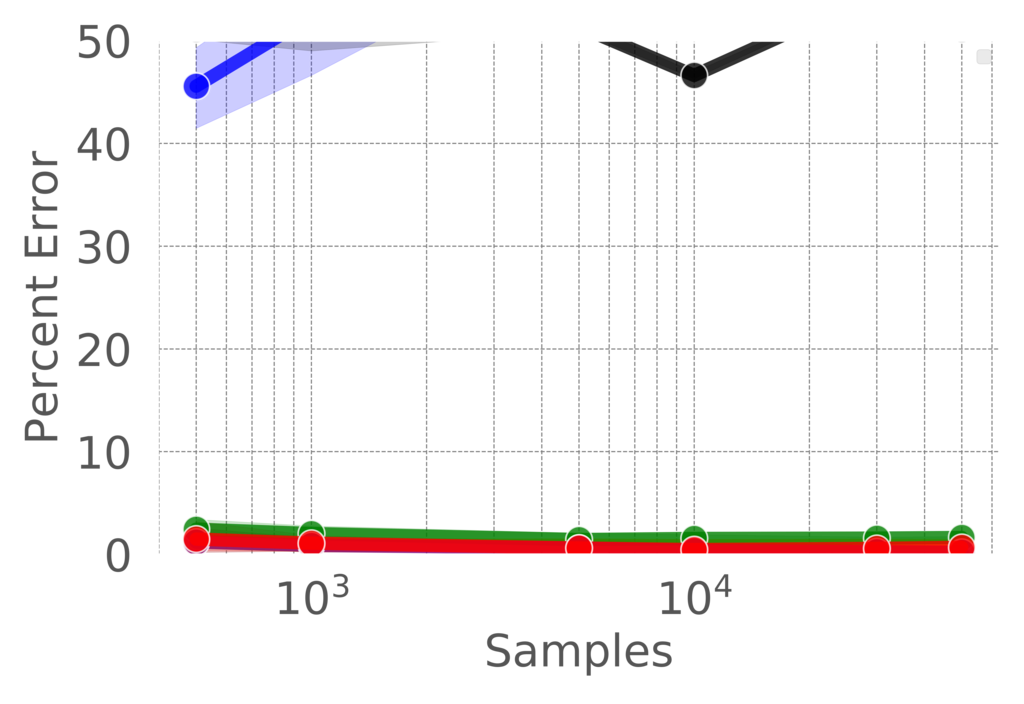} &
     \includegraphics[width=3.9cm]{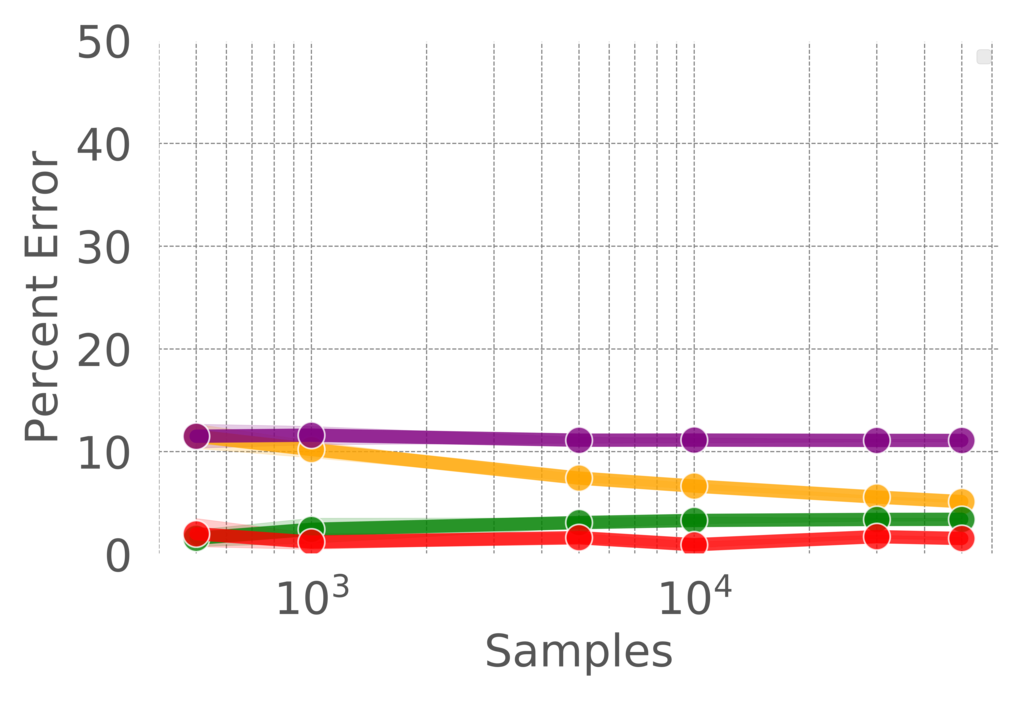} &
     \includegraphics[width=3.9cm]{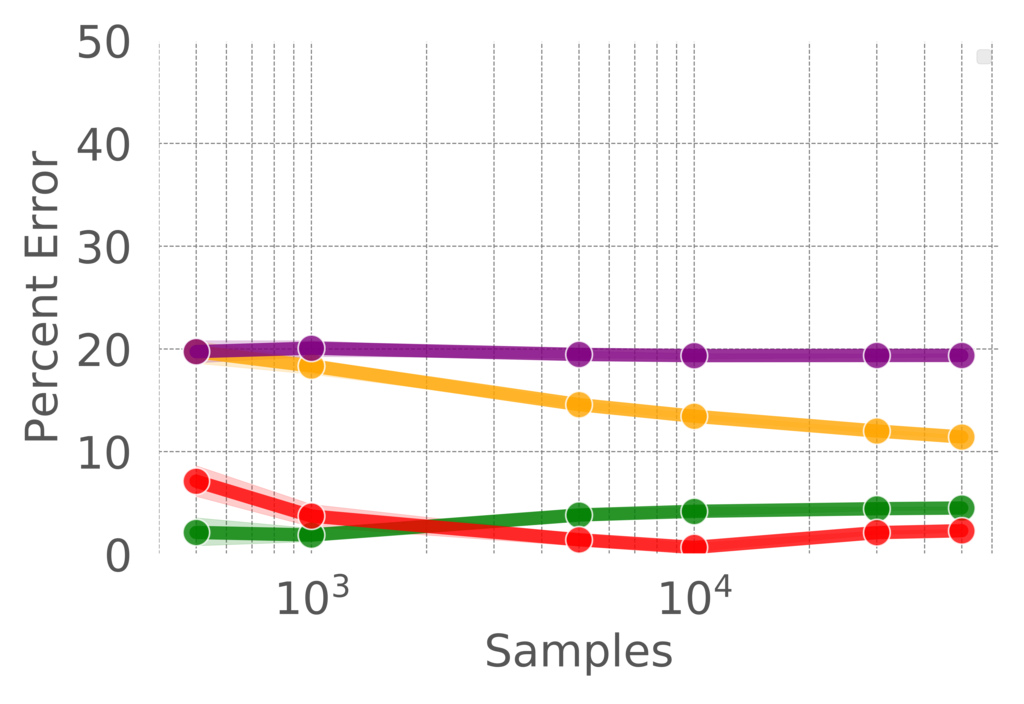}
    \end{tabular}
    \begin{tabular}{m{150mm}}
    \includegraphics[width=3.9cm, trim={0 100mm 0 0}, clip]{Figures/experiments/FIGS_TC/legend.png}\\
\end{tabular}
    \caption{Entropy estimation results in relative mean absolute error (see sec. \ref{sec:entropy} for detail). Results for different distributions are given: Gaussian, uniform and the Student-t PDFs ($\nu = 3,5,20$ for each row respectively). Each column correspond to an experiment of a particular number of dimensions $D$. Mean and standard deviation are given for five trials.}
    \label{fig:Entropy_plots}
\end{figure*}


\begin{figure*}
    \centering
    \begin{tabular}{cccc}
    D = 3 & D = 10 & D = 50 & D = 100 \\
    \includegraphics[width=3.9cm]{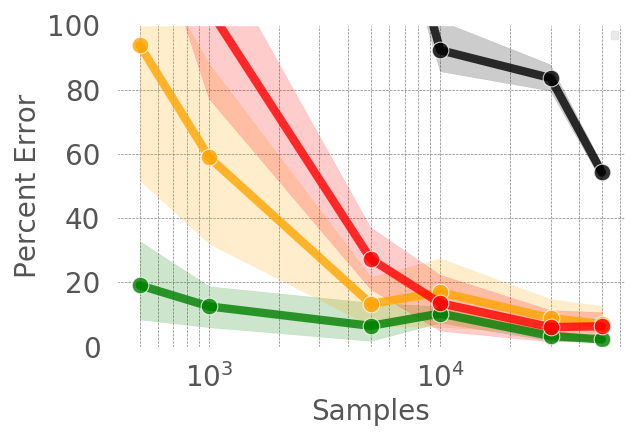} &
    \includegraphics[width=3.9cm]{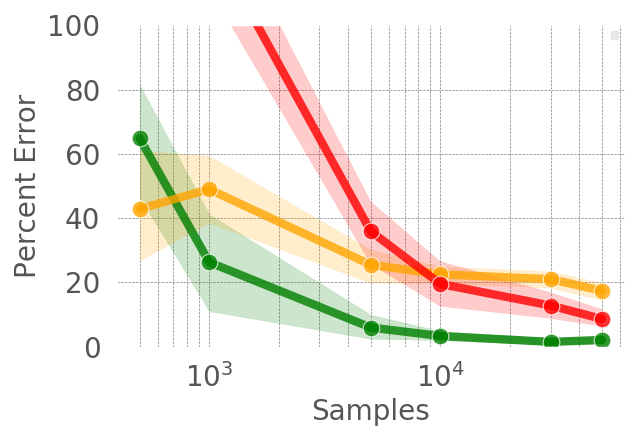} &
    \includegraphics[width=3.9cm]{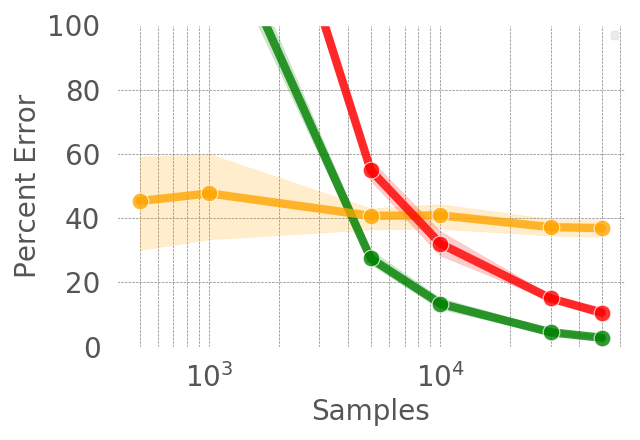} &
    \includegraphics[width=3.9cm]{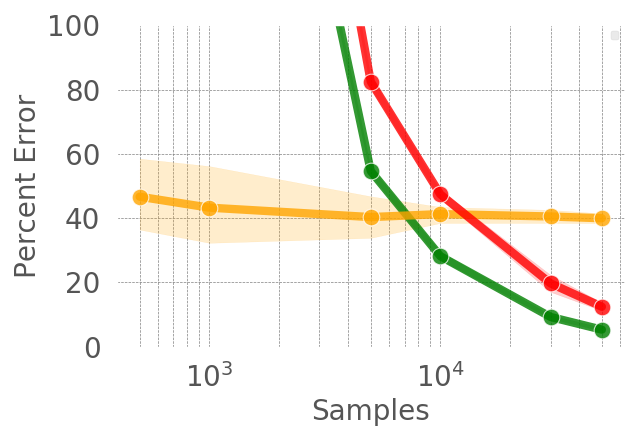}\\
    \includegraphics[width=3.9cm]{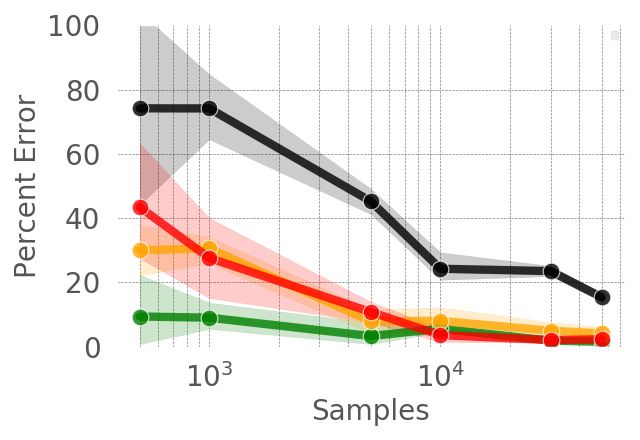} &
    \includegraphics[width=3.9cm]{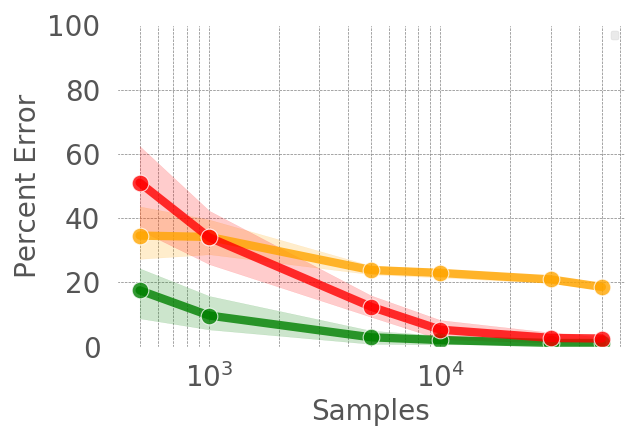} &
    \includegraphics[width=3.9cm]{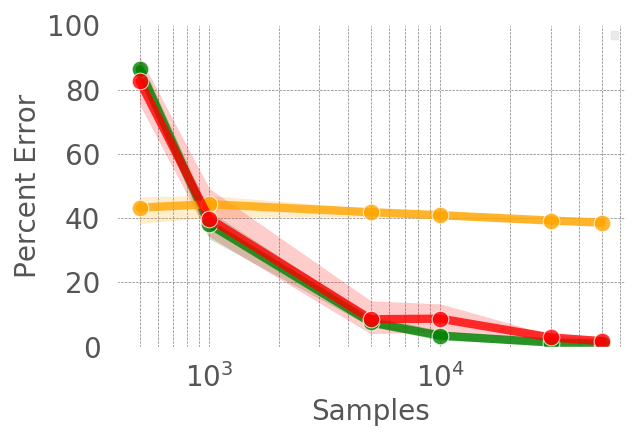} &
    \includegraphics[width=3.9cm]{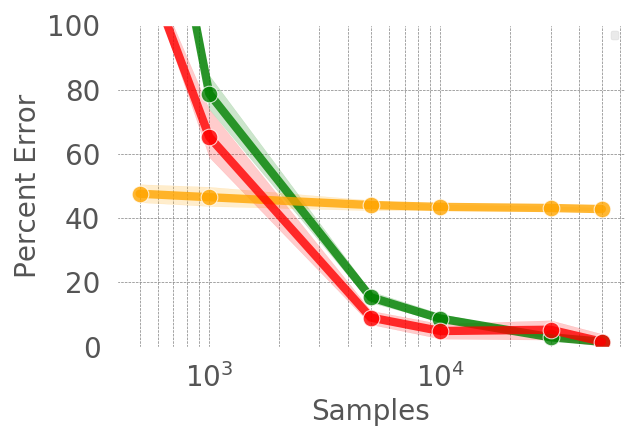}\\       \includegraphics[width=3.9cm]{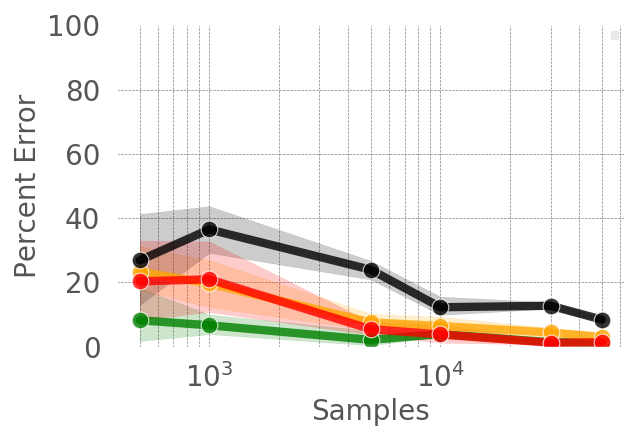} &
    \includegraphics[width=3.9cm]{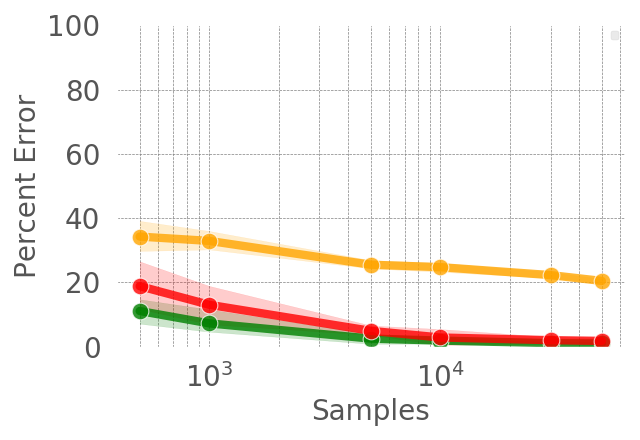} &
    \includegraphics[width=3.9cm]{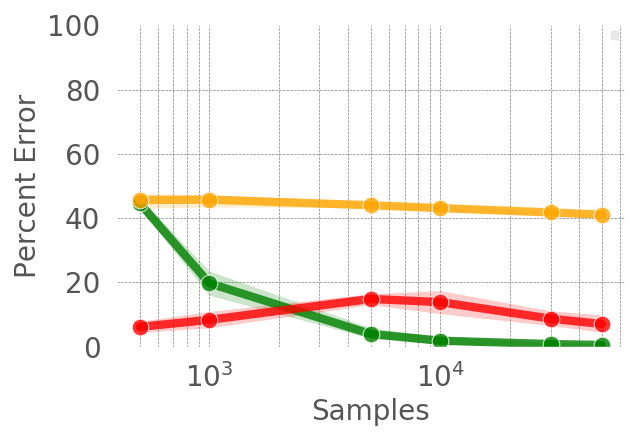} &
    \includegraphics[width=3.9cm]{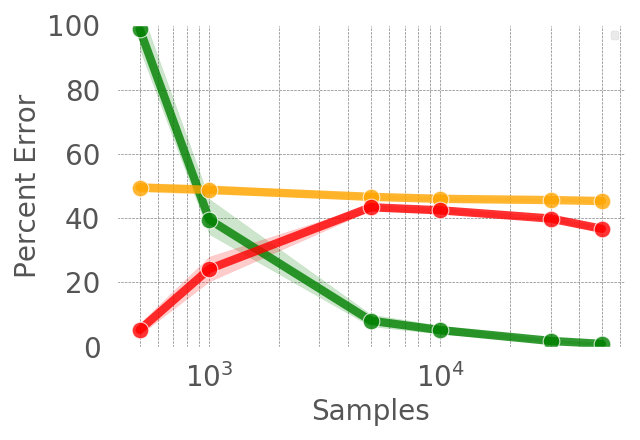}
    \end{tabular}
    \begin{tabular}{m{150mm}}
    \includegraphics[width=3.9cm, trim={0 100mm 0 0}, clip]{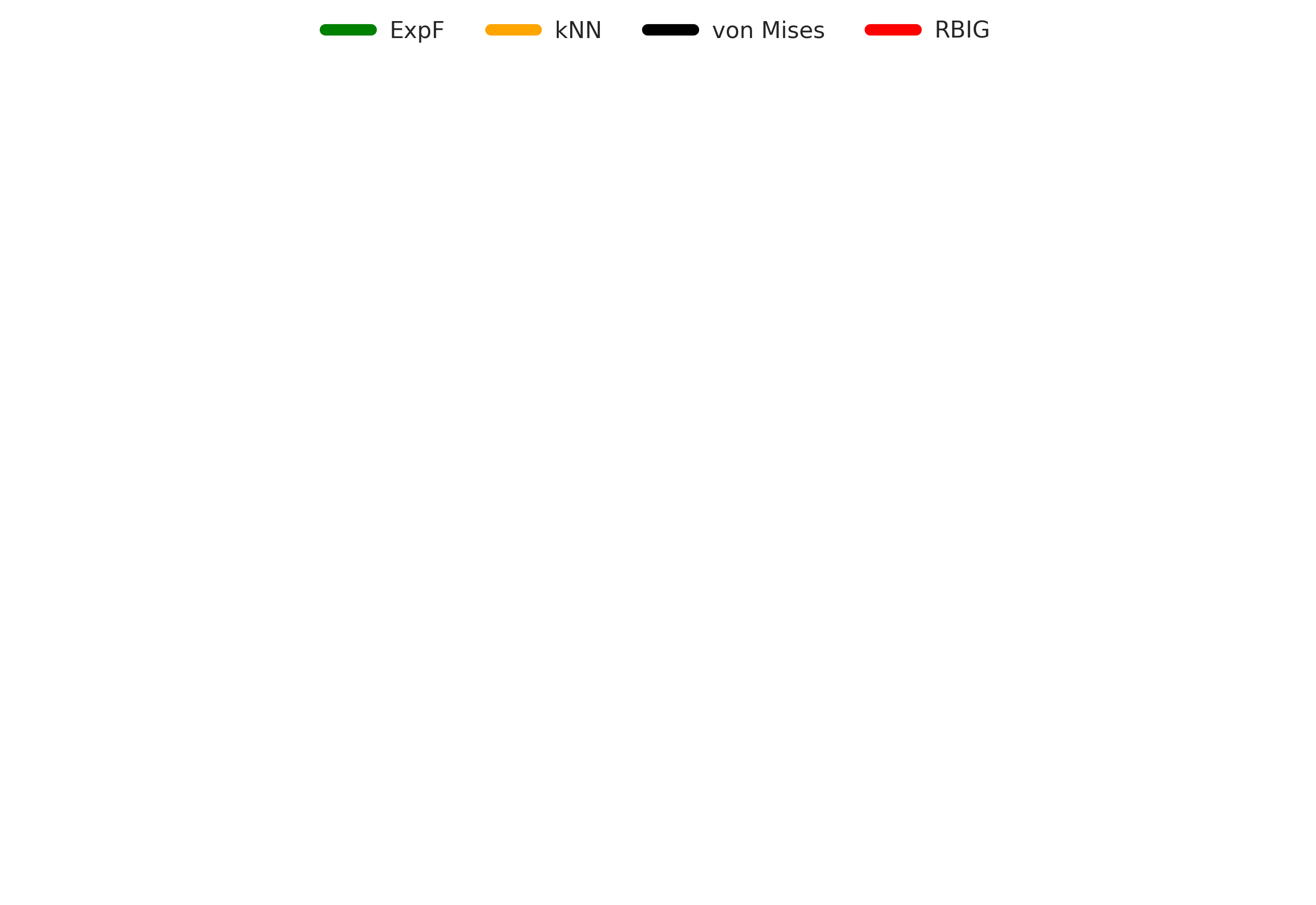}\\
    \end{tabular}
        \caption{$D_{KL}$ estimation results for the Gaussian distribution when modifying the mean (see sec. \ref{sec:kld} for detail). The mean value tested is respectively in each row: $\mu_2 = 0.2$, $\mu_2 = 0.4$ and $\mu_2 = 0.6$. }
    \label{fig:KLD_gauss_vs_gauss_mean}
\end{figure*}

\begin{figure*}
    \centering
    \begin{tabular}{cccc}
    D = 3 & D = 10 & D = 50 & D = 100 \\
    \includegraphics[width=3.9cm]{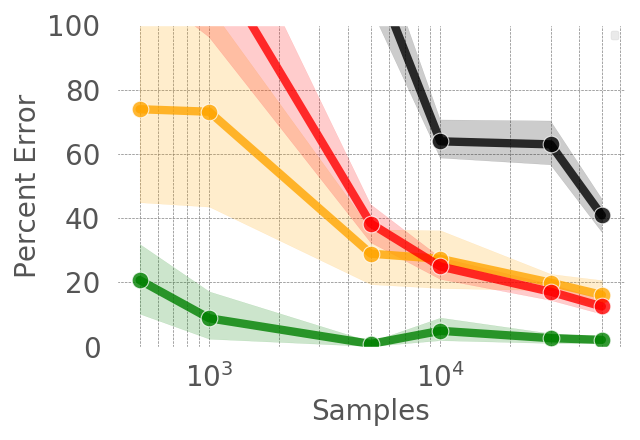} &
    \includegraphics[width=3.9cm]{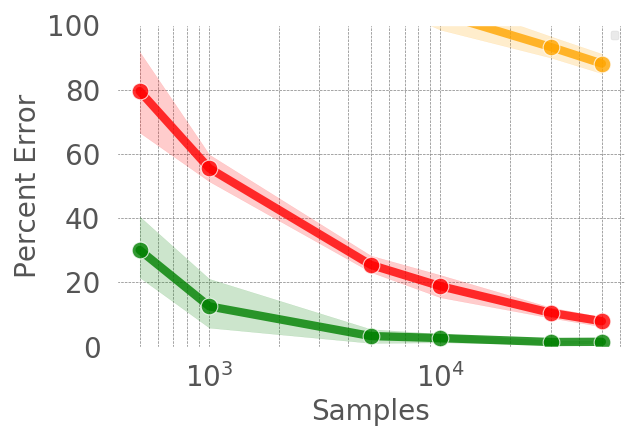} &
    \includegraphics[width=3.9cm]{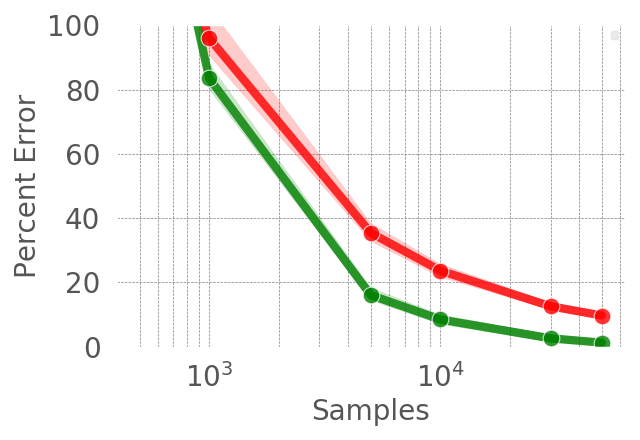} &
    \includegraphics[width=3.9cm]{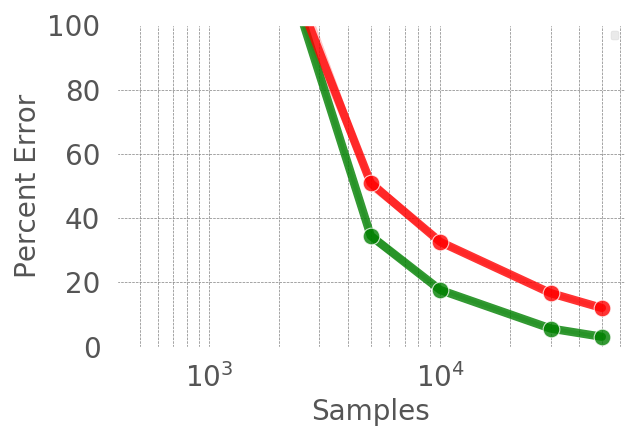}\\
    \includegraphics[width=3.9cm]{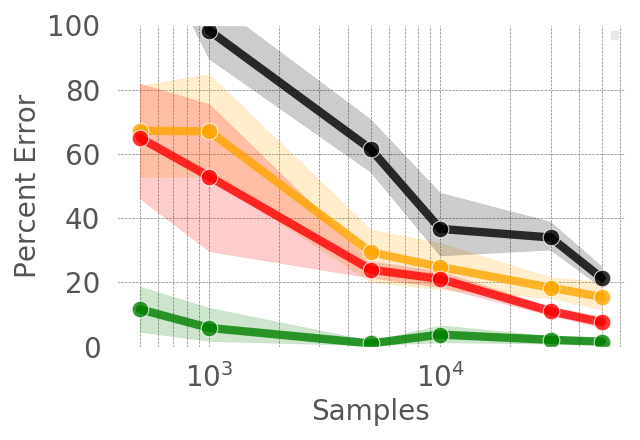} &
    \includegraphics[width=3.9cm]{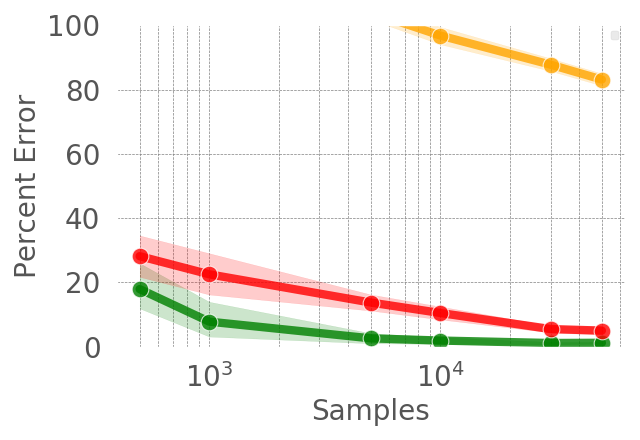} &
    \includegraphics[width=3.9cm]{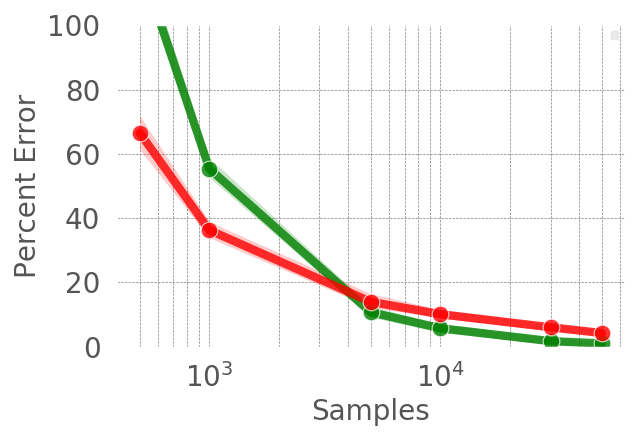} &
    \includegraphics[width=3.9cm]{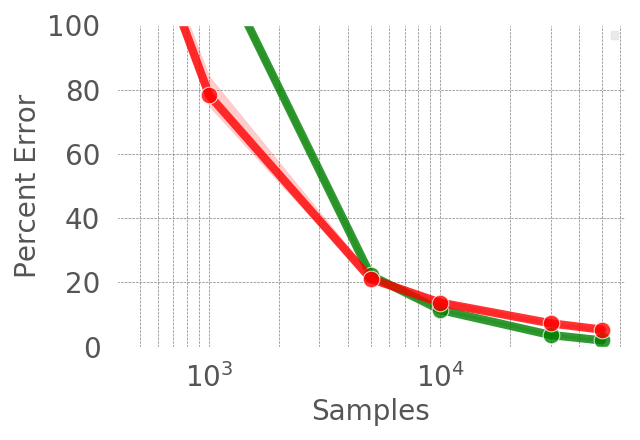}\\       \includegraphics[width=3.9cm]{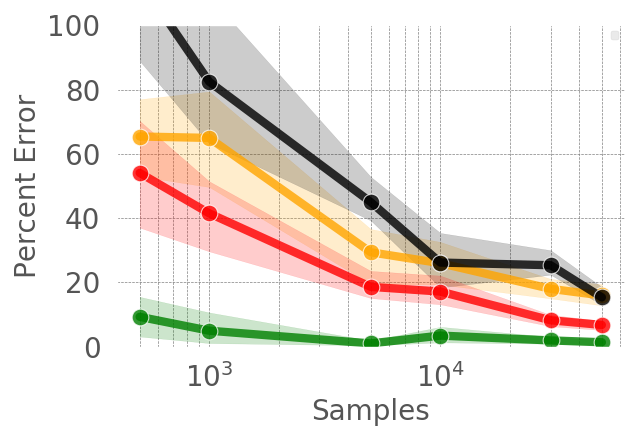} &
    \includegraphics[width=3.9cm]{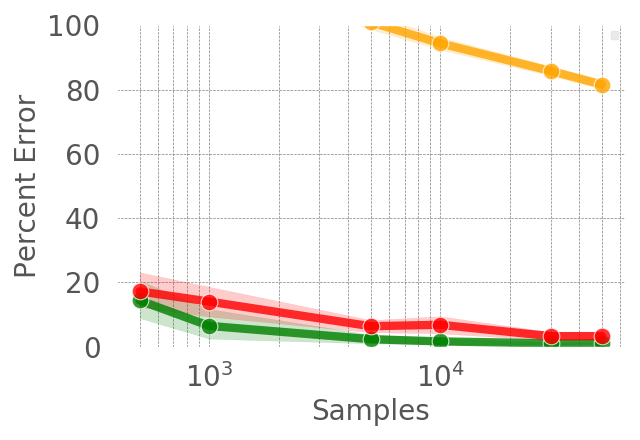} &
    \includegraphics[width=3.9cm]{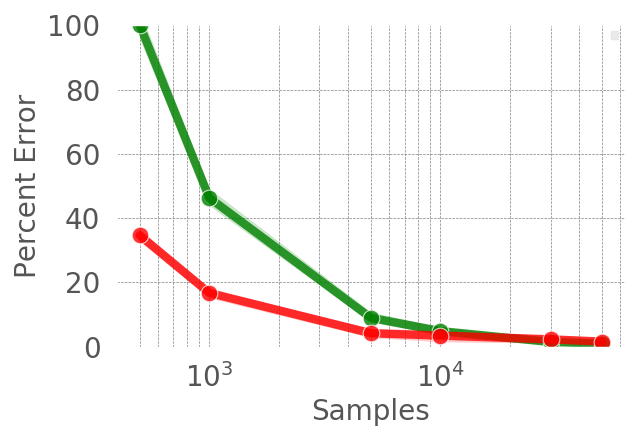} &
    \includegraphics[width=3.9cm]{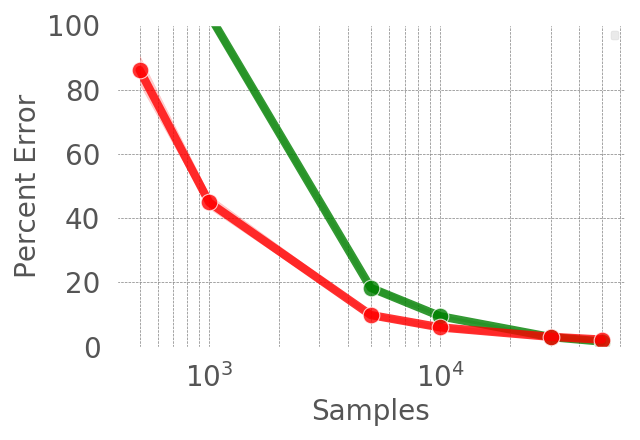}
    \end{tabular}
    \begin{tabular}{m{150mm}}
    \includegraphics[width=3.9cm, trim={0 100mm 0 0}, clip]{Figures/experiments/FIGS_KLD/legend.png}\\
    \end{tabular}
            \caption{$D_{KL}$ estimation results for the Gaussian distribution when modifying the standard deviation (see sec. \ref{sec:kld} for detail). The $\sigma_2$ value tested is respectively in each row: $[0.5,0.75,0.9]$. }
    \label{fig:KLD_gauss_vs_gauss_std}
\end{figure*}

\begin{figure*}
    \centering
    \begin{tabular}{cccc}
    D = 3 & D = 10 & D = 50 & D = 100 \\
    \includegraphics[width=3.9cm]{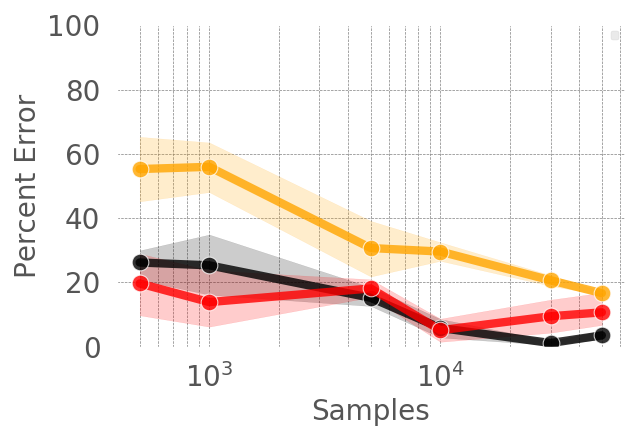} &
    \includegraphics[width=3.9cm]{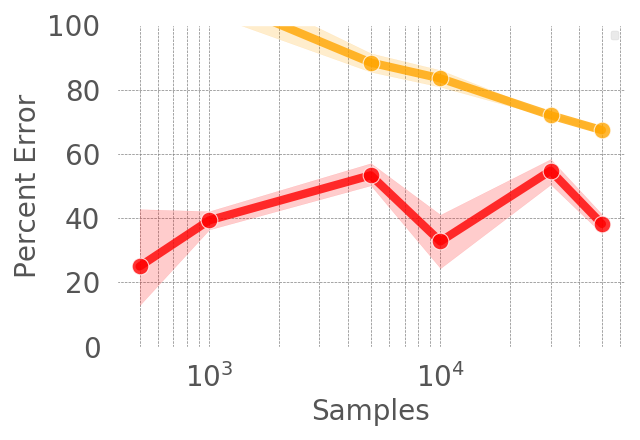} &
    \includegraphics[width=3.9cm]{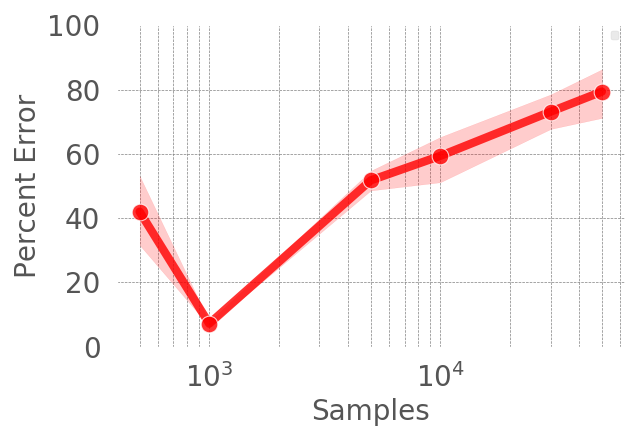} &
    \includegraphics[width=3.9cm]{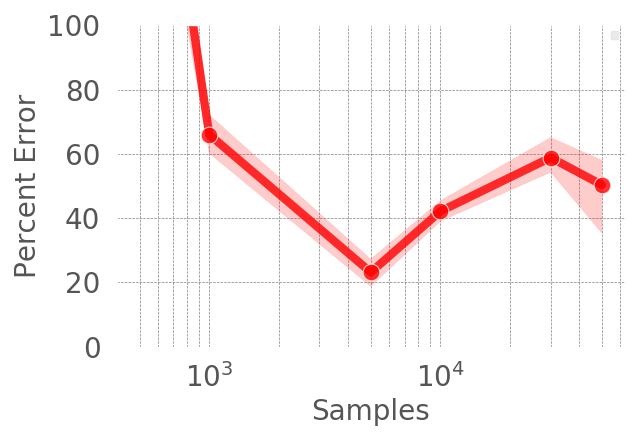}\\
    \includegraphics[width=3.9cm]{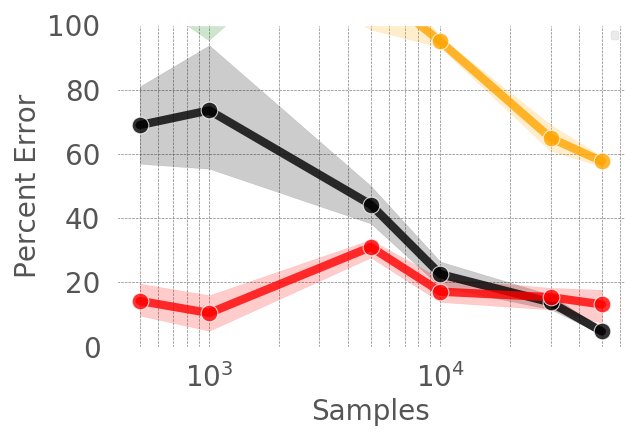} &
    \includegraphics[width=3.9cm]{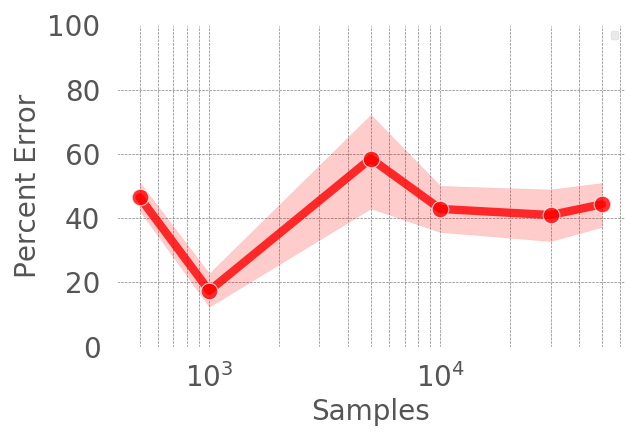} &
    \includegraphics[width=3.9cm]{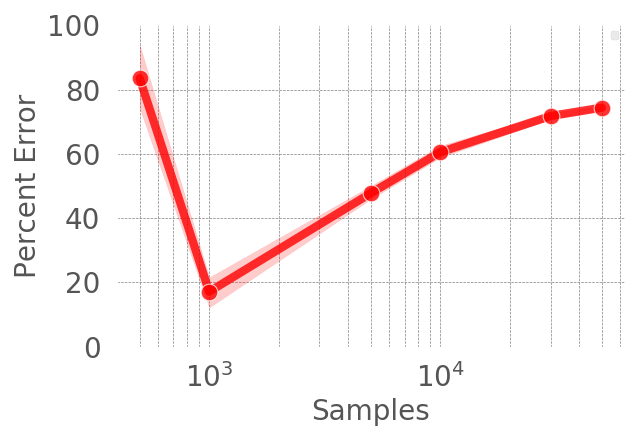} &
    \includegraphics[width=3.9cm]{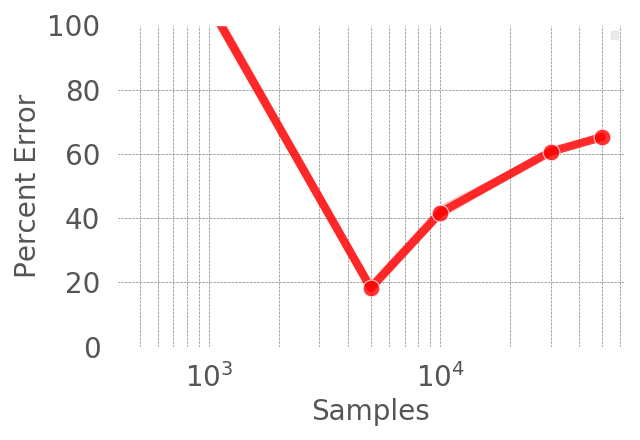}\\       \includegraphics[width=3.9cm]{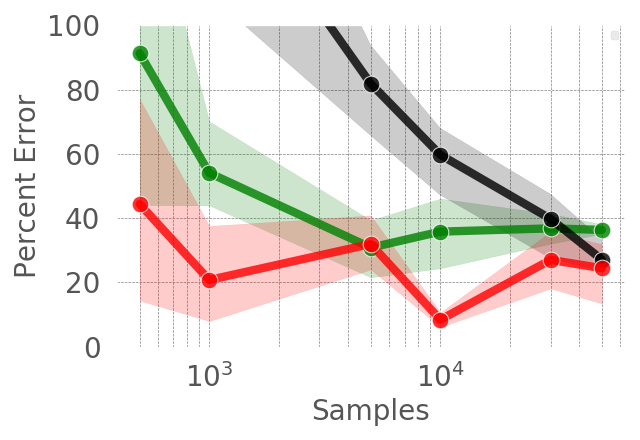} &
    \includegraphics[width=3.9cm]{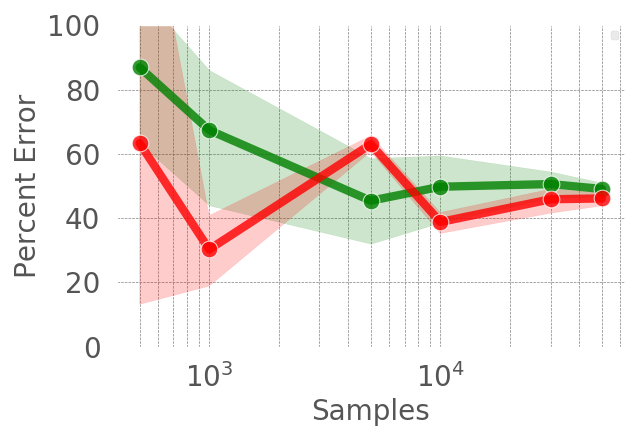} &
    \includegraphics[width=3.9cm]{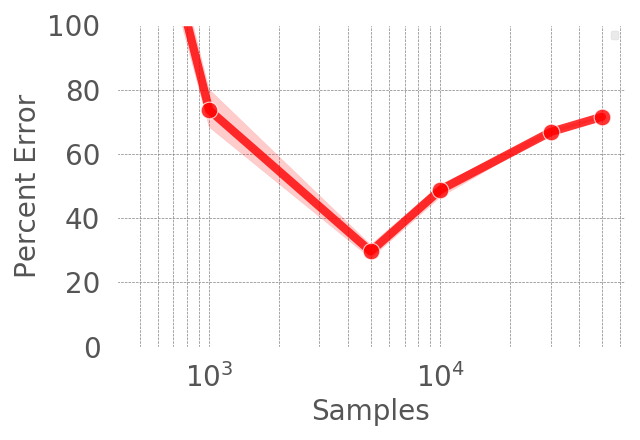} &
    \includegraphics[width=3.9cm]{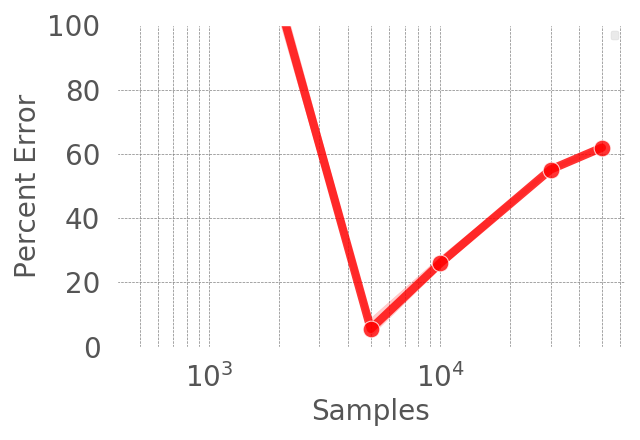}
    \end{tabular}
    \begin{tabular}{m{150mm}}
    \includegraphics[width=3.9cm, trim={0 100mm 0 0}, clip]{Figures/experiments/FIGS_KLD/legend.png}\\
    \end{tabular}
            \caption{$D_{KL}$ estimation results for the Gaussian vs multivariate Student distributions when modifying the $\nu$ (see sec. \ref{sec:kld} for detail). The $\nu$ value tested is respectively in each row: $\nu_2 = 0.2$, $\nu_2 = 0.4$ and $\nu_2 = 0.6$. }
    \label{fig:KLD_gauss_vs_tstudent}
\end{figure*}

\begin{figure*}
    \centering
    \begin{tabular}{cccc}
    D = 3 & D = 10 & D = 50 & D = 100 \\
    \includegraphics[width=3.9cm]{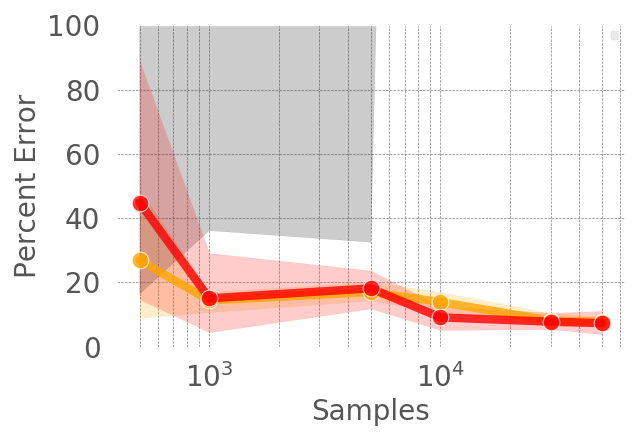} &
    \includegraphics[width=3.9cm]{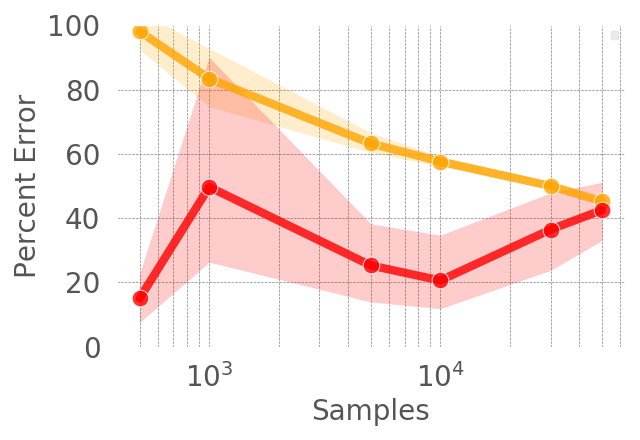} &
    \includegraphics[width=3.9cm]{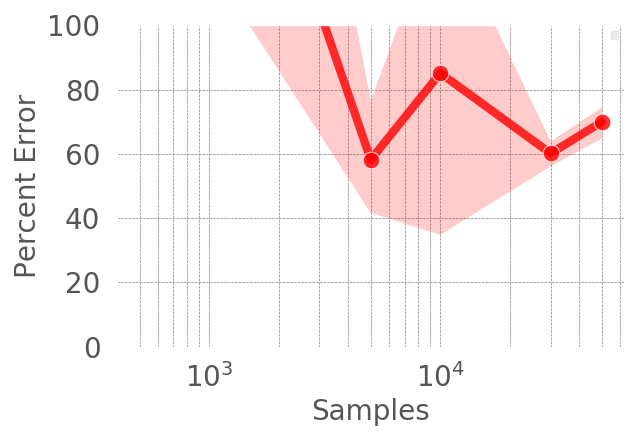} &
    \includegraphics[width=3.9cm]{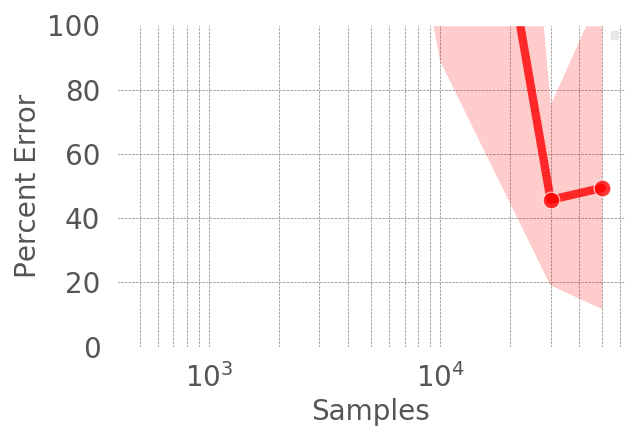}\\
    \includegraphics[width=3.9cm]{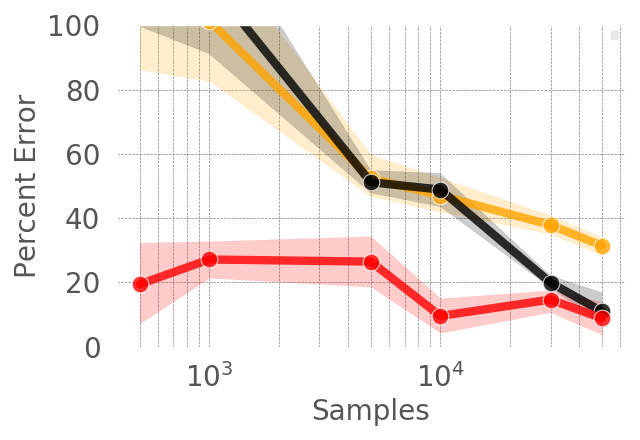} &
    \includegraphics[width=3.9cm]{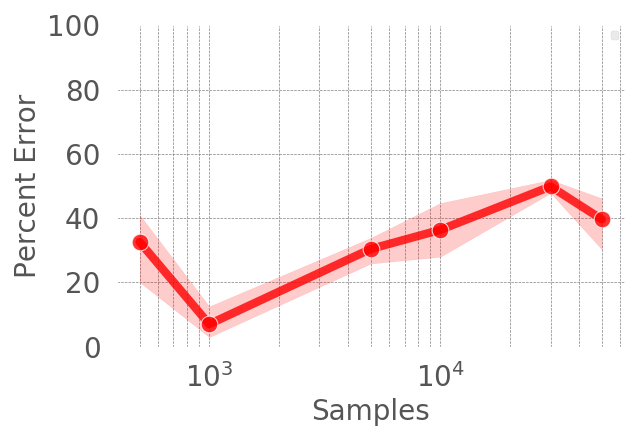} &
    \includegraphics[width=3.9cm]{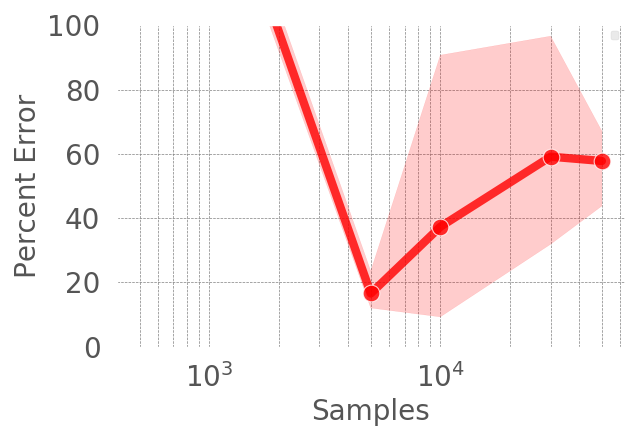} &
    \includegraphics[width=3.9cm]{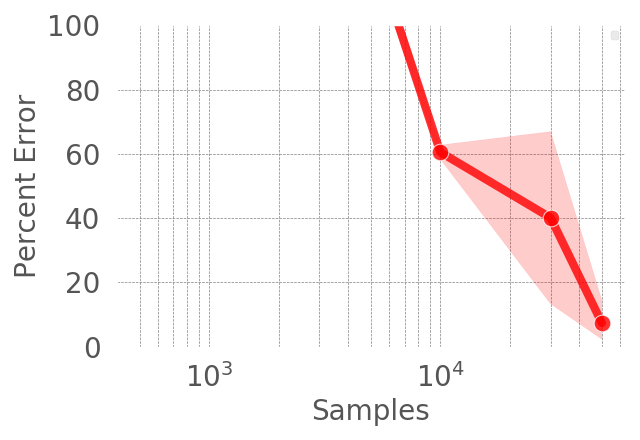}\\       \includegraphics[width=3.9cm]{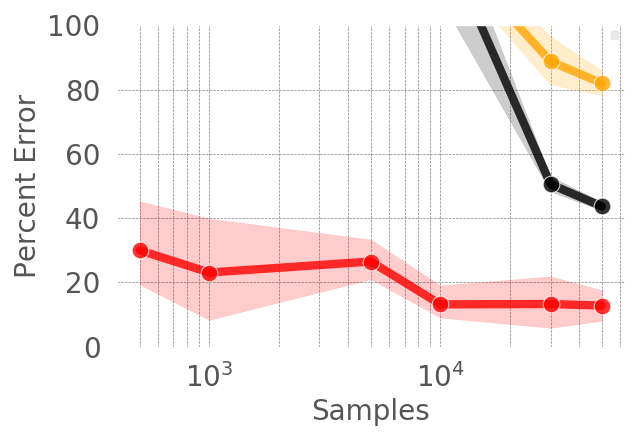} &
    \includegraphics[width=3.9cm]{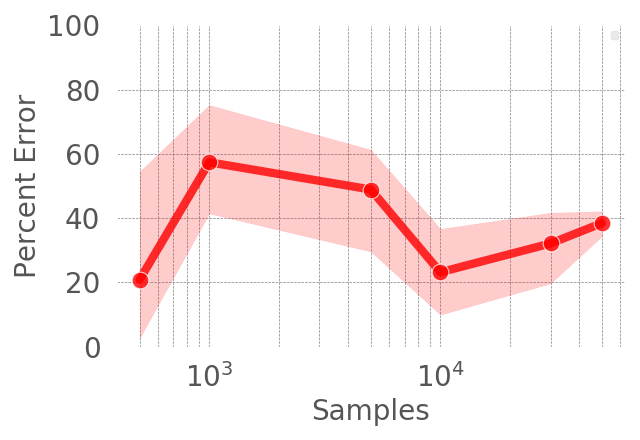} &
    \includegraphics[width=3.9cm]{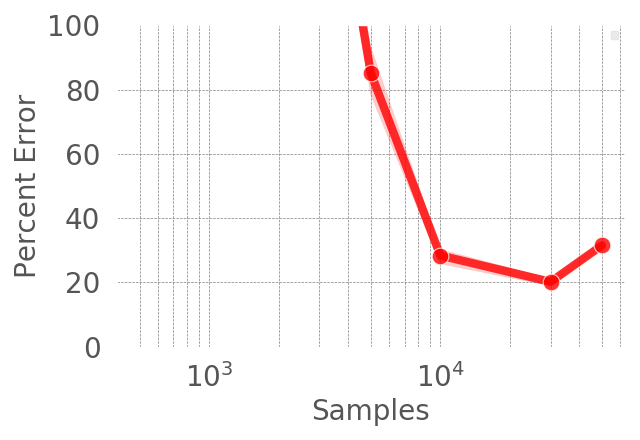} &
    \includegraphics[width=3.9cm]{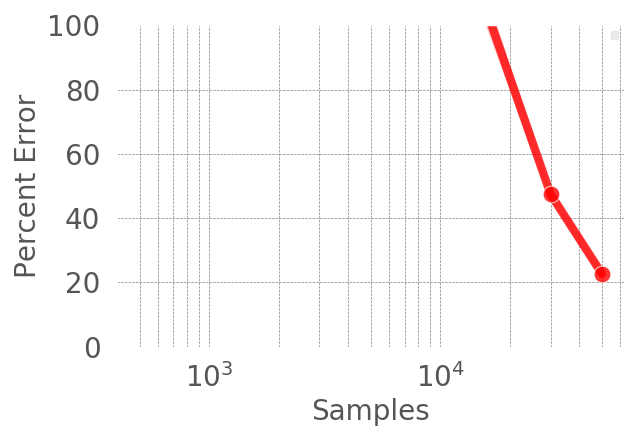}
    \end{tabular}
    \begin{tabular}{m{150mm}}
    \includegraphics[width=3.9cm, trim={0 100mm 0 0}, clip]{Figures/experiments/FIGS_KLD/legend.png}\\
    \end{tabular}
                \caption{$D_{KL}$ estimation results for the multivariate Student vs multivariate Student distributions when modifying the $\nu$ (see sec. \ref{sec:kld} for detail). The $\nu$ value tested is respectively in each row: $\nu_2 = 0.2$, $\nu_2 = 0.4$ and $\nu_2 = 0.6$. }
    \label{fig:KLD_tstudent_vs_tstudent}
\end{figure*}

\begin{figure*}
    \centering
    \begin{tabular}{cccc}
    D = 3 & D = 10 & D = 50 & D = 100 \\
    \includegraphics[width=3.9cm]{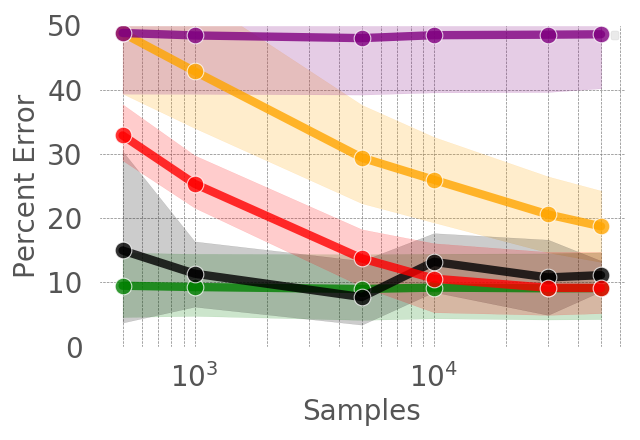} &
    \includegraphics[width=3.9cm]{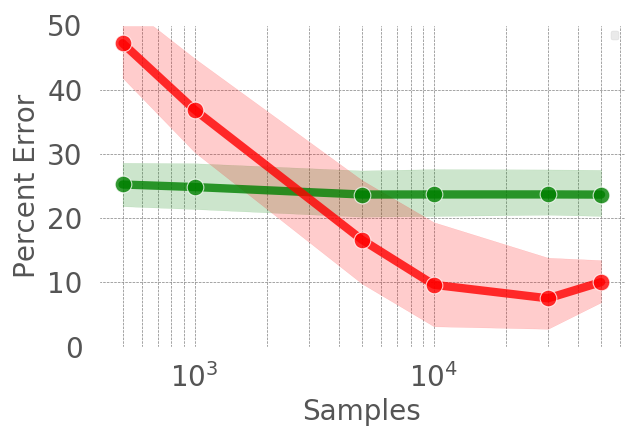} &
    \includegraphics[width=3.9cm]{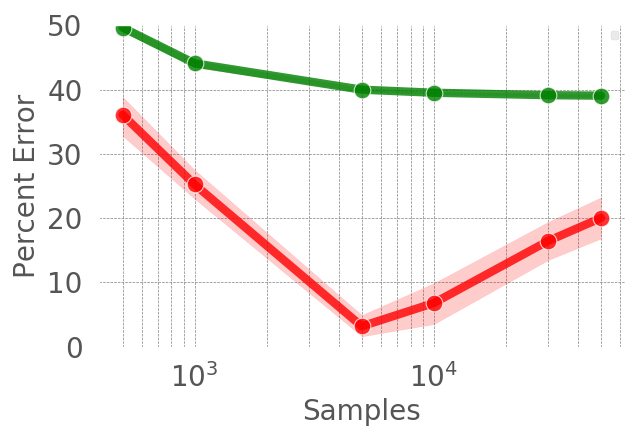} &
    \includegraphics[width=3.9cm]{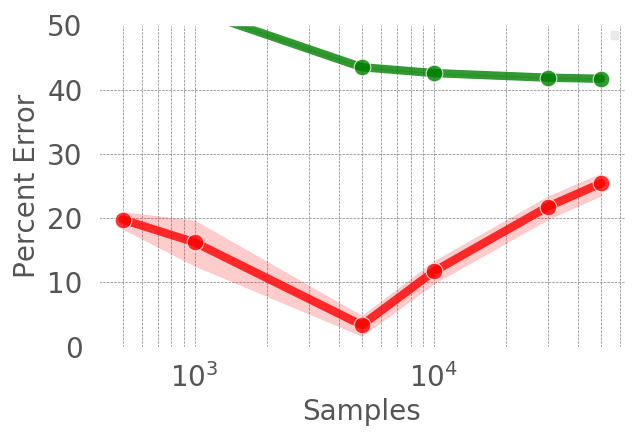}
    \end{tabular}
    \begin{tabular}{m{150mm}}
    \includegraphics[width=39mm, trim={0 100mm 0 0}, clip]{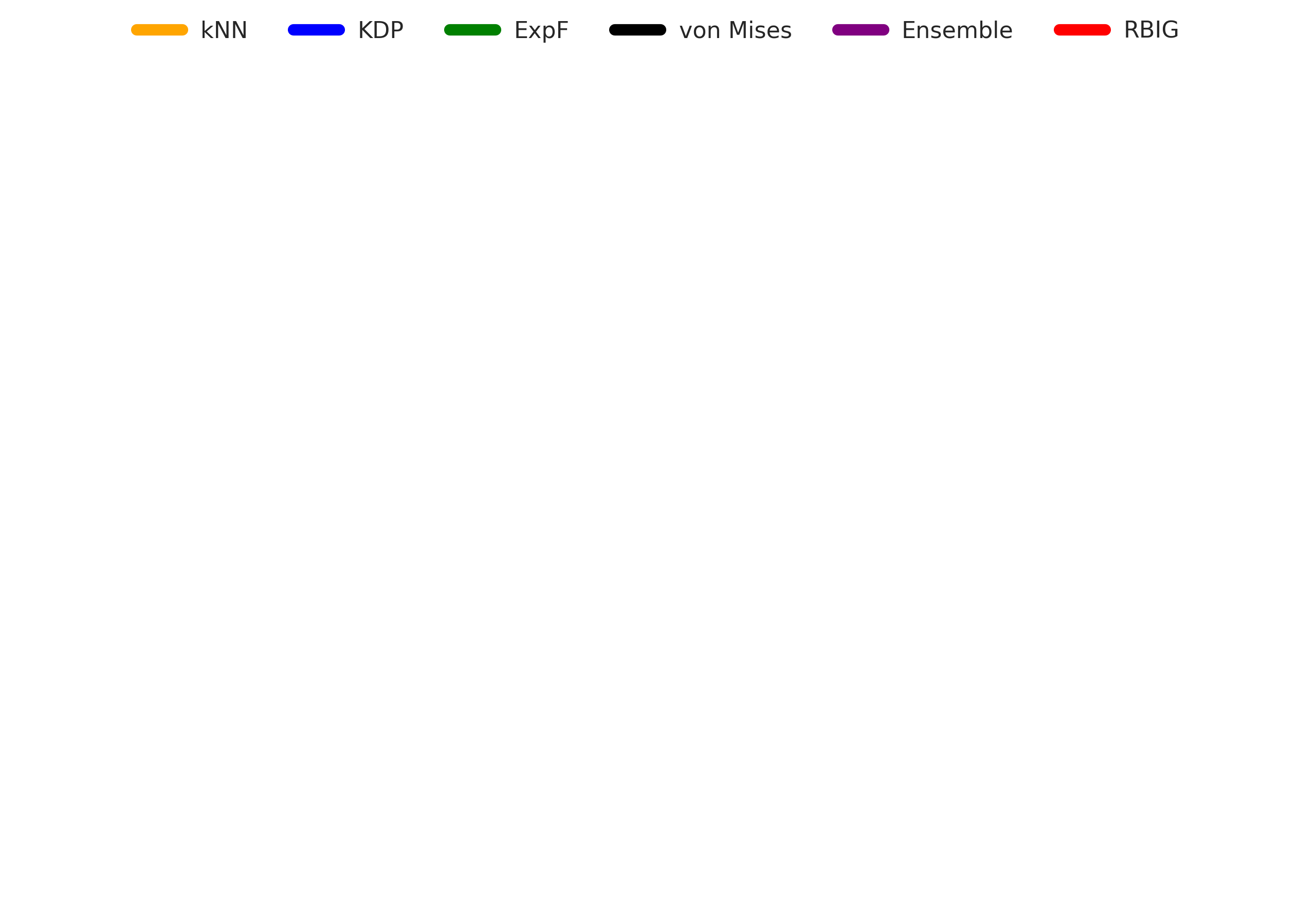} \\
    \end{tabular}
        \caption{Mutual information estimation results for the Gaussian distribution (see sec. \ref{sec:MI} for detail).}
    \label{fig:MI_gauss}
\end{figure*}

\begin{figure*}
    \centering
    \begin{tabular}{cccc}
    D = 3 & D = 10 & D = 50 & D = 100 \\
    \includegraphics[width=3.9cm]{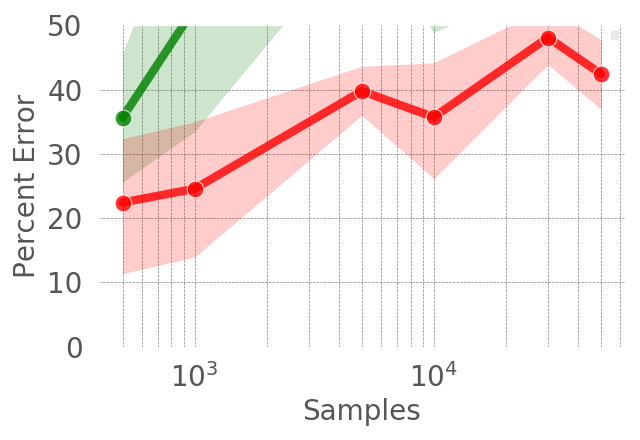} &
    \includegraphics[width=3.9cm]{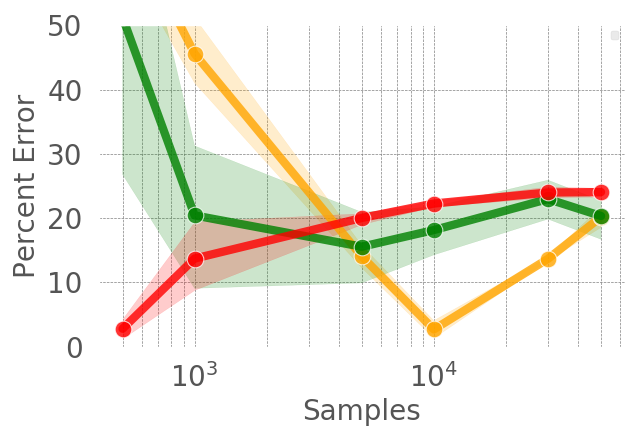} &
    \includegraphics[width=3.9cm]{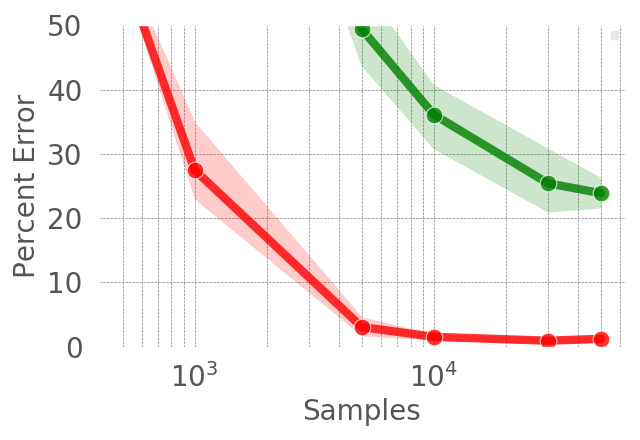} &
    \includegraphics[width=3.9cm]{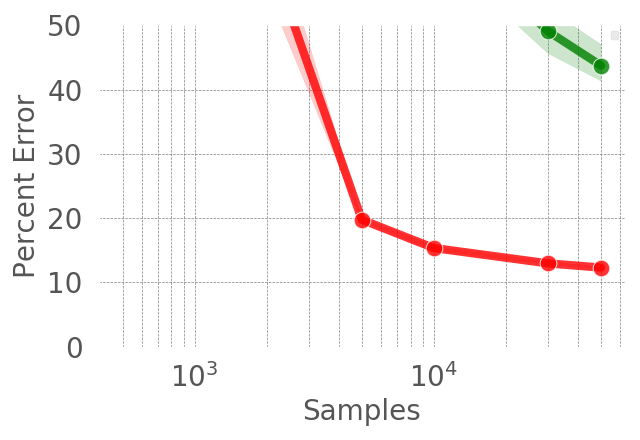}\\
    \includegraphics[width=3.9cm]{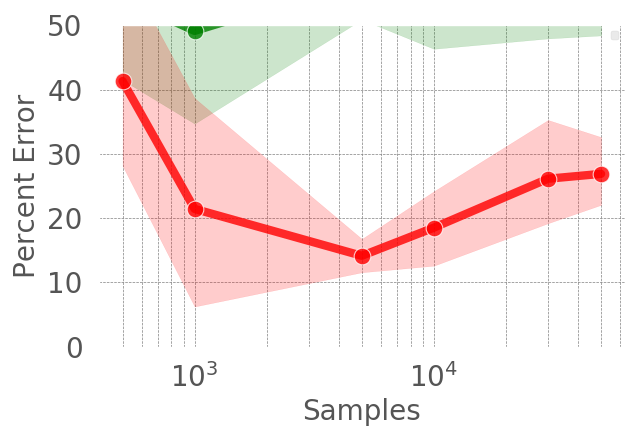} &
    \includegraphics[width=3.9cm]{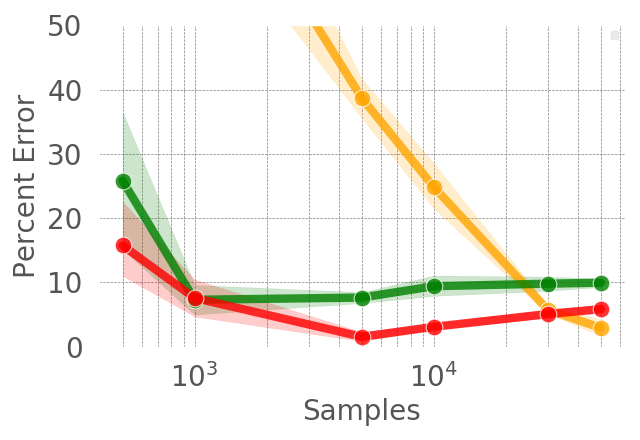} &
    \includegraphics[width=3.9cm]{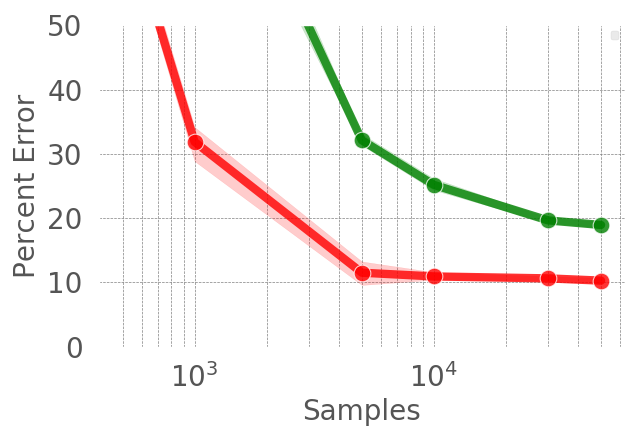} &
    \includegraphics[width=3.9cm]{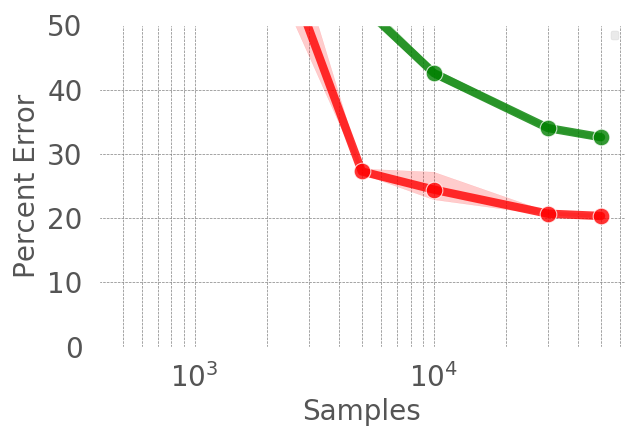}\\
    \includegraphics[width=3.9cm]{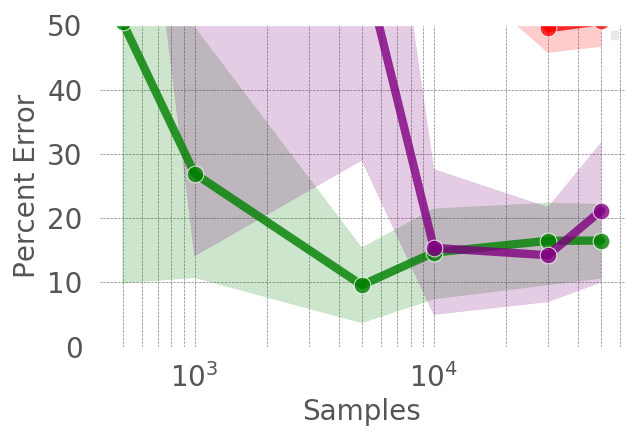} &
    \includegraphics[width=3.9cm]{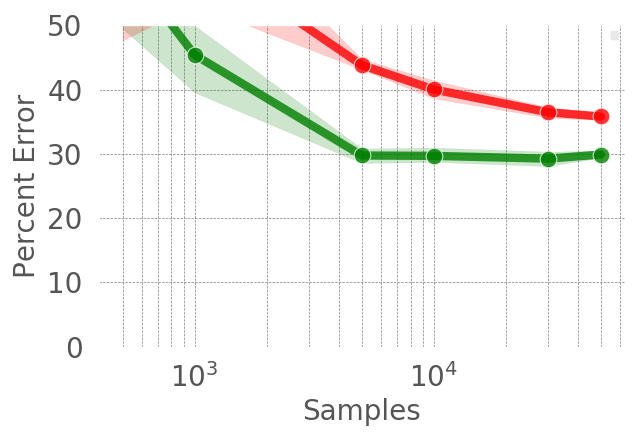} &
    \includegraphics[width=3.9cm]{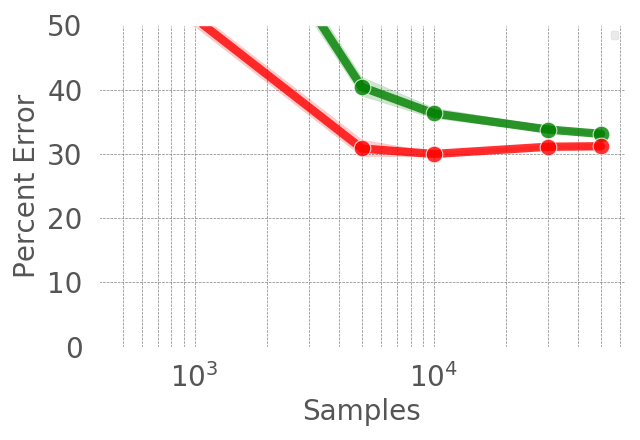} &
    \includegraphics[width=3.9cm]{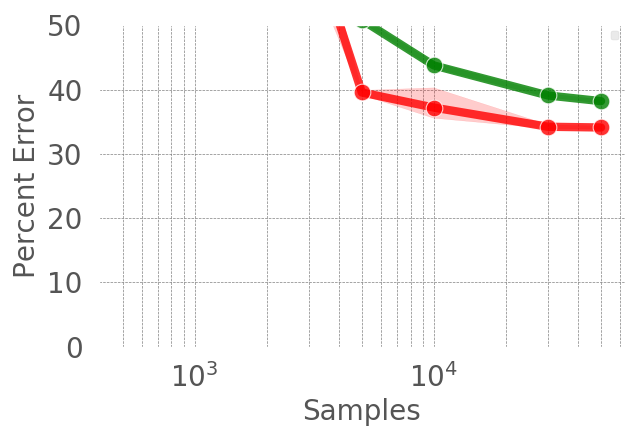}\\
    \end{tabular}
    \begin{tabular}{m{150mm}}
    \includegraphics[width=3.9cm, trim={0 100mm 0 0}, clip]{Figures/experiments/FIGS_MI/legend.png}\\
    \end{tabular}
        \caption{Mutual information estimation results for the multivariate Student distribution (see sec. \ref{sec:MI} for detail).  Each row shows results for different values of $\nu = 3, 5, 20$ respectively.}
    \label{fig:MI_stud}
\end{figure*}

\end{document}